\documentclass{article}
\usepackage{iclr2027_conference,times}

\usepackage{amsmath,amsfonts,bm}

\def\eqref#1{equation~\ref{#1}}

\def\floor#1{\lfloor #1 \rfloor}
\def\1{\bm{1}}

\DeclareMathAlphabet{\mathsfit}{\encodingdefault}{\sfdefault}{m}{sl}
\SetMathAlphabet{\mathsfit}{bold}{\encodingdefault}{\sfdefault}{bx}{n}

\usepackage{hyperref}
\usepackage{url}
\usepackage{booktabs}
\usepackage{array}
\usepackage{multirow}
\usepackage{enumitem}
\usepackage{tcolorbox}
\usepackage{caption}
\usepackage{float}
\usepackage{wrapfig}
\usepackage{placeins}
\usepackage{pifont}
\newcommand{\cmark}{\ding{51}}
\newcommand{\xmark}{\ding{55}}

\newlength{\defaultintextsep}
\makeatletter
\patchcmd{\WF@putfigmaybe}{\lower\intextsep}{\raise1.4pt}{}{}
\makeatother
\usepackage{soul}
\definecolor{boxcolor}{HTML}{8B86B0}
\newtcolorbox[auto counter, number within=section]{promptbox}[3][]{%
    colback=white,
    colframe=boxcolor,
    title=#2,
    label={#3},
    arc=2mm,
    boxrule=1pt,
    fontupper=\ttfamily\small,
    #1
}
\title{Designing Protein Structures\\from Natural Language}

\author{%
\normalfont\bfseries\fontsize{9.5pt}{11pt}\selectfont
Joseph Lee$^{1}$, Yanjing Li$^{1,*}$, Bo Qiang$^{1,*}$, Gregory Lau$^{2}$, Allyson Ettinger$^{3}$,\\[-1pt]
\normalfont\bfseries\fontsize{9.5pt}{11pt}\selectfont
Noah A. Smith$^{1,3}$, Rory Donovan-Maiye$^{3}$, David Baker$^{1,4}$, Hanna Hajishirzi$^{1}$, Pang Wei Koh$^{1}$\\[0.5ex]
\normalfont\small
$^{1}$University of Washington \quad
$^{2}$National University of Singapore\\
\normalfont\small
$^{3}$Allen Institute for AI (Ai2) \quad
$^{4}$Howard Hughes Medical Institute\\[-0.5pt]
\normalfont\small $^{*}$Equal contribution.
}

\iclrfinalcopy
\begin{document}

\maketitle
\lhead{Preprint}
\vspace{-15.5pt}

\addtocontents{toc}{\protect\setcounter{tocdepth}{-1}}

\begin{abstract}
\vspace{2pt}
A central challenge in \textit{de novo} protein design is generating proteins with desired structural and functional properties.
Existing design models typically require a dedicated input representation for each property, limiting the design objectives they support.
We develop \textbf{Prompteina}, the first all-atom structure generation model that supports new and broader design objectives specified in natural language, analogous to text-to-image generation.
To train Prompteina, we construct \textbf{Prompteina-Data}, a paired text--protein structure dataset of 1.8M structures and 50M+ captions describing global and residue-specific structural and functional properties.
We adapt Proteina-Complexa, a pretrained structure generation model, through text-conditioned pretraining and finetuning that aligns residues with their corresponding text.
Prompteina outperforms baselines on 15 of 18 structural properties and generates proteins that more closely match descriptions of their intended functions.
In binder design, it leads baselines on at least 13 of 20 targets and improves interface confidence by 14\% over the strongest baseline.
With iterative in silico feedback, Prompteina's gain over non-iterative generation is 57\% larger than the next-best baseline on the hardest targets.
Finally, we experimentally validate Prompteina designs against GM2A, a target with no reported \textit{de novo} binders.
We identify nine binders, with the strongest at an apparent $\mathrm{K}_{\mathrm{D}}$ of 107~nM.
These results establish natural language as an effective and powerful conditioning interface for protein structure generation.
Code, data and model weights: \href{https://github.com/jleechung/prompteina}{\raisebox{-0.15em}{\includegraphics[height=0.9em]{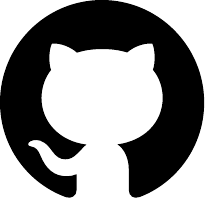}}}.

\end{abstract}
\vspace{-3pt}

\begin{figure}[h]
  \centering
  \includegraphics[width=\linewidth]{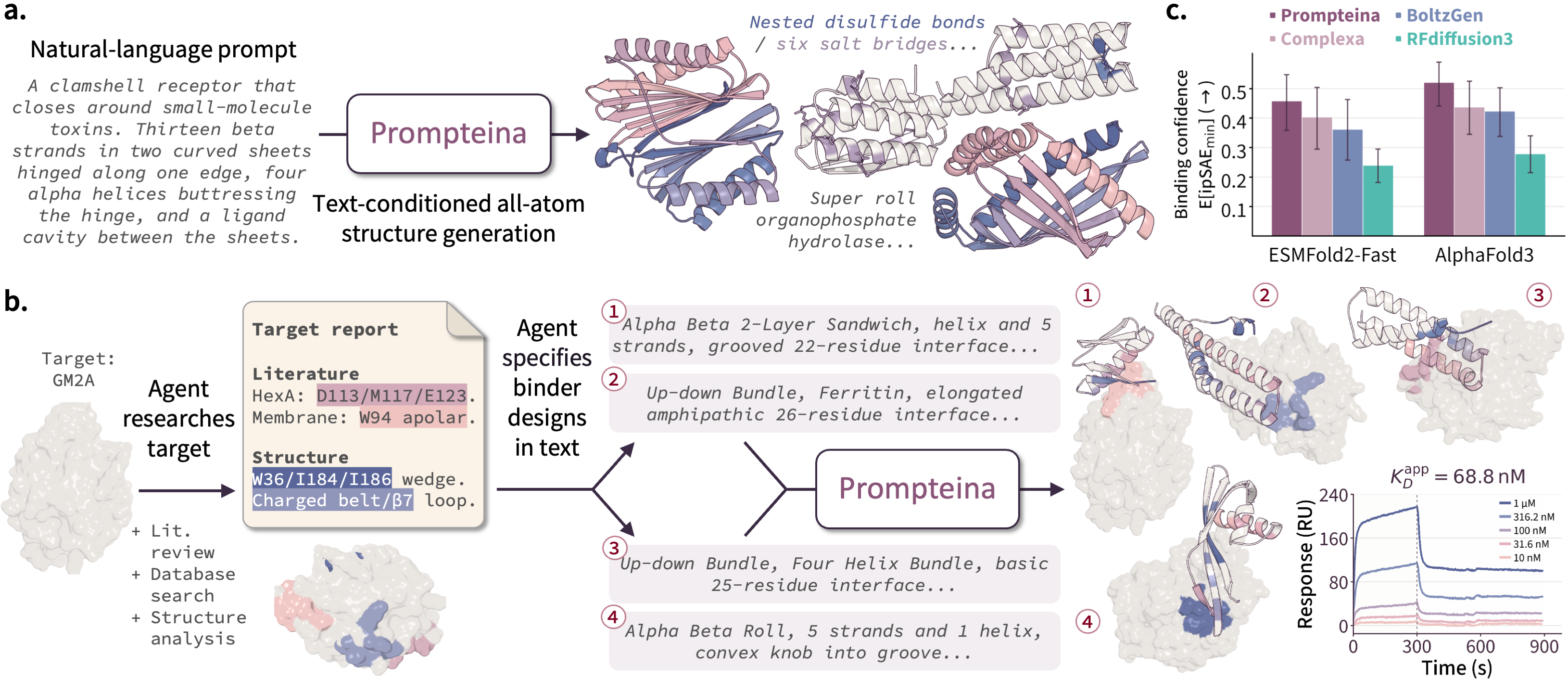}
  \caption{\textbf{Prompteina generates all-atom protein structures directly from natural language.}
  \textbf{(a)} A single natural-language interface supports diverse design objectives spanning global and residue-specific properties.
  \textbf{(b)} In agentic binder design, an agent researches target biology and writes text prompts describing candidate binders.
  Prompteina then realizes these research-informed specifications directly in generated structures.
  Experimental validation against GM2A identifies nine diverse binders, with the best replicate at an apparent $K_{\mathrm{D}}$ of 68.8~nM.
  \textbf{(c)} Across 20 targets, Prompteina achieves the highest interface confidence when the same agent specifies designs for every model.}
  \label{fig:overview}
\end{figure}

\clearpage
\section{Introduction}

A central challenge in \textit{de novo} protein design is generating proteins with specified properties, from their overall fold or function to properties of the individual residues that make up the protein.
Existing protein structure generation models allow control over such properties, but a custom input representation must typically be engineered for each property to condition generation \citep{watsonNovoDesignProtein2023a,geffnerProteinaScalingFlowbased2025a,starkBoltzGenUniversalBinder2026}.
For instance, a protein's desired fold may be represented by a single categorical label, and its desired secondary structure by a sequence that labels each residue as part of a helix, strand, or loop.
While this form of conditioning provides precise control, the need to define a separate representation for each property limits the range of design objectives that any one model can support.

An ideal conditioning approach would support a diverse set of properties without the need to engineer a custom representation for each.
Image generation provides a precedent: natural language offers a unified conditioning interface for expressing diverse visual attributes that would otherwise require task-specific labels or controls \citep{reedGenerativeAdversarialText2016a, rombachHighResolutionImageSynthesis2022a}.

We introduce \textbf{Prompteina}, the first text-conditioned all-atom protein structure generation model that supports diverse design objectives through a single natural-language interface.
Given an input prompt, Prompteina jointly generates an all-atom protein structure and sequence conditioned on the prompt's text embeddings.
By contrast, existing approaches either use text to guide structure generation only at inference time \citep{ingrahamIlluminatingProteinSpace2023a} or use it to generate sequences that must then be folded into structures \citep{daiNovoProteinDesign2025,peiBioMatrixComprehensiveBiological2026}.
This limits robust, fine-grained control over both structure and sequence.
By directly conditioning the generation of both modalities on text, Prompteina enables steering of global properties such as fold class alongside residue-specific properties including specific disulfide bonds and binder--target contacts (Fig.~\ref{fig:overview}; App.~\ref{app:fig1-prompts}).
Prompteina also complements work on agentic protein design by letting language-model agents pass specifications directly to the structure generator in natural language rather than translate them into fixed, model-specific inputs that constrain which design decisions they can express \citep{shanehsazzadehAutonomousNovoProtein2026}.

Training Prompteina for direct, fine-grained control requires paired text--structure data describing both global and residue-specific properties.
Existing text--protein datasets, however, pair sequences with only high-level functional descriptions \citep{daiNovoProteinDesign2025,songInstructProNaturalLanguage2026}.
We construct \textbf{Prompteina-Data}, comprising 1.8M structures and 52.4M captions describing both functional and structural properties.
We transform annotations mined from databases and scientific literature or computed directly from structures into diverse natural-language captions and annotate, where possible, correspondences between phrases in the captions and the residues they describe (\S\ref{sec:data_pipeline}).
Using this data, we extend Complexa \citep{didiScalingAtomisticProtein2026}, a pretrained all-atom protein structure generation model, with two text-conditioning pathways (\S\ref{sec:text_conditioning}).
We adapt the model at scale through text-conditioned continual pretraining on Prompteina-Data, then fine-tune it with direct supervision of the correspondences between caption text and residues for finer residue-level control (\S\ref{sec:training}).

Across five groups of structural properties, Prompteina leads evaluated baselines on 15 of 18 tasks, supports more properties than any single baseline, and generates proteins that more closely match descriptions of their intended functions (\S\S\ref{sec:exp-structural}--\ref{sec:exp-functional}).
At finer structural scales, Prompteina steers the number of residues at the binder--target interface across 20 targets ($\rho > 0.99$, \S\ref{sec:exp-nres-steering}).
Beyond steering individual interface properties, we evaluate Prompteina in agent-guided protein binder design.
The same language-model agent writes specifications based on the properties supported by each model's conditioning interface.
Prompteina outperforms every baseline on at least 13 of 20 targets and raises interface confidence by 14\% over the strongest baseline (\S\ref{sec:exp-ra-binder}).
When we allow the agent to iteratively refine designs, Prompteina shows larger gains over matched-budget i.i.d. generation than either baseline, especially on the most difficult targets (\S\ref{sec:exp-iteropt}).
As a proof of concept, we conducted a prospective, agent-guided design campaign against GM2 ganglioside activator protein (GM2A), a therapeutic target implicated in GM2 gangliosidosis \citep{deschenesBiochemicalCorrectionGM22023}.
GM2A has no reported \textit{de novo} binders and was one of three targets without a hit in a prior campaign \citep{starkBoltzGenUniversalBinder2026}.
In our campaign, nine of 89 assayed designs were classified as binders, with the strongest yielding an overall apparent $K_{\mathrm{D}}$ of 107~nM across three replicates (\S\ref{sec:exp-gm2a}).

These results establish natural language as an effective conditioning interface, enabling structural and functional control across scales and integrating language-based reasoning and iterative refinement into protein design.
We make our code, data, and model weights available on \href{https://github.com/jleechung/prompteina}{GitHub}.

\section{Text-guided Protein Structure Generation}
\label{sec:methods}

\begin{figure}[t]
  \centering
  \includegraphics[width=\linewidth]{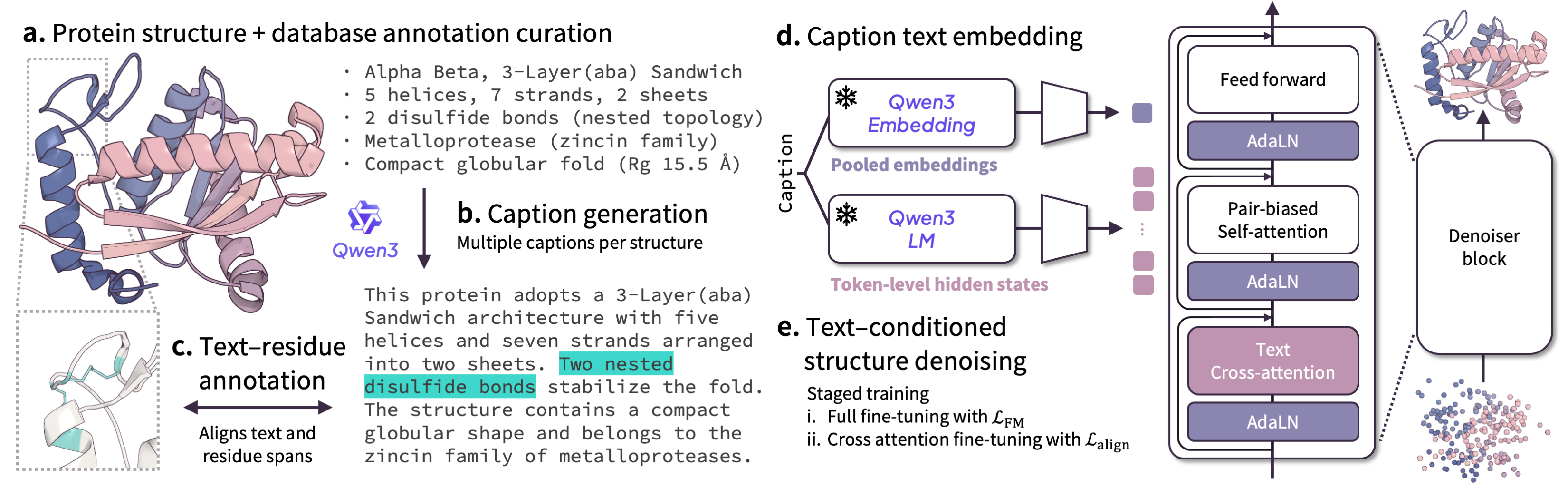}
  \caption{\textbf{Paired text--structure data and text-conditioned structure generation}.
  \textbf{(a)} Structural and functional annotations are curated and \textbf{(b)} passed to a large language model for caption generation.
  \textbf{(c)} Residues are aligned to their corresponding caption text spans for supervised alignment fine-tuning.
  \textbf{(d)} Captions are embedded into pooled and per-token text representations, then projected to condition the structure generation model.
  \textbf{(e)} The denoiser is first trained on caption--structure pairs with a flow-matching objective, then fine-tuned with supervised text--residue alignment (\S\ref{sec:alignment_head}).
  }
  \label{fig:2}
\end{figure}

A natural-language conditioning interface should let protein designers specify properties across structural scales, from global properties of the whole protein, such as fold class and overall shape, to residue-specific properties, such as disulfide bonds, catalytic sites, and binder--target contacts.
We therefore construct Prompteina-Data, a dataset pairing protein structures with natural-language descriptions spanning these scales (\S\ref{sec:data_pipeline}), and use it to train Prompteina, an all-atom structure generation model conditioned on natural language for monomer and binder design (Fig.~\ref{fig:2}; \S\S\ref{sec:text_conditioning}--\ref{sec:training}).

\subsection{Prompteina-Data: Text--Structure Data}
\label{sec:data_pipeline}

We construct Prompteina-Data from experimental structures in the Protein Data Bank \citep[PDB;][]{bermanProteinDataBank2000a}, predicted structures in the AlphaFold Database \citep[AFDB;][]{varadiAlphaFoldProteinStructure2022}, and Teddymer synthetic dimers \citep{didiScalingAtomisticProtein2026} (App.~\ref{app:data-sources}).
Prompteina generates one chain at a time, so we treat each chain as a separate entry, giving about 1.8M entries.
The dataset includes up to 59 annotations per entry, derived from database fields and atomic coordinates (Fig.~\ref{fig:2}a; App.~\ref{app:fact-extraction}).
Annotations span global properties, including fold class and compactness, and residue-specific properties with residue positions, such as disulfide bonds, catalytic sites, and protein--protein contacts.

At inference, prompts may specify one property or combine several, so we vary caption content and length in the training data \citep{wuQwenImageTechnicalReport2025b} (App.~\ref{app:residue-level-captions}).
For each entry, we sample several subsets that differ in the number and choice of annotations, then prompt Qwen3-235B-A22B \citep{yangQwen3TechnicalReport2025a} to write one caption from each subset (Fig.~\ref{fig:2}b; App.~\ref{app:caption-sampling}).
In total, we generate 52.4M captions, averaging 29 captions per entry.
We use an LLM judge to compare 80 generated captions with their ground-truth annotations (App.~\ref{app:caption-prompt}).
Across this sample, the generated captions match their ground-truth annotations closely (97.8\% recall, 89.8\% precision, and 95.7\% accuracy).

\subsection{Prompteina: Text-Conditioned Structure Generation}
\label{sec:text_conditioning}

Our model, Prompteina, builds on the pretrained Complexa model \citep{didiScalingAtomisticProtein2026} (App.~\ref{app:base-model}), adding two text-conditioning pathways and an auxiliary text--residue alignment objective.

\paragraph{Global and residue-level text conditioning.}
Complexa's denoiser generates backbone coordinates and one side-chain latent per residue before a frozen variational autoencoder decodes both into an all-atom structure and amino acid sequence.
To give every residue the same conditioning signal for global properties, we broadcast a pooled embedding from frozen Qwen3-Embedding-8B \citep{zhangQwen3EmbeddingAdvancing2025a} through adaptive layer normalization.
For residue-specific properties, each residue attends to frozen Qwen3-8B token states \citep{yangQwen3TechnicalReport2025a} via cross-attention before protein self-attention in each denoiser layer (Fig.~\ref{fig:2}d; Apps.~\ref{app:text-encoders}--\ref{app:cross-attn}).
Two-layer MLPs map both representations to denoiser dimensions.
We ablate pooled conditioning, cross-attention, and their combination, which performs best across tasks (App.~\ref{app:ablations}).

\paragraph{Text--residue alignment.}\label{sec:alignment_head}
In text-conditioned pretraining, we train the denoiser using a flow-matching loss but do not directly supervise which residues correspond to each text span.
Prompteina-Data contains residue positions for many annotations, which we use to construct text--residue alignment annotations (Fig.~\ref{fig:2}c).
For each Prompteina-Data annotation selected for a caption, we call its associated residue positions the annotation's residue span.
A residue span contains, for example, the helix residues for a mainly alpha fold annotation and the participating residues for a disulfide bond or protein--protein contact.
We attempt to locate the corresponding text span with regular expressions and TF-IDF-weighted token matching.
If either span is unavailable, we retain the caption--structure pair but omit that alignment annotation.
We validate span matching on held-out captions (App.~\ref{app:alignment-annotation}).

We train an auxiliary alignment head on these annotations, following work using segmentation masks for token-level supervision in text-to-image diffusion \citep{wangTokenComposeTexttoImageDiffusion2024}.
For $N$ protein residues and $T$ caption tokens, we represent alignment annotations as a binary matrix $\mathbf{M}\in\{0,1\}^{N\times T}$, where $M_{n,i}=1$ when token $i$ is aligned to residue $n$.
The alignment head predicts a logit $A_{n,i}$ using cross-attention updates for every residue--token pair (App.~\ref{app:alignment-head}).
Let $\mathcal{V}\subseteq\{1,\ldots,N\}\times\{1,\ldots,T\}$ contain residue--token pairs used for alignment supervision, and let $\sigma$ denote the sigmoid function.
Aligned residue--token pairs are sparse, so for each supervised example we set $w_+$ to the ratio of negative to positive pairs (App.~\ref{app:alignment-head}).
The alignment loss is
\begin{equation}
\label{eq:align_loss}
\mathcal{L}_\mathrm{align}
=
-\frac{1}{|\mathcal{V}|}\sum_{(n,i)\in\mathcal{V}}
\left[
w_+M_{n,i}\log\sigma(A_{n,i})
+(1-M_{n,i})\log\!\left(1-\sigma(A_{n,i})\right)
\right].
\end{equation}
After alignment fine-tuning, Prompteina gains finer control over features such as disulfide-bond and salt-bridge counts and retains global fold control without the alignment head at inference (Table~\ref{tab:alignment-finetuning}).

\subsection{Training and Inference}
\label{sec:training}

\paragraph{Training.}
We first adapt the Complexa denoiser to natural-language conditioning using newly initialized parameters for pooled and token-level text conditioning (\S\ref{sec:text_conditioning}; App.~\ref{app:init-gating}).
These parameters are first trained with the denoiser frozen before we unfreeze it for joint text-conditioned continual pretraining on Prompteina-Data, following staged training in text-to-image models \citep{chenPixArtaFastTraining2023} (Fig.~\ref{fig:2}e; App.~\ref{app:training-stages}).
For alignment fine-tuning, we freeze the denoiser and update text projectors, cross-attention layers, and alignment head.
We describe other Complexa training modifications that improve performance and their ablations (Apps.~\ref{app:training-stages},~\ref{app:ablations}).

\paragraph{Inference.}
At inference, Prompteina takes a prompt and protein length and, for binder design, a target structure and optional target contact residues.
Prompteina samples backbone coordinates and side-chain latents with the same stochastic differential equation as Complexa and uses classifier-free guidance to adjust text-conditioning strength \citep{hoClassifierFreeDiffusionGuidance2022a} (App.~\ref{app:sampling}).
The frozen variational autoencoder decodes these into an all-atom structure and amino acid sequence.

\section{Experiments}

We show that Prompteina's natural-language conditioning interface controls structure generation from prompts specifying structural and functional properties (\S\ref{sec:exp-structure-generation}).
This control extends to protein binding, where Prompteina directly steers the binder--target interface (\S\ref{sec:exp-nres-steering}) and gives agent-written designs higher interface confidence than state-of-the-art binder design models do (\S\ref{sec:exp-ra-binder}).
Iterative revision yields the largest interface-confidence gains with Prompteina, especially on the hardest targets (\S\ref{sec:exp-iteropt}).
Finally, we prospectively design and experimentally test candidates against GM2A, identifying nine binders among 89 designs assayed by surface plasmon resonance (\S\ref{sec:exp-gm2a}).

\subsection{Text-Guided Structure Generation}
\label{sec:exp-structure-generation}

We first test Prompteina on prompts for structural properties (\S\ref{sec:exp-structural}) or protein function (\S\ref{sec:exp-functional}).

\begin{table}[t]
\centering
\begingroup
\includegraphics[width=\linewidth]{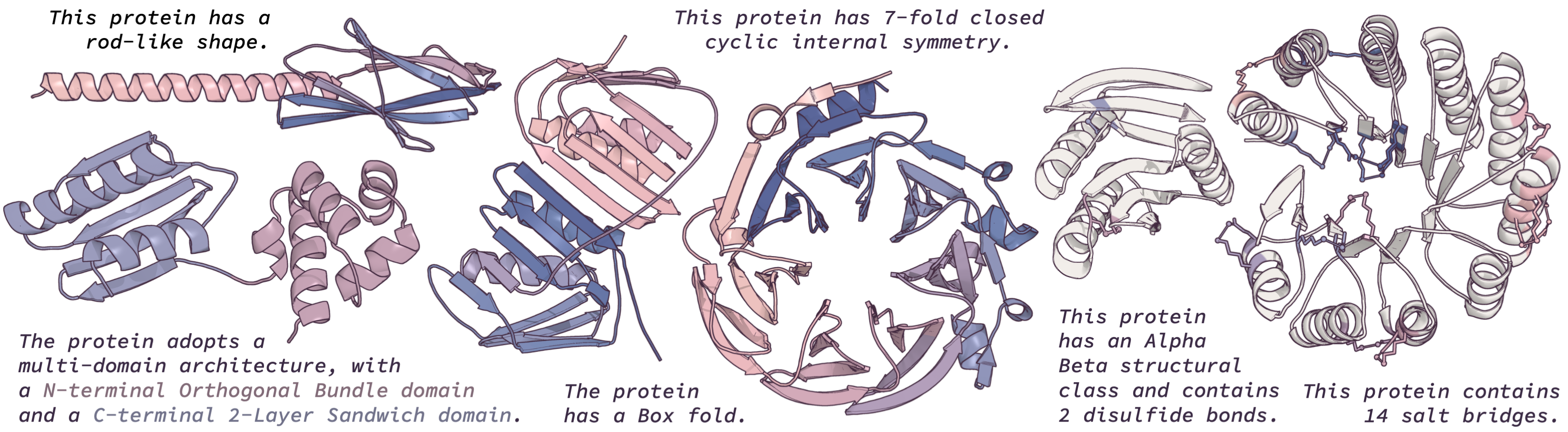}
\captionof{figure}{\textbf{Natural language controls global and residue-specific structural properties.}
Examples span global properties (shape, architecture, domain composition, and symmetry), residue-specific properties (disulfide and salt-bridge counts), and combinations of both (fold class with disulfide count).
All structures match their captions and are designable (scRMSD $<2$\,\AA).
}
\label{fig:3.1}
\endgroup

\medskip
\caption{\textbf{Prompteina controls the broadest range of structural properties.}
\textbf{(a) Text-guided models.} All models receive identical free-form prompts for 18 tasks in five property columns.
Cells average task scores by column; parentheses give task counts (full results in Table~\ref{tab:monomer_eval-full}).
$^\dagger$ denotes accuracy within 2.1 pp of chance (33.3\%).
\textbf{(b) Property-specific models.} Baselines use property-specific labels or structured inputs.
Columns show tasks supported by at least one baseline; dashes mark unsupported tasks.
We report within-prompt half-widths of 95\% stratified-bootstrap intervals.}
\label{tab:structure-control}
\small
\setlength{\tabcolsep}{2.8pt}
\begin{tabular*}{\linewidth}{@{\extracolsep{\fill}}l*{4}{>{\centering\arraybackslash}m{0.13\linewidth}}>{\centering\arraybackslash}m{0.21\linewidth}@{}}
\cmidrule(l){2-6}
\textbf{(a)} & \multicolumn{2}{c}{Accuracy (\%) ($\uparrow$)} & \multicolumn{2}{c}{Spearman $\rho$ ($\uparrow$)} & Accuracy (\%) / $\rho$ ($\uparrow$)\\
\cmidrule(lr){2-3} \cmidrule(lr){4-5} \cmidrule(lr){6-6}
\textbf{Method} & Fold (6) & Shape (3) & Secondary (3) & Tertiary (2) & Fold + sec./tert. (4) \\
\midrule
\textit{Prompteina} & \textbf{53.1} & \textbf{62.2} & \textbf{0.82} & \textbf{0.48} & \textbf{90.2 / 0.60} \\
\midrule
Chroma (text) & 10.9 & 34.2$^\dagger$ & 0.52 & 0.35 & 33.9$^\dagger$ / 0.33 \\
Pinal & 12.7 & 35.1$^\dagger$ & 0.38 & 0.17 & 35.4$^\dagger$ / 0.12 \\
BioMatrix & 9.9 & 32.7$^\dagger$ & 0.32 & 0.18 & 31.4$^\dagger$ / 0.15 \\
\bottomrule
\end{tabular*}

\vspace{0.5em}
\begin{tabular*}{\linewidth}{@{\extracolsep{\fill}}l*{6}{>{\centering\arraybackslash}m{0.12\linewidth}}@{}}
\cmidrule(l){2-7}
\textbf{(b)} & \multicolumn{3}{c}{Accuracy (\%) ($\uparrow$)} & \multicolumn{3}{c}{Spearman $\rho$ ($\uparrow$)} \\
\cmidrule(lr){2-4} \cmidrule(lr){5-7}
\textbf{Method} & CATH class & CATH arch. & Compactness & Helix & Strand & Sheet \\
\midrule
\textit{Prompteina} & 96.0 {\scriptsize$\pm$\,3.0} & \textbf{51.1} {\scriptsize$\pm$\,1.6} & \textbf{76.7} {\scriptsize$\pm$\,5.3} & \textbf{0.94} {\scriptsize$\pm$\,0.01} & 0.88 {\scriptsize$\pm$\,0.02} & 0.64 {\scriptsize$\pm$\,0.04} \\
\midrule
Chroma (label) & 64.7 {\scriptsize$\pm$\,7.7} & 7.5 {\scriptsize$\pm$\,1.0} & 39.3 {\scriptsize$\pm$\,5.3} & 0.86 {\scriptsize$\pm$\,0.02} & 0.58 {\scriptsize$\pm$\,0.04} & 0.59 {\scriptsize$\pm$\,0.05} \\
Proteina & \textbf{96.7} {\scriptsize$\pm$\,3.0} & 27.5 {\scriptsize$\pm$\,1.4} & --- & --- & --- & --- \\
RFdiffusion & --- & --- & --- & 0.87 {\scriptsize$\pm$\,0.01} & \textbf{0.92} {\scriptsize$\pm$\,0.01} & \textbf{0.82} {\scriptsize$\pm$\,0.02} \\
\bottomrule
\end{tabular*}
\end{table}

\subsubsection{Control of Structural Properties}
\label{sec:exp-structural}

\paragraph{Baselines.}
For structural control, we benchmark text-guided models that accept free-form prompts and property-specific models that take labels or structured inputs.
Text-guided models receive the same free-form prompts as Prompteina: Chroma (text) uses a structure-captioning model for inference-time text guidance \citep{ingrahamIlluminatingProteinSpace2023a}, whereas Pinal \citep{daiNovoProteinDesign2025} and BioMatrix \citep{peiBioMatrixComprehensiveBiological2026} generate amino acid sequences from natural-language descriptions, which we fold with ESMFold.
Property-specific models, comprising Chroma (label) \citep{ingrahamIlluminatingProteinSpace2023a}, Proteina \citep{geffnerProteinaScalingFlowbased2025a}, and RFdiffusion \citep{watsonNovoDesignProtein2023a}, receive their native labels or structured inputs.
For each baseline, we specify every property it supports and evaluate it only on tasks expressible through its conditioning interface.
All use paper-default settings (App.~\ref{app:eval-protocol}).

\paragraph{Evaluation.}
We compare Prompteina with six baselines across 18 tasks organized into five property groups: fold and topology, secondary structure, tertiary structure, global shape and surface properties, and combinations of fold class with secondary-structure or disulfide counts (Table~\ref{tab:structure-control}a).
Complete task definitions and scoring procedures are included in App.~\ref{app:eval-protocol}.
For categorical properties, we report accuracy, and for count-valued properties such as helix count, we report the Spearman rank correlation ($\rho$) between requested and realized values measured in each generated structure.

\paragraph{Prompteina enables new and broader forms of structural control.}
Across the benchmark, Prompteina controls all five property groups and leads on 15 of 18 tasks (Table~\ref{tab:structure-control}a and Table~\ref{tab:monomer_eval-full}).
It also enables CATH labels for multiple domains in one protein and prompts specifying multiple structural constraints, such as fold class and helix or disulfide count (Table~\ref{tab:monomer_eval-full}).
With one natural-language interface, Prompteina nearly matches Proteina on CATH class and leads property-specific models on CATH architecture, compactness, and helix count (Table~\ref{tab:structure-control}b).
RFdiffusion performs better on strand and sheet counts because it accepts per-residue secondary-structure inputs.
Prompteina's structural control is robust to paraphrasing (App.~\ref{app:monomer-paraphrase}) and preserves designability and novelty, with a higher designable fraction than every baseline on shared tasks and more novel designable structures than Complexa (Tables~\ref{tab:monomer-designability} and~\ref{tab:monomer-novelty}).
We show representative Prompteina samples (Fig.~\ref{fig:3.1}); additional designable examples compare each generated structure with its corresponding refold (Fig.~\ref{fig:monomer-designable-examples}).

\FloatBarrier
\subsubsection{Generation from Functional Specifications}
\label{sec:exp-functional}

We next test whether Prompteina can generate structures for protein families, such as lysozymes and green fluorescent proteins, from functional prompts.
In this setting, we compare against Pinal \citep{daiNovoProteinDesign2025} and BioMatrix \citep{peiBioMatrixComprehensiveBiological2026}, which generate sequences from functional prompts; we fold their outputs with ESMFold.
Since functional prompts lack a directly verifiable structural property, we evaluate each generated protein with ProTrek \citep{suTrimodalProteinLanguage2025a}, a contrastive encoder of sequences, structures, and text.
We report similarities between each prompt and its generated sequence or structure.
A high ProTrek score is less meaningful for a low-confidence predicted fold, so we also weight each score by predicted-structure confidence (pLDDT).
We use four prompt formats for all 122 families, including at least one from each model's training distribution.

\begin{table}[h]
\centering
\caption{\textbf{Prompteina leads on functional prompts.}
Each cell reports a model's highest mean across four prompt formats, with a 95\% bootstrap CI over 122 families.
pLDDT-weighted scores multiply each ProTrek similarity by mean per-residue pLDDT divided by 100.
No samples are filtered.}
\label{tab:functional_summary}
\small
\begin{tabular}{@{}lcccc@{}}
\toprule
& \multicolumn{2}{c}{\textbf{Sequence similarity} ($\uparrow$)} & \multicolumn{2}{c}{\textbf{Structure similarity} ($\uparrow$)} \\
\cmidrule(lr){2-3} \cmidrule(lr){4-5}
\textbf{Method} & Unweighted & pLDDT-weighted & Unweighted & pLDDT-weighted \\
\midrule
\textit{Prompteina} & \textbf{8.46} {\scriptsize$\pm$\,0.98} & \textbf{7.12} {\scriptsize$\pm$\,0.88} & \textbf{7.36} {\scriptsize$\pm$\,1.04} & \textbf{6.20} {\scriptsize$\pm$\,0.90} \\
\midrule
Pinal      & 7.55 {\scriptsize$\pm$\,0.79} & 5.98 {\scriptsize$\pm$\,0.70} & 7.09 {\scriptsize$\pm$\,0.73} & 5.49 {\scriptsize$\pm$\,0.68} \\
BioMatrix  & 5.73 {\scriptsize$\pm$\,0.68} & 3.70 {\scriptsize$\pm$\,0.58} & 4.64 {\scriptsize$\pm$\,0.63} & 3.20 {\scriptsize$\pm$\,0.56} \\
\bottomrule
\end{tabular}
\end{table}

\paragraph{Prompteina generates structures consistent with functional specifications.}
At each model's best prompt format, Prompteina achieves the highest mean on both ProTrek similarity metrics (Table~\ref{tab:functional_summary}).
With pLDDT weighting, it significantly outperforms BioMatrix on both metrics and Pinal on sequence similarity, while retaining the highest mean ProTrek structure similarity (Table~\ref{tab:functional_summary}; App.~\ref{app:functional-eval}).

\FloatBarrier

\subsection{Text-Guided Protein Binder Design}
\label{sec:exp-binder}

Designing strong binders requires control over both global binder properties, such as fold class, and the size and chemistry of the binder--target interface \citep{caoDesignProteinbindingProteins2022,watsonNovoDesignProtein2023a}.
We test whether Prompteina can steer these properties through natural language (\S\ref{sec:exp-nres-steering}) and produce better predicted binders than methods using predefined, property-specific inputs (\S\S\ref{sec:exp-ra-binder}--\ref{sec:exp-iteropt}).

\subsubsection{Binder--Target Interface Steering}
\label{sec:exp-nres-steering}

We first test control over binder--target interface size, which is associated with binding specificity \citep{prasadbahadurDissectionSpecificNonspecific2004}.
We use interface residue count, defined as the number of binder residues that contact the target, as a proxy (App.~\ref{app:interface-steering}).
Across 20 targets from the Complexa binder-design benchmark \citep{didiScalingAtomisticProtein2026}, we generate binders with Prompteina while requesting interface residue counts from 8 to 30 and holding the target and binder-length distribution fixed.
For each target, we compare requested and realized counts using Spearman correlation ($\rho$) and a linear fit.

\begin{figure}[t]
\centering
\includegraphics[width=\linewidth]{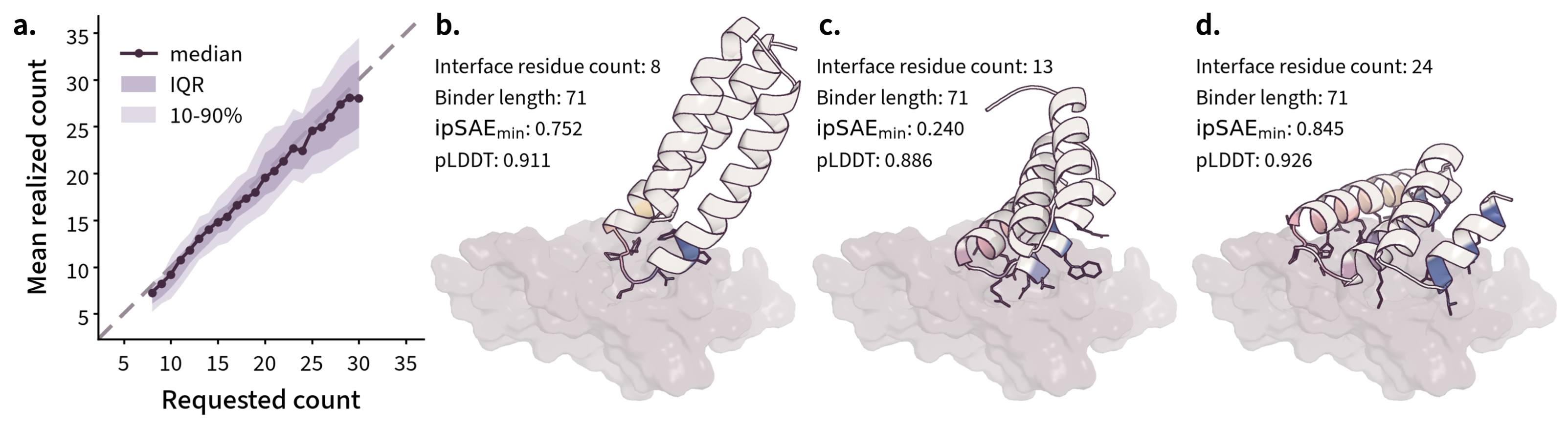}
\caption{\textbf{Prompteina steers binder--target interface residue count.}
\textbf{(a)} Requested versus mean realized count across 20 targets.
The line shows the median across targets, the bands show the interquartile and 10th--90th percentile ranges, and the dashed line marks equality.
\textbf{(b--d)} Predicted TrkA binder--target complexes whose realized counts are within one residue of the requested counts (requested: 8, 14, 24; realized: 8, 13, 24).
Binders (white cartoons) contact the target (surface) through interface residues (colored sticks).
For each complex, ipSAE$_\mathrm{min}$ ($\uparrow$) reports interface confidence and complex-averaged pLDDT ($\uparrow$) reports confidence in the predicted complex structure.}
\label{fig:interface-steering}
\end{figure}

\paragraph{Prompteina reliably steers binder--target interface residue count.}
The mean realized count tracks the requested count closely and nearly linearly across all 20 targets (Spearman's $\rho=0.994$--$1.000$ and linear-fit $R^2=0.916$--$0.996$; Fig.~\ref{fig:interface-steering-grid}).
Across the 20 targets, Prompteina steers interface residue counts below and above unconditional generation, allowing the size of the resulting binder--target interface to be tuned (Fig.~\ref{fig:interface-steering-grid}).
Figure~\ref{fig:interface-steering} shows binders for the same target with low and high realized counts.
This interface control is robust to paraphrased and reordered prompts (Table~\ref{tab:interface-paraphrase}).

\subsubsection{Agentic Protein Binder Design}
\label{sec:exp-ra-binder}

Designing a binder against a new target begins with research into the shape and chemistry of its surface, which epitopes known partners engage, and what scaffolds have succeeded against related targets.
Previous work shows that LLM agents can conduct this research and operate design tools end to end \citep{ghafarollahiProtAgentsProteinDiscovery2024, shanehsazzadehAutonomousNovoProtein2026}.
However, an agent can communicate design decisions only through the conditioning inputs a model accepts.
Existing models require a separate input representation for each property and therefore support a narrower range of properties, whereas Prompteina's single natural-language interface supports a broader range.
We test whether this broader range of supported properties allows the same agent to produce better binders.

\FloatBarrier

\paragraph{Setup and baselines.} We compare Prompteina with Complexa \citep{didiScalingAtomisticProtein2026}, RFdiffusion3 \citep{butcherNovoDesignAllatom2025b}, and BoltzGen \citep{starkBoltzGenUniversalBinder2026} on Complexa's 20-target benchmark.
For each target, Claude Opus 4.6 compiles a research dossier and writes 200 binder specifications using each model's supported input properties (App.~\ref{app:agentic-binder}).
Each model samples one structure per specification.
We also run all three baselines in a non-agentic setting, where they receive only the target structure, binder-length range, and fixed benchmark hotspots.
We do not run Complexa with the agent because its released checkpoint supports no additional input properties (App.~\ref{app:agentic-baselines}).

\paragraph{Evaluation.} We score each design under three sequence conditions: the generated sequence (\texttt{self}), full ProteinMPNN redesign (\texttt{MPNN}), and fixed-interface redesign (\texttt{MPNN-FI}) \citep{dauparasRobustDeepLearning2022a}.
ESMFold2-Fast ipSAE$_\mathrm{min}$ separates binders from non-binders more accurately than the AlphaProteo AF2 filters used by Complexa (App.~\ref{app:ipsae-calibration}), corroborating \citet{shanehsazzadehAutonomousNovoProtein2026}, so we use it to measure binder--target interface confidence throughout our binder-design experiments.

To compare methods across design budgets, let $M(k)$ be the mean ipSAE$_\mathrm{min}$ of a target's $k$ best designs, out of $N$ generated.
We report the expected top-$k$ ipSAE$_\mathrm{min}$,
\begin{equation}
\label{eq:etopk-main}
\mathbb{E}\bigl[M(K)\bigr]
=
\Bigl(\textstyle\sum_{k=1}^{N}\tfrac{1}{k}\Bigr)^{-1}
\textstyle\sum_{k=1}^{N} M(k)/k .
\end{equation}
This quantity averages $M(k)$ under a log-uniform prior on $K$, weighting equally each order of magnitude of $k$.
We compute the statistic separately for each target before averaging across all 20, and report the full comparison across design budgets for each sequence condition in Table~\ref{tab:topk-curves}.

\begin{table}[t]
\centering
\captionsetup{format=plain}
\caption{\textbf{Prompteina achieves the highest interface confidence.}
We evaluate 200 designs per method on 20 targets.
Expected top-$k$ ipSAE$_\mathrm{min}$ (Eq.~\ref{eq:etopk-main}) measures interface confidence in each sequence condition.
Diversity is one minus the mean pairwise TM-score among each method's 32 best \texttt{MPNN-FI} designs on 18 targets with at least 32 positive-scoring designs per method.}
\label{tab:binder_eval}
\footnotesize
\setlength{\tabcolsep}{4pt}
\newcolumntype{N}{>{\centering\arraybackslash}p{4.15em}}
\begin{tabular}{@{}ll @{\hspace{7pt}} NNN @{\hspace{9pt}} c@{}}
\toprule
& & \multicolumn{3}{c}{\hspace*{-9pt}Expected top-$k$ ipSAE$_\mathrm{min}$ ($\uparrow$)} & \multicolumn{1}{c}{Diversity ($\uparrow$)} \\
\cmidrule(lr){3-5} \cmidrule(lr){6-6}
\textbf{Method} & \textbf{Agent} & self & MPNN & \mbox{MPNN-FI} & 1$-$TM sim.\ \\
\midrule
Prompteina        & \cmark & \textbf{0.455} & \textbf{0.584} & \textbf{0.549} & \underline{0.584} \\
\midrule
Complexa          & \xmark & \underline{0.400} & 0.538 & \underline{0.510} & 0.520 \\
BoltzGen          & \cmark & 0.359 & 0.575 & 0.490 & 0.516 \\
BoltzGen          & \xmark & 0.376 & \underline{0.577} & 0.493 & 0.399 \\
RFdiffusion3      & \cmark & 0.237 & 0.559 & 0.454 & \textbf{0.587} \\
RFdiffusion3      & \xmark & 0.251 & 0.563 & 0.469 & 0.541 \\
\bottomrule
\end{tabular}
\end{table}

\paragraph{Prompteina achieves the highest interface confidence.} Prompteina leads in expected top-$k$ ipSAE$_\mathrm{min}$ across all three sequence conditions (Table~\ref{tab:binder_eval}).
Prompteina's lead is largest for the generated sequence and fixed-interface redesign, narrows under full ProteinMPNN redesign (Table~\ref{tab:binder-target-win-rates}), and persists under AlphaFold3 evaluation \citep{abramsonAccurateStructurePrediction2024a} (Table~\ref{tab:binder-af3-self}).
RFdiffusion3 and BoltzGen achieve similar interface-confidence scores with or without the agent, and all four variants score below Prompteina (Table~\ref{tab:binder_eval}).
Under redesign, its top designs are more diverse than either BoltzGen variant and closely match the diversity of agent-guided RFdiffusion3 (App.~\ref{app:diversity}).

\paragraph{Prompteina achieves higher binding confidence when interface properties are specified.} Prompteina lets the agent specify binder--target interface size through residue count and buried surface area, neither of which RFdiffusion3 or BoltzGen accepts (Table~\ref{tab:knob-coverage}).
Prompteina generates larger interfaces than either baseline, and within each target greater realized buried surface area is associated with higher interface confidence (App.~\ref{app:interface-properties}).
To test this directly, we rerun Prompteina on all 20 targets, first allowing only overall fold to be specified, then adding shape and finally interface properties.
We find only a modest gain from adding shape, but the largest gain from interface properties, suggesting that specifying them is important for interface confidence (Table~\ref{tab:prompt-ablation}).

\begin{table}[H]
\centering
\caption{\textbf{Binder specification ablation.}
Mean top-$k$ ipSAE$_\mathrm{min}$ across 20 targets using generated sequences.
We evaluate 200 designs per target with successively more detailed specifications.}
\label{tab:prompt-ablation}
\footnotesize
\setlength{\tabcolsep}{4pt}
\begin{tabular}{@{}l ccccc@{}}
\toprule
\textbf{Specification} & $k$=8 & 16 & 32 & 64 & Expected top-$k$ ipSAE$_\mathrm{min}$ ($\uparrow$) \\
\midrule
Prompteina (default: fold + shape + interface) & \textbf{0.534} & \textbf{0.470} & \textbf{0.376} & \textbf{0.253} & \textbf{0.455} \\
Prompteina (fold + shape)                      & 0.430 & 0.356 & 0.267 & 0.176 & 0.364 \\
Prompteina (fold only)                         & 0.412 & 0.336 & 0.250 & 0.164 & 0.362 \\
\bottomrule
\end{tabular}
\end{table}

\paragraph{Prompteina's language interface preserves rich design intent for generation.} For CD45, we trace how each model represents agent-selected binder--target interface properties (App.~\ref{app:cd45-agent-traces}).
Claude identifies distinct CD45 sites, including the charged and polar BC8 epitope and hydrophobic M47/I48/F49 patch, then proposes a fold and interface geometry for each (Fig.~\ref{fig:cd45-designs}).
All three models receive the same CD45 structure and research dossier, but only Prompteina can condition on the resulting fold, interface geometry, size, and chemistry in natural language.
For RFdiffusion3, Claude encodes a non-loop flag and, in some designs, atom-level hydrogen-bond or burial labels.
For BoltzGen, Claude encodes residue-level helix, sheet, and loop ranges (App.~\ref{app:cd45-agent-traces}).
With these richer specifications, Prompteina's two best designs outscore every baseline design while targeting CD45 sites and adopting distinct beta-rich and alpha-helical folds specified in their prompts.

\FloatBarrier
\subsubsection{Iterative Binder Prompt Optimization}
\label{sec:exp-iteropt}

Recent design models perform inference-time search by scoring partial structures during generation and retaining promising candidates \citep{didiScalingAtomisticProtein2026, rector-brooksGeneralMultimodalProtein2026a}.
Analogously, when agents call design models as tools, they can use in-silico or experimental feedback on completed designs to identify promising design strategies and revise the next batch of specifications \citep{shanehsazzadehAutonomousNovoProtein2026}.
We test whether this iterative revision yields larger gains when the agent uses Prompteina's broader conditioning interface than when it uses RFdiffusion3 or BoltzGen.
For each method and target, we run four rounds of 50 designs, matching the 200-design budget of \S\ref{sec:exp-ra-binder}.
Round 1 contains 50 i.i.d.\ designs generated as in \S\ref{sec:exp-ra-binder}.
Before each later round, the agent reads all scores so far and writes 50 new specifications (App.~\ref{app:iteropt}).
At budgets of 50, 100, 150, and 200 designs, we compare the accumulated iterative pool with matched-budget i.i.d.\ generation.

\begin{figure}[t]
\raggedright
\includegraphics[width=\linewidth]{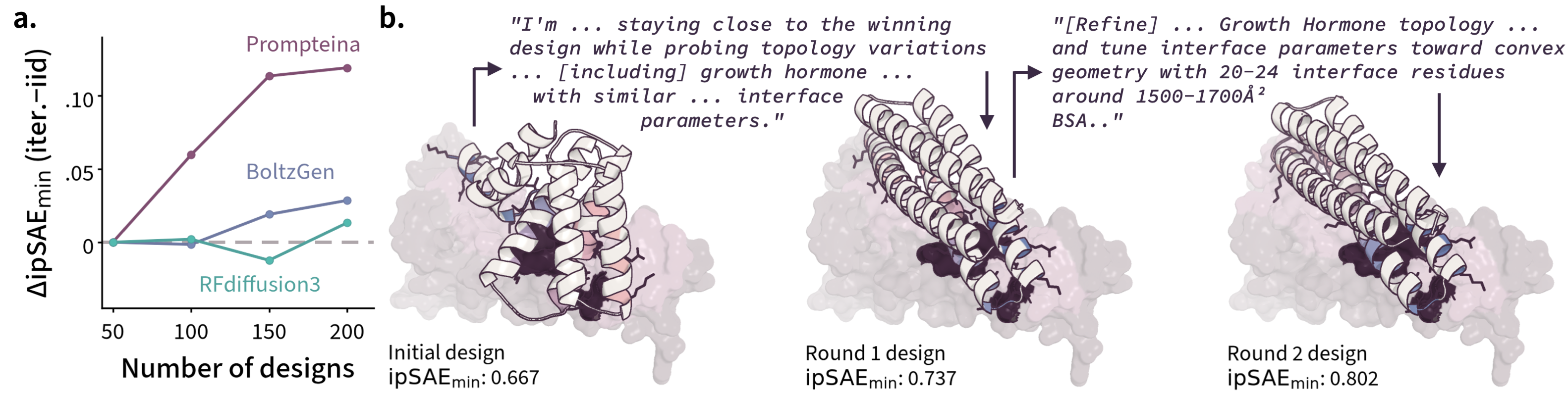}
\caption{\textbf{Prompteina gains most from iterative optimization on Claudin-1.}
\textbf{(a)} Difference in cumulative top-1 ipSAE$_\mathrm{min}$ between iterative and matched-budget i.i.d.\ generation, using the same initial pool of 50 designs.
\textbf{(b)} Planning excerpts and selected Claudin-1 designs from the first two feedback rounds.
The agent explores alternative topologies in round 1 and enlarges the requested interface in round 2.
For each complex, ipSAE$_\mathrm{min}$ ($\uparrow$) reports binder--target interface confidence.}
\label{fig:iteropt}
\end{figure}

\paragraph{Prompteina gains most from iterative optimization on the hardest targets.}
Across 20 targets, iterative feedback benefits Prompteina more than either baseline.
On the hardest targets, Prompteina has larger top-1 and top-8 gains than either baseline; its top-8 gain is 57\% larger than the next best (App.~\ref{app:iteropt-panel}; Table~\ref{tab:iteropt-panel}).
Using Claudin-1 as an example, we find that the agent first tests new topologies, then requests a larger interface, improving the best design after each round (Fig.~\ref{fig:iteropt}; App.~\ref{app:iteropt-claudin}).
Paired with either baseline, the agent also identifies productive patterns, but each model's interface offers fewer properties to revise (App.~\ref{app:iteropt-claudin}).
This suggests that Prompteina's broader natural-language conditioning lets the agent make more useful changes in response to in-silico feedback.

\subsection{Experimental Validation of GM2A Binder Designs}
\label{sec:exp-gm2a}

As a proof of concept, we conducted a prospective, agent-guided campaign against GM2 ganglioside activator protein (GM2A), implicated in GM2 gangliosidosis \citep{deschenesBiochemicalCorrectionGM22023}.
GM2A has no reported \textit{de novo} binders and yielded none in a prior 30-design campaign \citep{starkBoltzGenUniversalBinder2026}.

\begin{figure}[b]
\centering
\includegraphics[width=\linewidth]{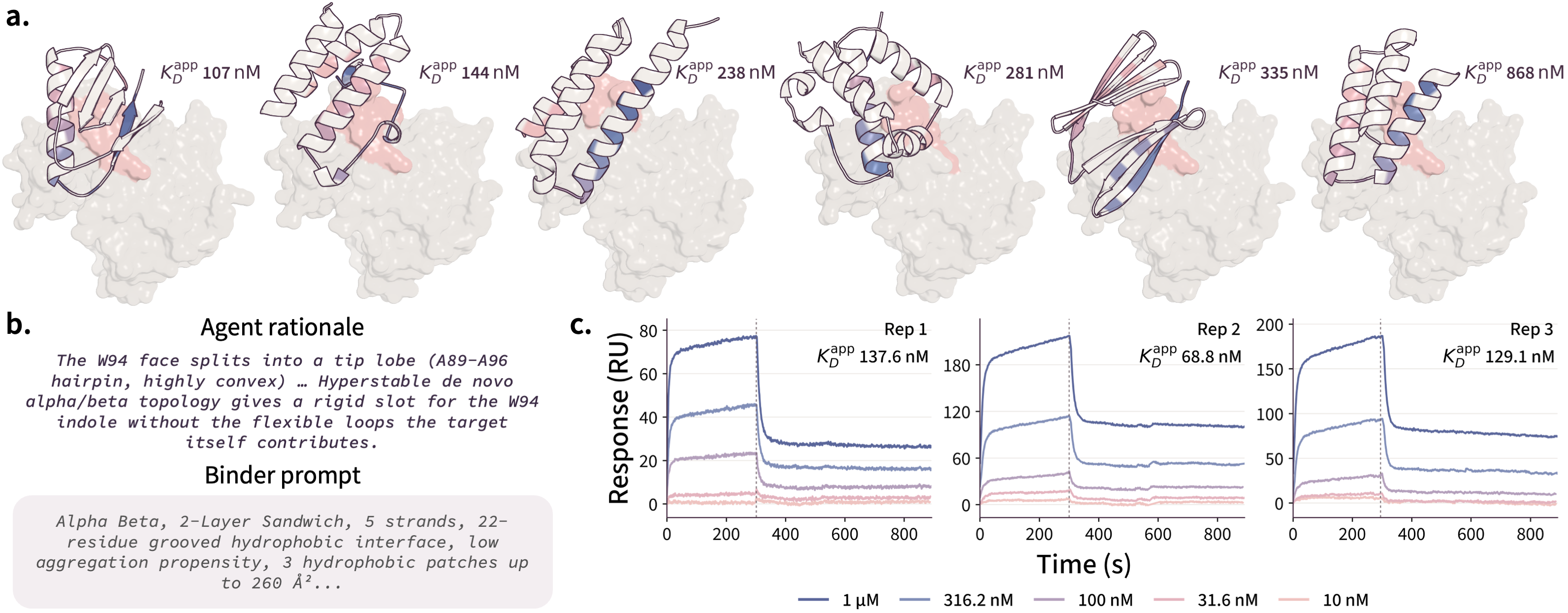}
\caption{\textbf{Experimental validation identifies diverse GM2A binders.}
\textbf{(a)} ESMFold2-Fast predictions of six sub-micromolar binders, ordered by apparent $K_{\mathrm{D}}$.
Despite distinct agent-specified architectures, all six target the same GM2A face around W94 (GM2A, gray; W94, pink).
\textbf{(b)} Agent reasoning and the design prompt for the strongest binder.
\textbf{(c)} SPR sensorgrams for this binder across three replicates and five analyte concentrations, labeled with per-replicate apparent $K_{\mathrm{D}}$ values.}
\label{fig:gm2a-campaign}
\end{figure}

We ran iterative optimization (\S\ref{sec:exp-iteropt}) with a 1,000-design budget, then selected 90 designs for experimental testing by Adaptyv Bio (App.~\ref{app:gm2a-assay}).
Of these, 89 were expressed and assayed by SPR (App.~\ref{app:gm2a-binding-results}).
Adaptyv classified nine binders, eight with fitted apparent $K_{\mathrm{D}}$ values.
Six sub-micromolar binders spanned diverse architectures, corresponding to a 6.7\% assay hit rate (Fig.~\ref{fig:gm2a-campaign}a).
The strongest binder had an overall apparent $K_{\mathrm{D}}$ of 107~nM across three replicates (Fig.~\ref{fig:gm2a-campaign}a,c).
Prompteina realized the agent's two-layer alpha/beta architecture around GM2A residue W94, with five strands, a hydrophobic interface, and 21 interface residues versus 22 requested (Fig.~\ref{fig:gm2a-campaign}b).

\section{Related Work}
\label{sec:related}
\paragraph{Protein structure generation.}
Early diffusion and flow-matching models generated protein backbones \citep{yimSE3DiffusionModel2023a,watsonNovoDesignProtein2023a,geffnerProteinaScalingFlowbased2025a}, while newer all-atom models jointly generate structure and sequence \citep{geffnerLaProteinaAtomisticProtein2026,didiScalingAtomisticProtein2026,butcherNovoDesignAllatom2025b,rector-brooksGeneralMultimodalProtein2026a,starkBoltzGenUniversalBinder2026}.
Existing protein structure generation models use custom input representations to specify desired folds, motif coordinates, target hotspots, and interactions, and some also require separate checkpoints or inference-time classifiers and rewards \citep{watsonNovoDesignProtein2023a,didiScalingAtomisticProtein2026,rector-brooksGeneralMultimodalProtein2026a,ingrahamIlluminatingProteinSpace2023a}.
RFdiffusion3 and BoltzGen combine many controls in one model, but each property has a distinct model input and its own training annotations \citep{butcherNovoDesignAllatom2025b,starkBoltzGenUniversalBinder2026}.
ESM3 jointly generates sequence, structure, and function through dedicated token tracks \citep{hayesSimulating500Million2025a}.
Chroma provides the closest precedent for using natural language to control protein structure generation.
It guides sampling through a separate structure-to-caption model rather than conditioning its denoiser on text \citep{ingrahamIlluminatingProteinSpace2023a}.
This approach requires gradients during sampling and does not use supervised text--residue alignment.
The authors report that its natural-language conditioning remains sensitive to guidance strength.
Prompteina instead directly conditions the denoising process on text, enabling a single natural-language interface to specify global and residue-specific properties.

\paragraph{Protein--text multimodal models.}
Protein--text multimodal models support alignment, captioning, and text-guided generation.
Alignment models learn shared representations for retrieval and classification \citep{xuProtSTMultiModalityLearning2023a,suTrimodalProteinLanguage2025a,xuMultilingualTranslationZeroshot2023a}, while captioning models describe protein sequences or structures \citep{liuProtT3ProteintoTextGeneration2024,feiProt2TextV2ProteinFunction2025a,abdineProt2TextMultimodalProteins2024a}.
Neither generates protein sequences or structures from text.
Text-guided protein models such as ProteinDT, Pinal, InstructPro, InstructBioMol, and BioMatrix produce amino-acid sequences rather than structures \citep{liuTextguidedProteinDesign2025b,daiNovoProteinDesign2025,songInstructProNaturalLanguage2026,zhuangAdvancingBiomolecularUnderstanding2025,peiBioMatrixComprehensiveBiological2026}.
Their prompts usually provide high-level functional descriptions, and the resulting sequences must then be folded into structures.
ProteinDT uses a text-aligned latent representation, while Pinal predicts Foldseek structure tokens before generating the sequence.
HelixProtX emits backbone torsion angles from text, but its pretrained parameters are unavailable for evaluation \citep{chenUnifyingSequencesStructures2024b}.
Prompteina instead directly conditions the generation of protein sequence and all-atom structure on text, with supervised text--residue alignment connecting phrases to the residues they describe.

We provide an extended survey of related work in Appendix~\ref{app:conditioning}, including conditioning interfaces supported by protein structure generation models and their property-specific representations.

\section{Discussion}
We introduce \textbf{Prompteina}, a text-conditioned all-atom protein structure generation model, and its paired text--structure dataset, \textbf{Prompteina-Data}.
Prompteina uses a single natural-language interface to control global and residue-specific structural and functional properties.
Prompteina matches or exceeds existing approaches on shared properties while supporting a broader range.
Its natural-language interface lets an agent turn research directly into design prompts, yielding higher interface confidence than the same agent achieves with existing models.
Experimental validation then identified Prompteina binders for GM2A, a target with no previously reported \textit{de novo} binders.
As language-model agents improve, specialist generative models need interfaces that let agents use scientific knowledge and language-based reasoning to guide generation.
Prompteina makes this connection more seamless: agents can incorporate experimental feedback into rich specifications realized directly in generated structures.
Together, these results establish natural language as an effective, powerful conditioning interface that lets agents directly connect scientific knowledge, language-based reasoning, and experimental results to protein structure generation and design.

\subsubsection*{Author Contributions}
Y.J.L., B.Q., and J.L. conceptualized the project. J.L. led the project, implemented the methods, conducted experiments, and wrote the manuscript. Y.J.L., B.Q., and R.D. contributed to method implementation. G.L. contributed to project discussions. R.D. supervised the project and provided compute and API resources through Ai2. A.E. and N.A.S. provided funding through Ai2. D.B., H.H., and P.W.K. supervised the project and provided funding. P.W.K., Y.J.L., B.Q., R.D., G.L., H.H., D.B., N.A.S., and A.E. reviewed or edited the manuscript.

\subsubsection*{Acknowledgments}
We thank Molly Park for providing Anthropic credits used in our agentic binder design campaigns, Bingxu Liu for screening structures for our experimental campaign, Chang Ma for proofreading, and Kieran Didi, Jasper Butcher, Jie Chen, Jeremy Koob, Scott Geng, Jina Kim, and Maya Venkatraman for helpful discussions.
We also thank members of the Koh Lab, H2Lab, Ai2 Open Ecosystem team, DSO National Laboratories, and GLOW.AI for their feedback on this work, and Madeline May, Jillian Willis, and Taira Anderson for their executive support.
This work was supported by the Singapore National Research Foundation and the National AI Group in the Singapore Ministry of Digital Development and Information through the AI Visiting Professorship Programme (award AIVP-2024-001), by the AI2050 program at Schmidt Sciences, and by the National Science Foundation under Award No.~2413244.

\bibliography{references,manual-refs}
\bibliographystyle{iclr2027_conference}

\appendix
\addtocontents{toc}{\protect\setcounter{tocdepth}{2}}
\renewcommand{\contentsname}{Appendix Contents}
{\small\tableofcontents}

\section{Extended related work}
\label{app:conditioning}

Recent protein structure generation models control different properties through distinct architectural components, training procedures, and inference-time operations.
We survey these mechanisms before describing the text-conditioning techniques that Prompteina adopts from image generation.
The comparison focuses on the dedicated input formats and procedures that existing models require for each controlled property.

\paragraph{RFdiffusion \citep{watsonNovoDesignProtein2023a}.} RFdiffusion fine-tunes a pretrained RoseTTAFold structure prediction network as a denoiser for the reverse diffusion process.
Five separately fine-tuned variants handle unconditional generation, fold-conditioned generation, binder design, fold-conditioned binder design, and enzyme active-site scaffolding.
Fold control takes per-residue secondary-structure labels and a residue-by-residue block adjacency matrix that specifies long-range contacts.
Binder design takes a user-specified subset of target-chain residues as interface hotspots and can combine them with fold properties to control binder scaffold topology.
Motif scaffolding fixes the sequence identity and three-dimensional coordinates of motif backbone and side-chain atoms during denoising.
Symmetric oligomer design takes a cyclic, dihedral, tetrahedral, octahedral, or icosahedral point group and initializes noise with the corresponding symmetry.
Each task therefore uses a distinct input format and separately trained model.

\paragraph{RFdiffusion2 \citep{ahernAtomlevelEnzymeActive2026}.} RFdiffusion2 uses the RoseTTAFold All-Atom architecture and combines Euclidean flow matching for atom coordinates with Riemannian flow matching on $\mathrm{SE}(3)$ for backbone frames.
The model is trained from random initialization and focuses on enzyme active-site scaffolding.
Atomic motif conditioning supplies individual heavy atoms from catalytic side chains.
During training, a random residue subset forms the motif, and each selected residue is represented either as a backbone frame or as an atomized subgraph of its heavy atoms.
Unindexed motif conditioning removes sequence indices from these residues, requiring the model to infer their positions in the scaffold.
Relative accessible surface area (RASA) labels mark each ligand atom as buried or exposed during 50\% of training.
An origin pseudo-atom gives an approximate center-of-mass target for the scaffold relative to the ligand.
Stochastic centering prevents the model from recovering the motif--scaffold offset directly from partially noised inputs by adding a small Gaussian translation to the training structure.
A single model handles these tasks, but each property has a dedicated input format and training procedure.

\paragraph{RFdiffusion3 \citep{butcherNovoDesignAllatom2025b}.} RFdiffusion3 uses a transformer-based U-Net for all-atom protein generation.
A single model handles unconditional generation, protein, nucleic-acid, and small-molecule binder design, enzyme active-site scaffolding, and symmetric oligomer design through several conditioning mechanisms.
Atom-level features label target or ligand atoms with property-specific annotations at varying rates during training.
Donor and acceptor labels direct complementary hydrogen bonds, RASA labels divide ligand-atom solvent accessibility into four burial bins, and hotspots mark target atoms near the requested binder--target interface.
Appended tokens represent catalytic residues for unindexed motif scaffolding, leaving the model to infer each catalytic residue's sequence position.
Inference-time operations symmetrize each diffusion step for oligomer design and position the generated protein relative to the target center of mass.
For nucleic-acid and small-molecule targets, RFdiffusion3 can jointly diffuse the target conformation with the protein.
The shared model still requires a distinct input format and annotation pipeline for each property.

\paragraph{Proteina series \citep{geffnerProteinaScalingFlowbased2025a, geffnerLaProteinaAtomisticProtein2026, didiScalingAtomisticProtein2026}.} Proteina \citep{geffnerProteinaScalingFlowbased2025a} is a non-equivariant flow-matching model for protein backbone generation.
CATH fold conditioning uses learnable embeddings for a hierarchy of 5 classes, 43 architectures, and 1,336 topologies, injected through adaptive layer normalization (AdaLN).
Hierarchical dropout trains unconditional generation and each level of the hierarchy.
The model drops all labels with probability $0.5$, retains only class with probability $0.1$, retains class and architecture with probability $0.15$, and retains all three levels with probability $0.25$.
La-Proteina \citep{geffnerLaProteinaAtomisticProtein2026} extends Proteina to all-atom generation through a partially latent representation and adds indexed and unindexed motif scaffolding, but omits fold conditioning.
Complexa \citep{didiScalingAtomisticProtein2026} extends La-Proteina to binder design with separate protein--protein, CATH-conditioned protein--protein, and ligand--protein checkpoints.
Protein and small-molecule targets use different representation pipelines.
Binder design takes target interface hotspots, and the CATH-conditioned checkpoint uses Proteina's hierarchical embeddings and AdaLN but applies only the class level at inference.

\paragraph{Chroma \citep{ingrahamIlluminatingProteinSpace2023a}.} Chroma composes a library of separate conditioners at sampling time.
Each independently trained or hand-designed conditioner steers generation through its gradient with respect to the noised structure.
Symmetry conditioning tessellates an asymmetric subunit into C$_n$, D$_n$, T, O, or I point groups, and shape conditioning minimizes an optimal-transport distance to a user-provided point cloud.
Fold-class conditioning inverts a separately trained CATH classifier through its gradients.
Substructure conditioning fixes motif coordinates during registered or unregistered scaffolding.
Text conditioning trains a structure-to-caption predictor and uses the gradient of caption log-likelihood to steer generation toward the prompt.
Guidance strength must be balanced against the prior: weak gradients have little effect, whereas strong gradients overwhelm the prior and degrade structural coherence.
The authors report that guidance effects vary widely and that robust natural-language conditioning remains challenging.
The ProCap captioner pairs a frozen structure encoder with GPT-Neo 125M and is trained on approximately 45{,}000 captions from PDB descriptions and UniProt functional comments.
The authors report that this dataset size constrained their architecture choices.
Any subset of conditioners can be composed at inference, but each requires a trained classifier, analytical energy function, or separate gradient computation.

\paragraph{HelixProtX \citep{chenUnifyingSequencesStructures2024b}.} HelixProtX is a multimodal language model that maps among protein sequence, structure, and text.
Generating structure from text emits backbone torsion angles, and control is reported at the level of the whole structure rather than individual residues or motifs.
Its released repository does not include the pretrained HelixProtX parameters, which prevents evaluation on our benchmark.\footnote{\href{https://github.com/PaddlePaddle/PaddleHelix/tree/dev/apps/helixprotx}{HelixProtX repository}.}

\paragraph{Pinal \citep{daiNovoProteinDesign2025}.} Pinal generates amino-acid sequences from prompts that describe a protein's function and specify its length.
It first converts the prompt into Foldseek structure tokens, which encode the local backbone geometry around each residue.
A structure-aware protein language model then predicts an amino acid at each position from these tokens and the prompt.
Pinal returns a sequence, which we fold with ESMFold to obtain coordinates for structural evaluation.

\paragraph{BioMatrix \citep{peiBioMatrixComprehensiveBiological2026}.} BioMatrix represents natural language, amino-acid sequences, protein backbones, and small molecules as token sequences processed by one decoder-only model.
Instructions specify which representation the model receives and which representation it generates for a given task.
For the text-guided protein design task evaluated here, BioMatrix receives a functional description and generates an amino-acid sequence.
BioMatrix has no length input, so we constrain decoding to the sampled length and fold each sequence with ESMFold for structural evaluation.

\paragraph{DISCO \citep{rector-brooksGeneralMultimodalProtein2026a}.} DISCO jointly generates protein sequences and structures with Euclidean coordinate diffusion and absorbing-state masked sequence diffusion.
Biomolecular context enters as discrete tokens with three-dimensional coordinates for small molecules, metallocofactors, reactive intermediates, or nucleic acids, which are denoised with the protein.
Property-specific control uses Feynman--Kac reward tilting, with a custom reward function, distance thresholds, and hyperparameters for properties such as disulfides or cation--$\pi$ interactions.
Target-specific binding guidance evaluates two aligned denoiser calls per timestep and requires a hyperparameter sweep for each target pair.

\paragraph{PAR \citep{quProteinAutoregressiveModeling2026a}.} PAR uses an autoregressive model and flow-matching decoder to generate protein backbones by progressively upsampling a coarse representation.
Its two conditioning mechanisms operate on this hierarchy without fine-tuning.
Structural prompting replaces the coarsest generation step with user-provided three-dimensional points that define an approximate layout, which PAR then upsamples into a full backbone.
Motif scaffolding restores the ground-truth coordinates of motif residues after every prediction scale, keeping the motif fixed as the surrounding scaffold is generated.

\paragraph{BoltzGen \citep{starkBoltzGenUniversalBinder2026}.} BoltzGen extends Boltz-2 into a single model for folding, unconditional design, binder design, motif scaffolding, and non-protein interface design through seven training tasks.
Four conditioning mechanisms use separate representations and training distributions.
Binding-site conditioning assigns binding, non-binding, or unspecified labels to target residues.
Structure groups use a distance matrix to preserve pairwise distances among residues in the same group.
Secondary-structure conditioning labels each designed residue as helix, strand, coil, or unspecified.
Covalent-bond conditioning supplies an atom-pair bond matrix for cyclic proteins, disulfide staples, and helicons.
Each property has its own representation, dropout scheme, and training-time sampling distribution.

\paragraph{ESM3 \citep{hayesSimulating500Million2025a}.} ESM3 is a multimodal masked language model that generates protein sequence, structure, and function, with each modality or property represented by a dedicated token track.
For conditioning structure generation, partial information can be provided on any combination of property tracks: per-residue secondary structure labels, solvent-accessible surface area bins, function keywords derived from InterPro and Gene Ontology terms, residue-level functional annotations, partial structure (as discrete tokens or atomic coordinates), and partial sequence.
Each track requires its own tokenization: structure uses a VQ-VAE with a 4096-entry codebook, function keywords require parsing free text into a 68{,}103-entry vocabulary, quantizing via 8 locality-sensitive hash functions, and a dedicated 3-layer transformer to invert the hashing.
Each track also has its own noise schedule and dropout rate.
Adding a new conditioning property requires designing a tokenization scheme, selecting a noise schedule, and training any auxiliary encoders or decoders the representation needs.
Chroma and RFdiffusion already represent ESM3's per-residue secondary-structure input in the property-specific comparison, so we omit a redundant ESM3 run.

\paragraph{Genie series \citep{linGeneratingNovelDesignable2023a, linOutManyOne2024a, linFastUltraCapableProtein2026a}.} Genie \citep{linGeneratingNovelDesignable2023a} generates protein backbones by diffusing $\mathrm{C}_\alpha$ coordinates and conditions only on protein length.
Genie~2 \citep{linOutManyOne2024a} introduces motif scaffolding by training entirely on scaffolding tasks.
It represents motif sequence as one-hot vectors in the single-residue features and motif structure as $\mathrm{SE}(3)$-invariant pairwise $\mathrm{C}_\alpha$ distances.
This representation leaves Genie~2 to infer the relative configuration of multiple motifs.
Genie~3 \citep{linFastUltraCapableProtein2026a} uses a branched-polymer representation for all-atom generation and adds binder design.
Its training distribution allocates 90\% of examples to motif scaffolding and 10\% to binder design, with subtasks that vary the motif and binder information supplied.
Binder design takes target interface hotspots and can expand them to nearby surface residues or iteratively add residues from successful prior designs.

\paragraph{Text conditioning for diffusion models.} Protein structure generation models increasingly use transformer backbones also found in text-conditioned image models \citep{geffnerProteinaScalingFlowbased2025a, didiScalingAtomisticProtein2026, butcherNovoDesignAllatom2025b, rector-brooksGeneralMultimodalProtein2026a, starkBoltzGenUniversalBinder2026}.
That literature supplies the components Prompteina uses for natural-language conditioning.
Frozen pretrained language models encode the text \citep{sahariaPhotorealisticTexttoImageDiffusion2022b}, and later systems combine encoders to capture complementary information \citep{podellSDXLImprovingLatent2023, esserScalingRectifiedFlow2024a}.
Cross-attention carries token representations into the denoiser \citep{rombachHighResolutionImageSynthesis2022a}, and adaptive layer normalization supplies a pooled representation \citep{peeblesScalableDiffusionModels2023a}.
PixArt-$\alpha$ combines these pathways and first trains the text projection while freezing the pretrained generation model \citep{chenPixArtaFastTraining2023}.
Classifier-free guidance trains conditional and unconditional predictions by dropping the conditioning on a fixed fraction of examples, then scales their difference at inference \citep{hoClassifierFreeDiffusionGuidance2022a}.
Prompteina adapts these components to protein structure generation.

\FloatBarrier
\section{Prompteina-Data construction}
\subsection{Data sources and scale}
\label{app:data-sources}

We construct the paired text--structure dataset from three sources.
Monomer training uses experimental protein chains from the Protein Data Bank (PDB) and predicted monomers from the AlphaFold Database (AFDB).
Binder training uses Teddymer dimers assembled from TED domain pairs \citep{didiScalingAtomisticProtein2026}.
The three corpora provide 1.8M entries and 52.4M captions (Table~\ref{tab:data-scale}).

\begin{table}[H]
\caption{\textbf{Training data composition.}
Entries count independently captioned protein chains.}
\label{tab:data-scale}
\centering
\small
\begin{tabular}{lrrr}
\toprule
\textbf{Corpus} & Entries & Captions & Properties \\
\midrule
PDB chains & 611K & 19.5M & 59 \\
AFDB monomers & 288.7K & 7.5M & 26 \\
Teddymer chains & 903K & 25.4M & 32 \\
\midrule
Total & 1.8M & 52.4M & -- \\
\bottomrule
\end{tabular}
\end{table}

PDB chains include all protein chains (monomer and multimer) with no length or resolution filter at the captioning stage.
At training time, chains are filtered to remove nucleic acid leaks, C$_\alpha$-only entries, entries with fewer than 33 resolved residues, and chains longer than 512 residues.
AFDB monomers are filtered by pLDDT confidence ($>$70) and maximum length ($<$512 residues), matching the base model training data.
Teddymer dimers are synthetic complexes constructed by pairing TED-domain structures \citep{didiScalingAtomisticProtein2026}.
We caption each chain independently, producing two caption sets per dimer.
During binder training, each binder crop is conditioned on the caption for the chain selected as the binder.

\subsection{Annotation extraction and categories}
\label{app:fact-extraction}

Before captioning, we extract chain- or complex-level annotations from source databases and compute additional properties from atomic coordinates.
The extraction produces 59 property types for PDB, 26 for AFDB, and 32 for Teddymer, organized into ten groups.
Unless noted, a property appears in all three corpora.
For AFDB, we use TED-derived CATH fold labels in place of experimental CATH assignments and exclude properties requiring experimental ligands, biological assemblies, or ECOD domain annotations.
Teddymer is captioned per chain and excludes ligand, sheet topology, and PDB-specific properties.
\begin{itemize}[nosep,leftmargin=*]
\item Fold and topology (11): CATH \citep{sillitoeCATHComprehensiveStructural2015} class, architecture, and topology; ECOD \citep{chengECODEvolutionaryClassification2014} architecture, X-group, H-group, and T-group (PDB); contact order; internal symmetry detected via CE-Symm \citep{myers-turnbullSystematicDetectionInternal2014} (PDB); sheet topology including $\beta$-sandwich detection via SVD sheet-normal comparison (PDB, AFDB); knot topology from KnotProt \citep{dabrowski-tumanskiKnotProt20Database2019} (PDB).
\item Secondary structure (3): helix count with an $\alpha$, $3_{10}$, and $\pi$ breakdown, strand count, and sheet count (PDB, AFDB).
We assign secondary structure with mkdssp \citep{hekkelmanDSSP4FAIR2025}, merging helix gaps with \texttt{max\_gap}$=2$ to tolerate kinks \citep{wilmanHelixKinksAre2014}.
\item Tertiary structural motifs (7): disulfide bonds with sequential separation class (local: $<$30 residues, medium-range: 30--80, long-range: $>$80) and disulfide topology (e.g., cystine knot, ladder, nested); metal coordination with geometry and ligand composition (e.g., tetrahedral zinc with 3 His, 1 Cys) (PDB); intra-chain salt bridges classified as buried or surface by relative solvent accessibility (PDB, AFDB); hydrogen-bond networks (PDB, AFDB); cis-peptides; $\beta$-turns.
\item Global shape and surface (9): compactness with shape classification from SVD on C$_\alpha$ coordinates (spherical: elongation $<$1.5, elongated: 1.5--3.0, rod-like: $>$3.0, where elongation is the ratio of the two largest singular values); surface planarity; spatial aggregation propensity; buried unsatisfied hydrogen bonds; flexibility; hinge residues; repeat detection with secondary-structure composition; mechanical clamp type; Ramachandran profile.
\item Quaternary structure (3): biological assembly symmetry (PDB), subunit structure (PDB), oligomeric context (PDB, Teddymer).
\item Protein--protein interfaces (9; PDB, Teddymer): interface residues, buried surface area, hydrogen bonds (backbone/sidechain split), salt bridges, hydrophobic contacts, interface geometry, shape complementarity, secondary-structure character, calcium-dependent interfaces.
\item Ligand binding (3): ligand identity with molecular properties computed from Chemical Component Dictionary SMILES via RDKit (heavy atom count, rotatable bonds, hydrogen-bond acceptors and donors, octanol-water partition coefficient, ligand type classification) (PDB), ligand contacts with pocket secondary-structure composition and pocket flexibility from B-factors (PDB), catalytic activities.
\item Domain architecture (5): multi-domain layout, linker regions, domain--domain interfaces, UniProt \citep{theuniprotconsortiumUniProtUniversalProtein2025a} domain regions remapped to PDB coordinates via SIFTS \citep{danaSIFTSUpdatedStructure2019} (all PDB); chain layout.
\item PDB-specific annotations (4; PDB only): membrane topology from PDBTM \citep{kozmaPDBTMProteinData2013}, tunnel geometry from ChannelsDB \citep{pravdaChannelsDBDatabaseBiomacromolecular2018}, curated repeats from RepeatsDB \citep{paladinRepeatsDB2021Improved2021}, antibody CDR regions from SAbDab \citep{dunbarSAbDabStructuralAntibody2014}.
\item Function and annotation (5): GO \citep{thegeneontologyconsortiumGeneOntologyResource2021} molecular function terms filtered to experimental evidence (PDB), subcellular locations from UniProt comments (PDB), post-translational modifications remapped via SIFTS (PDB), structure-specific statements mined from PDB-linked scientific literature (PDB), free-text function.
\end{itemize}

For structure-specific literature statements in the function and annotation group, we retrieve PDB-linked PubMed abstracts and, when available, Results and Discussion sections from PubMed Central.
Qwen3-235B-A22B extracts these statements from abstracts, while Qwen3-32B extracts and validates them from full papers \citep{yangQwen3TechnicalReport2025a}.
Rule-based filters remove cross-structure attributions, generic or methodological statements, residue-specific details, and semantic duplicates.

\subsection{Residue-specific captions}
\label{app:residue-level-captions}

Most captions describe properties without identifying the participating residues, as in \texttt{three disulfide bonds}.
For 15 properties, captions can also include positions in 1-based SEQRES numbering, as in \texttt{disulfide between Cys8 and Cys38}.
We sample independently for each property in each caption whether to include these positions.
Properties whose individual positions define the requested structure, including metal coordination, catalytic activity, and disulfides, receive positions more often ($p{=}0.35$--$0.40$).
Interaction properties such as interface hydrogen bonds, salt bridges, and hydrophobic contacts use lower probabilities ($p{=}0.10$--$0.20$).
We omit positions for emergent properties such as flexibility, hinge residues, and $\beta$-turns because individual residues are not independently designable in these cases.

Residue identifiers use standard structural biology notation, such as \texttt{Cys8} and \texttt{Arg118}.
Cross-chain interactions add a \texttt{target:} prefix to partner-chain residues, as in \texttt{Arg45-target:Asp112}.
Captions that include positions omit qualitative count terms such as \texttt{several} and \texttt{many}, and long position lists are truncated to a fixed number of entries.
The training distribution therefore mixes exact positional prompts with position-free descriptions.

\subsection{Caption sampling}
\label{app:caption-sampling}

Including every available property in every caption would make the descriptions repetitive and dense.
We instead generate multiple captions per structure and vary the number of properties included in each one.
For a structure with $n$ available properties, we sample a continuous proportion $X \sim \mathrm{Beta}(1,\beta)$ and set
\[
K = \floor{nX} + 1,
\]
where $K$ is the number of properties included in the caption.
From the available set, we sample $K$ properties uniformly without replacement.
Setting $\beta=5.0$ favors sparse captions while retaining occasional dense descriptions.

The expected number of sampled properties is
\begin{equation}
\mathbb{E}[K] = \sum_{i=1}^{n} P(K \geq i) = \sum_{i=1}^{n} P\!\left(X \geq \tfrac{i-1}{n}\right) = \sum_{i=1}^{n} \left(1-\tfrac{i-1}{n}\right)^{\beta},
\end{equation}
where the last equality uses the survival function of $\mathrm{Beta}(1,\beta)$.
We choose the number of captions per structure by bounding the probability that any property is never sampled.
Let $q = 1 - \mathbb{E}[K]/n$ approximate the probability that one caption excludes a property, and let $E_i$ denote the event that property $i$ is absent from all $N$ captions.
The union bound gives
\begin{equation}
P(E_1 \cup \cdots \cup E_n) \leq n q^N.
\end{equation}
Setting this bound to $1-p$ gives $N = \left\lceil{(\log(1-p)-\log n)}/{\log q}\right\rceil$, with $p=0.95$.
The number of populated properties varies among entries because, for example, many PDB chains lack ligands or metal-coordination sites.
Applying the coverage criterion produces averages of approximately 32 captions per PDB entry, 26 per AFDB entry, and 28 per Teddymer chain.

\subsection{LLM caption generation}
\label{app:caption-prompt}

We generate captions with Qwen3-235B-A22B \citep{yangQwen3TechnicalReport2025a} served through vLLM.
The input is a structured list of sampled annotations, and the prompt instructs the model to produce a faithful caption without adding information beyond them (Box~\ref{box:caption-template}).
We sample with temperature $0.7$, top-$p$ $0.8$, top-$k$ $20$, and presence penalty $1.5$.
Caption generation runs locally.
The measured throughput of approximately 14{,}300 captions per H200 GPU-hour places the cost of generating all 52.4M captions at approximately 3{,}700 H200 GPU-hours.

\begin{promptbox}{Caption generation prompt and example output}{box:caption-template}
\definecolor{filled}{HTML}{8A5479}%
You are given a list of facts about a protein formatted as key-value pairs.
Generate a factual protein caption from the following facts.
Only state what is explicitly provided.
Do not speculate, infer, or elaborate beyond the given information.\\

Provide only the caption text.\\

Facts:\\
{\color{filled}- Structural class: Alpha Beta}\\
{\color{filled}- Helices: 4}\\
{\color{filled}- Strands: 5}\\
{\color{filled}- Oligomeric context: monomeric (C1)}\\
{\color{filled}- Repeat structure: no internal repeats detected}\\

Caption:\\
{\color{filled}The protein adopts an alpha-beta structural class, featuring a total of four helices and five strands.
It functions as a monomeric entity in its oligomeric context.}
\end{promptbox}

\paragraph{Validation.} We evaluate caption fidelity with Claude Opus 4.6 as an LLM judge on 80 captions.
The sample covers different properties, data sources, caption lengths, and the presence or absence of residue-specific detail.
For each caption, the judge compares every factual claim against the source annotations and evaluates three metrics:
\begin{itemize}[nosep,leftmargin=*]
\item Recall: fraction of input properties whose content appears in the generated caption.
\item Precision: fraction of distinct factual claims in the caption that are traceable to a source annotation. Mild editorial language (e.g., ``contributing to structural stability'') counts as traceable.
\item Accuracy: fraction of mentioned properties that are stated with the correct value.
\end{itemize}

\begin{table}[H]
\centering
\caption{\textbf{Caption validation.}
The evaluation covers 80 captions.}
\label{tab:caption-faithfulness}
\small
\begin{tabular}{lcc}
\toprule
Property & Claims & \% \\
\midrule
Recall & 584 / 597 & 97.8 \\
Precision & 804 / 895 & 89.8 \\
Accuracy & 774 / 809 & 95.7 \\
\bottomrule
\end{tabular}
\end{table}

Unsupported numerical specificity accounts for most of the caption judge's precision loss.
Qualitative annotations such as \texttt{high} shape complementarity or \texttt{extensive} buried surface area sometimes prompt the language model to add a plausible value such as \texttt{score 0.89} or \texttt{3700~\AA$^2$}.
A less frequent failure divides a total into unsupported components, such as surface and buried contributions to a salt-bridge count.
Despite these errors, the captions retain high recall, precision, and accuracy on the audited sample (Table~\ref{tab:caption-faithfulness}).

\subsection{Text--residue alignment annotations}
\label{app:alignment-annotation}

After generating captions, we map text spans to residue positions to produce supervised alignment masks.
The selected source annotations identify which property text to locate in each caption.
We find that text with regular-expression anchors and TF--IDF-weighted token matching against the formatted annotation.
Captions that include residue positions use direct matches to identifiers such as \texttt{Cys8} (\S\ref{app:residue-level-captions}).
For example, given the caption in Box~\ref{box:caption-template} and the annotation \texttt{Structural class: Alpha Beta}, the annotator computes TF-IDF weights for the tokens \texttt{structural}, \texttt{class}, \texttt{alpha}, and \texttt{beta}, scans the caption, and identifies \texttt{alpha-beta structural class} as the best-matching region.
We then map this span to the helix and strand residues within the CATH domain boundary.

The property type determines the residue set assigned to each matched text span.
Secondary-structure phrases map to helix, strand, or sheet residues, and CATH fold labels map to secondary-structure residues within the annotated domain boundary.
Disulfides, salt bridges, metal coordination, and interface interactions map to both participating residues as pairs.
Whole-chain properties such as contact order and compactness map to all modelled positions.
Database-level properties, including ECOD labels, quaternary structure, Gene Ontology terms, subcellular locations, and free-text function, receive no residue mask.
We omit an annotation from the mask when its text span cannot be located reliably.

\paragraph{Validation.} We manually label the character span corresponding to each selected property in 80 captions, split evenly between held-in and hold-out examples, for 617 spans in total.
The annotator finds spans for nearly every category in the 40 hold-out captions, with lower recall where connecting words fall between matched tokens (99\% coverage, 89\% character-level precision, and 73\% recall over 267 spans).
Examples of missed connecting text include \texttt{corresponding to} and \texttt{with a mean distance of}.

\FloatBarrier
\section{Model and training details}
\label{app:model-details}

Prompteina adds text conditioning to Complexa while retaining its partially latent flow-matching representation.
This appendix specifies the inherited notation, new conditioning components, training, and sampling; \citet{didiScalingAtomisticProtein2026} describe the full base architecture.

\subsection{Inherited Complexa model}
\label{app:base-model}

We initialize Prompteina from a Complexa checkpoint trained first on AFDB monomers and then on PDB and Teddymer complexes.
Complexa uses partially latent flow matching to generate backbone $\mathrm{C}_\alpha$ coordinates $\mathbf{x}^{\mathrm{C}_\alpha}\in\mathbb{R}^{N\times 3}$ and per-residue side-chain latents $\mathbf{z}\in\mathbb{R}^{N\times 8}$.
A frozen variational autoencoder $\mathcal{E}$ maps an atomistic structure $\mathbf{x}$ to these latents.
For a training structure $(\mathbf{x}_1^{\mathrm{C}_\alpha}, \mathbf{z}_1) \sim p_\mathrm{data}$ with $\mathbf{z}_1=\mathcal{E}(\mathbf{x})$, the model samples independent timesteps $t_x,t_z\in[0,1]$.
It constructs linear interpolants $\mathbf{x}_{t_x}^{\mathrm{C}_\alpha}=t_x\mathbf{x}_1^{\mathrm{C}_\alpha}+(1-t_x)\mathbf{x}_0^{\mathrm{C}_\alpha}$ and $\mathbf{z}_{t_z}=t_z\mathbf{z}_1+(1-t_z)\mathbf{z}_0$ from noise $\mathbf{x}_0^{\mathrm{C}_\alpha}\sim\mathcal{N}(\mathbf{0},\mathbf{I})$ and $\mathbf{z}_0\sim\mathcal{N}(\mathbf{0},\mathbf{I})$.
The denoiser $\mathbf{v}_\phi$ predicts the velocity field:
\begin{equation}
\label{eq:fm_loss}
\mathcal{L}_\mathrm{FM}
=
\mathbb{E}_{t_x,t_z}
\!\left[
\bigl\|\mathbf{v}_\phi^x-(\mathbf{x}_1^{\mathrm{C}_\alpha}-\mathbf{x}_0^{\mathrm{C}_\alpha})\bigr\|^2
+
\bigl\|\mathbf{v}_\phi^z-(\mathbf{z}_1-\mathbf{z}_0)\bigr\|^2
\right].
\end{equation}
Here $\mathbf{v}_\phi^x$ and $\mathbf{v}_\phi^z$ are evaluated at $(\mathbf{x}_{t_x}^{\mathrm{C}_\alpha},\mathbf{z}_{t_z},\mathbf{c},t_x,t_z)$.

The denoiser is a transformer with hidden width $d$ over $N$ residue states $\mathbf{h}\in\mathbb{R}^{N\times d}$.
Each layer applies adaptive layer normalization (AdaLN), pair-biased multi-head self-attention, and a feed-forward network:
\begin{align}
\label{eq:layer}
\mathbf{h} &\leftarrow \mathbf{h} + \mathrm{AdaScale}\bigl(\mathrm{MHSA}\bigl(\mathrm{AdaLN}(\mathbf{h},\mathbf{c})\bigr),\mathbf{c}\bigr), \nonumber\\
\mathbf{h} &\leftarrow \mathbf{h} + \mathrm{AdaScale}\bigl(\mathrm{FFN}\bigl(\mathrm{AdaLN}(\mathbf{h},\mathbf{c})\bigr),\mathbf{c}\bigr).
\end{align}
Let $d_c$ denote the width of the conditioning array $\mathbf{c}\in\mathbb{R}^{N\times d_c}$.
In Complexa, $\mathbf{c}$ contains only embeddings of $t_x$ and $t_z$.
Prompteina also includes the pooled text representation in $\mathbf{c}$.

\subsection{Text encoders}
\label{app:text-encoders}

\paragraph{Sentence encoder.} We use Qwen3-Embedding-8B through the \texttt{sentence\_transformers} library to produce L2-normalized, 4,096-dimensional embeddings.
Each caption receives the following task instruction before encoding:

\begin{promptbox}{Sentence encoder task instruction}{box:pooled-instruction}
Given a description of a protein, embed it based on semantic meaning only.
Ignore surface form variation such as casing, punctuation, paraphrasing or rewording.
\end{promptbox}

\noindent The instruction asks the frozen encoder to ignore stylistic variation in the generated captions.
The two-layer MLP $g_s$ projects the embedding with Linear $4096{\to}1024$, GELU, LayerNorm, and Linear $1024{\to}256$.

\paragraph{Token encoder.} We use Qwen3-8B with its default chat-template formatting.
We extract the hidden state from each of its 36 transformer layers, excluding the embedding layer, and average them for each token.
The two-layer MLP $g_t$ projects each token with Linear $4096{\to}1024$, GELU, LayerNorm, and Linear $1024{\to}768$.

We freeze both encoders throughout all training stages.

\subsection{Token cross-attention}
\label{app:cross-attn}

For a prompt with $T$ tokens, let $L$ denote the number of language-model layers and $d_\mathrm{LM}$ their hidden width.
Let $\mathbf{e}^{(l)}_i\in\mathbb{R}^{d_\mathrm{LM}}$ denote the representation of token $i$ at language-model layer $l$.
We aggregate the $L$ representations for each token with a function $f$:
\[
\tilde{\mathbf{c}}^\mathrm{t}_i = f\!\bigl(\mathbf{e}^{(1)}_i,\ldots,\mathbf{e}^{(L)}_i\bigr) \in\mathbb{R}^{d_\mathrm{LM}}, \quad i=1,\ldots,T.
\]
The best-performing choice in the ablation is an average across layers (App.~\ref{app:ablations}).
The learned projector gives $\mathbf{c}^\mathrm{t}_i=g_t(\tilde{\mathbf{c}}^\mathrm{t}_i)\in\mathbb{R}^{d}$.

The function $\mathrm{CrossAttn}(\mathbf{h},\mathbf{c}^\mathrm{t},\mathbf{c})$ denotes cross-attention from the protein states $\mathbf{h}$ to the token embeddings $\mathbf{c}^\mathrm{t}$, with $\mathbf{c}$ modulating the protein queries.
Let $\alpha_\ell$ be the learned scalar gate for denoiser layer $\ell$.
Each layer applies the following update before protein self-attention:
\begin{equation}
\label{eq:cross_attn}
\mathbf{h} \leftarrow \mathbf{h} + \tanh(\alpha_\ell)\,
\mathrm{CrossAttn}(\mathbf{h},\mathbf{c}^\mathrm{t},\mathbf{c}).
\end{equation}
Let $d_h$ denote the attention-head width.
Let $\mathrm{LN}$ denote layer normalization, and let $\mathbf{W}_Q$, $\mathbf{W}_K$, and $\mathbf{W}_V$ denote the learned query, key, and value projections.
The projected queries, keys, and values are $\mathbf{Q}$, $\mathbf{K}$, and $\mathbf{V}$.
Let $\mathrm{Attn}$ denote scaled dot-product attention.
Protein states form the queries, and text tokens form the keys and values:
\begin{gather}
\mathrm{CrossAttn}(\mathbf{h},\mathbf{c}^\mathrm{t},\mathbf{c}) = \mathrm{Attn}(\mathbf{Q},\mathbf{K},\mathbf{V}), \nonumber\\
\mathbf{Q} = \mathrm{AdaLN}(\mathbf{h},\mathbf{c})\mathbf{W}_Q\in\mathbb{R}^{N\times d_h}, \quad
\mathbf{K} = \mathrm{LN}(\mathbf{c}^\mathrm{t})\mathbf{W}_K\in\mathbb{R}^{T\times d_h}, \quad
\mathbf{V} = \mathrm{LN}(\mathbf{c}^\mathrm{t})\mathbf{W}_V\in\mathbb{R}^{T\times d_h}.
\end{gather}
An output projection $\mathbf{W}_O$ maps the attention output back to dimension $d$.
We initialize $\alpha_\ell$ to zero.
The conditioning array $\mathbf{c}$ enters through AdaLN on the queries, so the pooled text representation and denoising timestep modulate the token--protein interaction.

\subsection{Text--residue alignment head}
\label{app:alignment-head}

Let $\Delta\mathbf{h}_\ell=\tanh(\alpha_\ell)\mathrm{CrossAttn}(\mathbf{h},\mathbf{c}^\mathrm{t},\mathbf{c})\in\mathbb{R}^{N\times d}$ denote the gated cross-attention update at denoiser layer $\ell$.
Let $L_d$ denote the number of denoiser layers.
We normalize learned weight logits $\mathbf{w}$ as $\boldsymbol{\beta}=\mathrm{softmax}(\mathbf{w})$ and aggregate the updates as $\bar{\mathbf{h}}$:
\[
\bar{\mathbf{h}} = \textstyle\sum_\ell \beta_\ell\,\Delta\mathbf{h}_\ell \in\mathbb{R}^{N\times d}.
\]
A zero-initialized projection $\mathbf{W}_a\in\mathbb{R}^{d\times d}$ maps $\bar{\mathbf{h}}$ into a shared space with the token embeddings.
Their dot product gives the alignment logits:
\[
A_{n,i} = \langle \mathbf{W}_a\bar{\mathbf{h}}_n,\mathbf{c}^\mathrm{t}_i\rangle.
\]
We train these logits with Equation~\ref{eq:align_loss}.
Aligned residue--token pairs are sparse, so positive entries are upweighted.
Let $n_\mathrm{pos}$ and $n_\mathrm{neg}$ denote the number of positive and negative entries in one sample.
Samples without valid alignment annotations receive zero alignment loss.
Otherwise, the per-sample positive weight is $w_+ = n_\mathrm{neg}/n_\mathrm{pos}$.

\subsection{Initialization and gating}
\label{app:init-gating}

\paragraph{AdaLN and AdaScale.} The base model's AdaLN computes $\gamma(\mathbf{c})\odot\mathrm{LayerNorm}(\mathbf{h})+\beta(\mathbf{c})$, where $\gamma=\sigma(\mathbf{W}_\gamma\mathbf{c}+\mathbf{b}_\gamma)$ and $\beta=\mathbf{W}_\beta\mathbf{c}$, with default initialization.
AdaScale computes $\sigma(\mathbf{W}_s\mathbf{c}+\mathbf{b}_s)\odot\mathbf{h}$.
Its weights start at zero and its bias at $-2.0$, so $\sigma(-2.0)\approx 0.12$ and each layer's initial output contribution is small.

\paragraph{Cross-attention gating.} We gate the cross-attention output with a per-layer scalar $\alpha_\ell$ initialized to zero:
\[
\Delta\mathbf{h}_\ell = \tanh(\alpha_\ell)\cdot\mathrm{Attn}(\mathbf{Q},\mathbf{K},\mathbf{V}).
\]
The zero initialization removes the cross-attention contribution at the start of training and allows it to grow thereafter.
The ablation compares this gate with AdaScale gating, AdaScale combined with the tanh gate, unbounded linear conditioning, per-dimension learned scalars, and no gating (App.~\ref{app:ablations}; \citealp{peeblesScalableDiffusionModels2023a}).

\paragraph{Alignment head projection.} We zero-initialize $\mathbf{W}_a\in\mathbb{R}^{d\times d}$ so that the alignment head produces zero logits at the start of training, allowing the generation objective to stabilize before alignment gradients take effect.

\subsection{Training}
\label{app:training-stages}

Training first introduces the text pathways for monomer generation and then adapts the model to generate a binder against a fixed target structure.
The monomer phase contains three stages.

\paragraph{Monomer training.} We follow the staged adaptation used by PixArt-$\alpha$ \citep{chenPixArtaFastTraining2023}.
The first stage warms up the text projectors $g_s$ and $g_t$, cross-attention parameters $\{\mathbf{W}_Q,\mathbf{W}_K,\mathbf{W}_V,\mathbf{W}_O,\alpha_\ell\}$, and AdaLN conditioning layers on paired text--structure data while freezing the pretrained denoiser $\mathbf{v}_\phi$.
The second stage unfreezes all parameters for text-conditioned continual pretraining on PDB and AFDB monomers with $\mathcal{L}_\mathrm{FM}$.
The third stage removes the flow-matching loss and fine-tunes the alignment pathway with $\mathcal{L}_\mathrm{align}$.
During this stage, we freeze the denoiser, AdaLN layers, cross-attention gates, and query and key projections.
We train the value and output projections in the last five cross-attention layers together with the text projectors and alignment head $\mathbf{W}_a$.

Restricting alignment fine-tuning to the late value and output projections preserves fold accuracy.
Higher learning rates or unfrozen query and key projections degrade fold accuracy within one to two epochs, whereas training the last five cross-attention layers matches all fourteen layers with one third as many trainable parameters.
During warmup and both continual-pretraining stages, we replace $\mathbf{c}^\mathrm{s}$ and $\mathbf{c}^\mathrm{t}$ with zero vectors with probability $p_\mathrm{uncond}=0.2$.
This dropout trains the unconditional prediction used for classifier-free guidance.

\paragraph{Binder training.} We continue training on Teddymer dimers \citep{didiScalingAtomisticProtein2026} and PDB multimers.
Only the binder tokens are noised, leaving the target tokens clean and fixed for the denoiser to attend to at every step.
An optional hotspot mask marks target residues that the binder should contact and enters through the pair and sequence features used for the target.
We sample hotspots from true contact residues with dropout so that the training distribution also includes prompts without a specified epitope.
Input text describes the binder, while the target structure enters separately.

Training the final monomer and binder models required approximately 12{,}900 H100-equivalent GPU-hours, or about 67 days on one eight-H100 node.
This estimate excludes ablations, dataset construction, and evaluation.

Every training stage uses a 750-step linear warmup, an effective batch size of 128, and exponential-moving-average decay of 0.999 (Table~\ref{tab:training-stages}).
The learning rate is $1.8{\times}10^{-4}$, except for alignment fine-tuning at $5.4{\times}10^{-6}$.

\begin{table}[H]
\caption{\textbf{Training stages.}
``Trainable'' identifies components with gradients enabled; all other components are frozen.
Steps are optimizer steps at the selected checkpoint.}
\label{tab:training-stages}
\centering
\footnotesize
\setlength{\tabcolsep}{3pt}
\begin{tabular}{@{}>{\raggedright\arraybackslash}p{2.9cm} >{\raggedright\arraybackslash}p{2.5cm} >{\raggedright\arraybackslash}p{3.9cm} c c r@{}}
\toprule
\textbf{Stage} & Dataset & Trainable & Loss & LR & Steps \\
\midrule
Projector warmup & PDB+AFDB monomers & Text modules, AdaLN & $\mathcal{L}_\mathrm{FM}$ & $1.8{\times}10^{-4}$ & 2.6K \\
Continual pretraining (monomers) & PDB+AFDB monomers & All & $\mathcal{L}_\mathrm{FM}$ & $1.8{\times}10^{-4}$ & 816K \\
Alignment fine-tuning & PDB+AFDB monomers & Cross-attn $\mathbf{W}_V$/out (late layers), projectors, $\mathbf{W}_a$ & $\mathcal{L}_\mathrm{align}$ & $5.4{\times}10^{-6}$ & 10.2K \\
\midrule
Continual pretraining (binders) & Teddymer+PDB dimers & All & $\mathcal{L}_\mathrm{FM}$ & $1.8{\times}10^{-4}$ & 465K \\
\bottomrule
\end{tabular}
\end{table}

\paragraph{Training distribution rebalancing.} A small number of architectures account for most PDB chains, so uniform chain sampling presents common folds many times per epoch and rare folds only a few times.
We cluster structures by the structural features described in their captions and interpolate each cluster's sampling weight between its empirical frequency and a uniform distribution over clusters.
A parameter $\beta\in[0,1]$ controls this interpolation.
Setting $\beta{=}0$ recovers uniform sampling over chains, and larger values progressively upweight rare feature combinations, by roughly an order of magnitude for the rarest architectures.
A cosine schedule alternates between the natural distribution and phases that emphasize rare folds.
Rebalancing produces the largest measured gain in fold control, and cosine cycling performs comparably to a monotonic ramp (CATH class accuracy 92.0\% with cosine cycling, 87.3\% with a linear ramp, and 51.3\% without rebalancing; App.~\ref{app:ablations}).

\paragraph{Timestep distribution.} Complexa samples timesteps from $p_{t_x}=0.02\cdot\mathrm{Unif}(0,1)+0.98\cdot\mathrm{Beta}(1.9,1.0)$ and $p_{t_z}=0.02\cdot\mathrm{Unif}(0,1)+0.98\cdot\mathrm{Beta}(1.0,1.5)$, concentrating training near clean data ($t\to 1$).
We replace the backbone schedule with $p_{t_x}=\mathrm{Unif}(0,1)$ and retain the side-chain schedule $p_{t_z}$.
The uniform backbone schedule gives more weight to intermediate timesteps, where the model makes structural decisions and text conditioning is most informative.

\paragraph{Alignment head representation.} The alignment head uses cross-attention updates as its source representation.
The representation ablation compares these updates with trunk hidden states and output-head activations (App.~\ref{app:ablations}).

\subsection{Sampling}
\label{app:sampling}

At inference, Prompteina generates backbone coordinates and side-chain latents by simulating a stochastic differential equation (SDE).
The frozen variational autoencoder (VAE) decoder $\mathcal{D}(\mathbf{x}^{\mathrm{C}_\alpha},\mathbf{z})$ reconstructs all-atom coordinates and amino acid sequence.
Sampling follows \citet{geffnerProteinaScalingFlowbased2025a, didiScalingAtomisticProtein2026} for 400 Euler--Maruyama steps.

Classifier-free guidance \citep{hoClassifierFreeDiffusionGuidance2022a} combines conditional and unconditional velocities:
\begin{equation}
\label{eq:cfg}
\mathbf{v}_\mathrm{guided}
=
w\,\mathbf{v}_\mathrm{cond}
+(1-w)\,\mathbf{v}_\mathrm{uncond},
\end{equation}
where $w$ controls the strength of text conditioning.
The evaluation protocol specifies the guidance weights and length distributions used for each experiment (App.~\ref{app:eval-protocol}).

\FloatBarrier
\section{Ablations}

\label{app:ablations}

We ablate the three choices that determine how text enters the model and which training examples and timesteps supply its learning signal: the conditioning architecture, the $\beta$ data-sampling schedule, and the noise schedule.
All ablations share the same parent checkpoint, dataset, and training duration of 20 epochs.
Evaluation uses the same protocol as \S\ref{sec:exp-structure-generation}: 50 structures per prompt, 318 prompts, and CFG $w$=2.0.

\paragraph{Variants.}
\textit{Complete model}: pooled + cross-attention, cosine $\beta$ cycling, uniform noise.
\textit{Cross-attn only}: removes the pooled/AdaLN text signal.
\textit{Pooled only}: removes cross-attention and the token-level encoder.
\textit{No $\beta$}: uniform cluster sampling ($\beta$=0), removing rare-class rebalancing.
\textit{Linear $\beta$}: monotonic $\beta$ ramp 0$\to$1 (vs.\ cosine cycling).
\textit{Default noise}: backbone C$_\alpha$ noise sampled from Beta(1.9, 1.0) instead of uniform.

\paragraph{Ablation results.}
The complete model obtains 92.0\% CATH class accuracy and 27.0\% CATH architecture accuracy (Table~\ref{tab:ablation}).
CATH class accuracy is 86.7\% with cross-attention alone, 68.7\% with the pooled pathway alone, and 92.0\% with both.
It is 51.3\% without rare-class rebalancing, 87.3\% with a linear $\beta$ schedule, and 92.0\% with cosine $\beta$ cycling.
Uniform and inherited Beta backbone-noise schedules give 92.0\% and 68.7\% CATH class accuracy, respectively.

\begin{table}[!htbp]
\centering
\caption{\textbf{Architecture ablations.}
All metrics use CFG $w=2.0$.
Categorical properties report accuracy, and count properties report Spearman $\rho$.
Rows match Table~\ref{tab:monomer_eval-full} where applicable.
\textbf{Bold} and \underline{underline} denote the best and second-best values.}
\label{tab:ablation}
\footnotesize
\setlength{\tabcolsep}{3.5pt}
\begin{tabular}{ll c c c c c c}
\toprule
\textbf{Group} & Property & Complete & Linear $\beta$ & No $\beta$ & Cross-attn only & Default noise & Pooled only \\
\midrule
\multicolumn{8}{l}{\textit{Fold \& topology, accuracy (\%) ($\uparrow$)}} \\
& CATH class & \textbf{92.0} & \underline{87.3} & 51.3 & 86.7 & 68.7 & 68.7 \\
& CATH architecture & \textbf{27.0} & \underline{25.9} & 22.2 & 25.0 & 18.7 & 7.5 \\
& Contact order & 29.6 & \textbf{36.4} & 28.8 & 28.4 & 31.2 & \underline{33.2} \\
& Internal symmetry & 16.0 & \underline{17.6} & 16.9 & \textbf{22.2} & 14.9 & 10.4 \\
& Sheet topology & 48.0 & 50.0 & \textbf{70.0} & \textbf{70.0} & 66.0 & 62.0 \\
& Domain class comp. & \textbf{4.7} & \underline{4.6} & 3.3 & 2.1 & 0.7 & 1.6 \\
& Domain arch.\ comp. & \textbf{13.3} & 10.5 & \underline{11.3} & 10.5 & 6.3 & 0.5 \\
\midrule
\multicolumn{8}{l}{\textit{Secondary structure, $\rho$ ($\uparrow$)}} \\
& Helix count & \underline{0.87} & 0.86 & 0.87 & 0.87 & \textbf{0.88} & 0.59 \\
& Strand count & \underline{0.75} & 0.70 & \textbf{0.76} & 0.74 & 0.72 & 0.58 \\
& Sheet count & 0.52 & \underline{0.60} & \textbf{0.62} & 0.59 & 0.57 & 0.44 \\
\midrule
\multicolumn{8}{l}{\textit{Tertiary structure, $\rho$ ($\uparrow$)}} \\
& Disulfide count & 0.16 & \underline{0.21} & 0.04 & $-$0.11 & \textbf{0.24} & 0.08 \\
& Salt bridge count & 0.14 & \underline{0.20} & \textbf{0.29} & 0.03 & 0.01 & 0.10 \\
\midrule
\multicolumn{8}{l}{\textit{Global shape \& surface, accuracy (\%) ($\uparrow$)}} \\
& Compactness (size) & \textbf{45.3} & \underline{43.3} & 36.7 & 38.0 & 39.3 & 39.3 \\
& Compactness (shape) & 56.0 & \textbf{72.0} & \underline{60.7} & 58.7 & 36.7 & 45.3 \\
& Surface planarity & \textbf{63.0} & 53.0 & \underline{61.0} & 57.0 & 44.0 & 52.0 \\
& SAP & \textbf{38.7} & 36.0 & 34.7 & 34.0 & \underline{36.7} & 35.3 \\
\midrule
\multicolumn{8}{l}{\textit{Additional metrics}} \\
& CATH topology (\%) & \textbf{1.6} & 1.1 & 1.1 & \underline{1.5} & 1.1 & 0.4 \\
& Buried unsatisfied ($\rho$) & 0.31 & \textbf{0.40} & \underline{0.36} & 0.25 & 0.33 & 0.29 \\
& Flexibility ($\rho$) & 0.43 & 0.45 & 0.43 & \underline{0.46} & 0.45 & \textbf{0.46} \\
\bottomrule
\end{tabular}
\end{table}

\paragraph{Natural-language and CATH-label comparison.}
We compare natural language with a dedicated hierarchical CATH-label conditioner using the same structure generation model, training structures, training step, $\beta$ schedule, requested labels and lengths, evaluator, and CFG.
Pooled text and CATH labels perform similarly through AdaLN, whereas adding token cross-attention to the text path substantially raises CATH class and architecture accuracy, and longer CATH-label training to 3.2 times the matched budget does not close this gap (Table~\ref{tab:cath-label-ablation}).

\begin{table}[!htbp]
\centering
\caption{\textbf{Natural-language and CATH-label conditioning.}
The Complete, Pooled only, and 47k CATH-label columns use the same 47,275-step training budget; the final two columns extend only CATH-label training.
Values are GearNet top-1 accuracy (\%) at CFG $w=2.0$ over three CATH classes and 35 CATH architectures, with 50 structures per requested label.
\textbf{Bold} marks the best value in each row.}
\label{tab:cath-label-ablation}
\footnotesize
\setlength{\tabcolsep}{7pt}
\begin{tabular}{lccccc}
\toprule
Property & \multicolumn{2}{c}{\textbf{Natural language}} & \multicolumn{3}{c}{\textbf{Dedicated CATH label}} \\
\cmidrule(lr){2-3}\cmidrule(lr){4-6}
& Complete & Pooled only & 47k (matched) & 98k & 149k \\
\midrule
CATH class accuracy (\%) & \textbf{92.0} & 68.7 & 66.7 & 73.3 & 67.3 \\
CATH architecture accuracy (\%) & \textbf{27.0} & 7.5 & 8.5 & 9.9 & 9.7 \\
\bottomrule
\end{tabular}
\end{table}

\paragraph{Alignment fine-tuning comparison.}
We assess alignment supervision separately from the shared-parent ablations in Table~\ref{tab:ablation}.
This sequential comparison fine-tunes the text-conditioned monomer checkpoint after 816K optimizer steps for 10.2K additional steps with text--residue alignment supervision.
Both checkpoints are evaluated with the same length-matched protocol at CFG $w$=2.0 and 50 samples per prompt.
After alignment fine-tuning, disulfide-count correlation changes from 0.054 to 0.294 and salt-bridge-count correlation from 0.566 to 0.663 (Table~\ref{tab:alignment-finetuning}).
CATH class accuracy changes from 94.0\% to 96.0\%, and architecture accuracy from 55.5\% to 51.1\%.

\begin{table}[!htbp]
\centering
\caption{\textbf{Alignment fine-tuning comparison.}
The second checkpoint is trained directly from the first, so this sequential comparison is separate from the shared-parent ablations in Table~\ref{tab:ablation}.
The after-alignment values are those reported for Prompteina in Tables~\ref{tab:structure-control} and~\ref{tab:monomer_eval-full}.
Both checkpoints use the same length-matched evaluation at CFG $w=2.0$; disulfide correlation covers requested counts 1--7.}
\label{tab:alignment-finetuning}
\footnotesize
\setlength{\tabcolsep}{8pt}
\begin{tabular}{lcc}
\toprule
Metric & Before alignment & After alignment \\
\midrule
CATH class accuracy (\%) & 94.0 & 96.0 \\
CATH architecture accuracy (\%) & 55.5 & 51.1 \\
Disulfide count ($\rho$) & 0.054 & 0.294 \\
Salt bridge count ($\rho$) & 0.566 & 0.663 \\
\bottomrule
\end{tabular}
\end{table}

\FloatBarrier
\section{Monomer generation evaluation}
\label{app:eval-protocol}

We evaluate text-guided monomer generation across 18 tasks: 14 single-property and 4 composite tasks.
Each prompt produces 50 structures with classifier-free guidance at $w$=2.0 and 400 SDE diffusion steps.
A prompt requesting property $x$ draws lengths from the empirical training distribution of structures annotated with $x$, using a piecewise-linear CDF constructed from 11 quantile breakpoints (P5 through P95).
For CATH class and architecture, this protocol follows Proteina's treatment of empirical compatibility between chain length and CATH label \citep{geffnerProteinaScalingFlowbased2025a} and avoids unrealistic label--length combinations.

\paragraph{Fold and topology.} CATH class (3 prompts) and architecture (35 prompts) are evaluated with GearNet \citep{zhangProteinRepresentationLearning2023a}, a pre-trained graph neural network for fold classification.
We report the fraction of generated structures whose GearNet-predicted CATH label matches the requested label.
Contact order (5 prompts) is computed from C$_\alpha$ contacts and binned into 5 classes.
We detect nine internal symmetries---cyclic groups C2--C8 and dihedral groups D2--D3---with CE-Symm \citep{myers-turnbullSystematicDetectionInternal2014}.

We evaluate multi-domain CATH class and architecture with separate sets of 16 prompts.
Each prompt specifies the class or architecture of every domain in N-to-C order, such as an N-terminal 3-Layer($\alpha\beta\alpha$) Sandwich domain followed by a C-terminal Orthogonal Bundle domain.
We segment each generated structure into domains with Chainsaw \citep{wellsChainsawProteinDomain2024} and classify each domain with GearNet.
A generated structure is correct only if Chainsaw finds the requested number of domains and GearNet predicts the complete sequence of requested CATH labels in order.
All text-conditioned methods receive the same prompts.
We restrict the class and architecture compositions to those with at least 500 training examples.

\paragraph{Secondary structure.} We compute helix count (20 prompts), strand count (20 prompts), and sheet count (14 prompts) from DSSP assignments \citep{kabschDictionaryProteinSecondary1983a}.
For each property, Spearman $\rho$ measures the association between requested and realized counts across all generated structures.

\paragraph{Matched-length control.}
Property-matched length sampling is necessary because some requests require sufficient chain length; for example, an 80-residue protein cannot reasonably contain 20 helices.
For count-valued properties, however, this protocol could also make length predictive of the requested value and thereby inflate requested--realized correlations.
We test this potential confound on helix and strand count, for which Prompteina shows strong control in Table~\ref{tab:structure-control}b.
For each property, we select the widest contiguous five-count window whose shared P20--P80 training-length range lies below 200 residues, requiring at least two strands.
This gives helix counts 1--5 at 134--193 residues and strand counts 2--6 at 112--169 residues.

Within each property, we reuse the exact same list of 50 lengths at every requested count, making length uninformative about the requested value by construction.
We also generate an empty-prompt control using the same lengths.
The text-conditioned correlations remain strong, whereas both control correlations are near zero (Table~\ref{tab:length-matched-ss}).

\paragraph{Tertiary structure.} We detect disulfide bonds from S$_\gamma$--S$_\gamma$ distances below 2.5~\AA{} in the generated coordinates and report Spearman $\rho$ over seven requested counts from 1 to 7.
For salt bridges, we use Asp/Glu--Lys/Arg atom-pair distances below 4.0~\AA{} and report the correlation across 15 prompts.

\paragraph{Global shape and surface.} We evaluate compactness with three prompts each for size and shape.
Size is the radius of gyration normalized by chain length, $R_g / N^{0.4}$, and shape is the ratio of the first two singular values of the C$_\alpha$ coordinate matrix.
We compute solvent-accessible surface area propensity (SAP, 3 prompts) from the solvent-accessible surface geometry.

\paragraph{Composite evaluations.} Four evaluations pair CATH class with a helix, strand, sheet, or disulfide count.
Both disulfide-count evaluations use counts 1--7.
We report GearNet accuracy for CATH class and Spearman $\rho$ for the count, using the same generated samples for the two reported metrics.

\paragraph{Baselines.} Each baseline model is evaluated only on properties its conditioning interface can express.
Chroma \citep{ingrahamIlluminatingProteinSpace2023a} text mode covers all categories via free-text caption guidance.
Chroma label mode uses discrete CATH, secondary structure, and radius of gyration conditioners (7 categories).
Proteina \citep{geffnerProteinaScalingFlowbased2025a} uses CATH label conditioning with classifier-free guidance (3 categories).
RFdiffusion \citep{watsonNovoDesignProtein2023a} uses per-residue secondary structure tensors via scaffoldguided mode (3 categories).
RFdiffusion3 \citep{butcherNovoDesignAllatom2025b} does not expose helix-, strand-, or sheet-specific conditioning, and its cyclic or dihedral symmetry sampler generates multi-chain assemblies rather than the single-chain internal symmetry evaluated here.

Pinal \citep{daiNovoProteinDesign2025} and BioMatrix \citep{peiBioMatrixComprehensiveBiological2026} are text-conditioned models evaluated on the same prompts as Prompteina, and both generate amino acid sequences that we fold with ESMFold.
Pinal accepts a target length directly, whereas BioMatrix exposes no length control, so we constrain its decoder to emit exactly the sampled number of residues before terminating.
All baselines use their paper-default settings and the same length distributions as Prompteina.

\begin{table}[H]
\centering
\caption{\textbf{Structural-control benchmark.}
We report accuracy (\%) for categorical properties and Spearman $\rho$ for count properties.
Prompteina uses classifier-free guidance weight $w$=2.0.
Baselines are evaluated only on properties their input format can express (--- = not expressible).
Best and second-best values are \textbf{bold} and \underline{underlined}, respectively.}
\label{tab:monomer_eval-full}
\footnotesize
\setlength{\tabcolsep}{2.0pt}
\begin{tabular}{@{}l c cccccc@{}}
\toprule
& \textbf{Ours} & \multicolumn{6}{c}{\textbf{Baselines}} \\
\cmidrule(lr){3-8}
Property & Prompteina & Chroma & Chroma & Proteina & RFD & Pinal & BioMatrix \\
& & (text) & (label) & & & & \\
\midrule
\multicolumn{8}{@{}l}{\textit{Fold \& topology, accuracy (\%) ($\uparrow$)}} \\
\quad CATH class & \underline{96.0} & 33.3 & 64.7 & \textbf{96.7} & --- & 36.0 & 30.7 \\
\quad CATH architecture & \textbf{51.1} & 3.7 & 7.5 & \underline{27.5} & --- & 5.5 & 2.8 \\
\quad Contact order & \textbf{56.0} & 21.2 & --- & --- & --- & \underline{27.6} & 21.6 \\
\quad Internal symmetry & \textbf{32.4} & 0.4 & --- & --- & --- & \underline{1.1} & 0.4 \\
\quad Multi-domain CATH class & \textbf{34.4} & 5.3 & --- & --- & --- & \underline{5.4} & 2.6 \\
\quad Multi-domain CATH arch. & \textbf{42.0} & \underline{1.5} & --- & --- & --- & 0.4 & 1.0 \\
\midrule
\multicolumn{8}{@{}l}{\textit{Secondary structure, $\rho$ ($\uparrow$)}} \\
\quad Helix count & \textbf{0.94} & 0.74 & 0.86 & --- & \underline{0.87} & 0.72 & 0.59 \\
\quad Strand count & \underline{0.88} & 0.42 & 0.58 & --- & \textbf{0.92} & 0.25 & 0.18 \\
\quad Sheet count & \underline{0.64} & 0.40 & 0.59 & --- & \textbf{0.82} & 0.16 & 0.18 \\
\midrule
\multicolumn{8}{@{}l}{\textit{Tertiary structure, $\rho$ ($\uparrow$)}} \\
\quad Disulfide count & \textbf{0.29} & 0.06 & --- & --- & --- & \underline{0.10} & 0.07 \\
\quad Salt bridge count & \textbf{0.66} & \underline{0.64} & --- & --- & --- & 0.23 & 0.28 \\
\midrule
\multicolumn{8}{@{}l}{\textit{Global shape \& surface, accuracy (\%) ($\uparrow$)}} \\
\quad Compactness (size) & \textbf{76.7} & 34.0 & \underline{39.3} & --- & --- & 36.7 & 33.3 \\
\quad Compactness (shape) & \textbf{71.3} & 31.3 & --- & --- & --- & \underline{34.7} & 30.9 \\
\quad SAP & \textbf{38.7} & \underline{37.3} & --- & --- & --- & 34.0 & 34.0 \\
\midrule
\multicolumn{8}{@{}l}{\textit{Composite evaluations, class acc.\ (\%) / count $\rho$ ($\uparrow$)}} \\
\quad Class $\times$ helix & \textbf{84.1}\,/\,\textbf{0.86} & 31.9\,/\,\underline{0.61} & --- & --- & --- & \underline{38.6}\,/\,0.10 & 35.9\,/\,0.11 \\
\quad Class $\times$ strand & \textbf{92.1}\,/\,\textbf{0.65} & 32.4\,/\,\underline{0.38} & --- & --- & --- & \underline{35.5}\,/\,0.21 & 30.7\,/\,0.35 \\
\quad Class $\times$ sheet & \textbf{94.4}\,/\,\textbf{0.46} & \underline{36.7}\,/\,\underline{0.34} & --- & --- & --- & 33.9\,/\,0.06 & 28.1\,/\,0.07 \\
\quad Class $\times$ disulf. count & \textbf{90.3}\,/\,\textbf{0.43} & \underline{34.5}\,/\,$-$0.02 & --- & --- & --- & 33.4\,/\,\underline{0.10} & 31.0\,/\,0.07 \\
\bottomrule
\end{tabular}
\end{table}

\begin{table}[H]
\centering
\caption{\textbf{Matched-length secondary-structure control.}
Within each property, the same 50-length list is reused at every requested count and for the empty-prompt control ($n=250$ per row).
Control groups inherit the text-conditioned samples' count labels only to compute the null correlation; no count is requested in their prompts.
Slope is from a linear regression of realized count on the count label.}
\label{tab:length-matched-ss}
\small
\setlength{\tabcolsep}{6pt}
\begin{tabular}{@{}lccccc@{}}
\toprule
\textbf{Experiment} & Requested counts & Lengths (aa) & $n$ & Spearman $\rho$ & Slope \\
\midrule
Helix           & 1--5 & 134--193 & 250 & \textbf{0.878} & \textbf{1.076} \\
Helix---control & ---  & 134--193 & 250 & $-$0.060 & $-$0.060 \\
\midrule
Strand           & 2--6 & 112--169 & 250 & \textbf{0.718} & \textbf{0.824} \\
Strand---control & ---  & 112--169 & 250 & $-$0.005 & $-$0.010 \\
\bottomrule
\end{tabular}
\end{table}

\FloatBarrier
\subsection{Prompt paraphrase robustness}
\label{app:monomer-paraphrase}

We test whether control of three well-supported properties depends on the benchmark's exact prompt template: CATH class, helix count, and their composition.
For each semantic specification, we compare direct, short-form, paraphrased, and reordered wording (Box~\ref{box:monomer-paraphrase-examples}).
The paraphrases were written with Claude rather than the Qwen model used to generate training captions.
All four formats use the same 20 chain lengths for each specification, so wording is the only manipulated variable.
All samples are generated specifically for this matched experiment rather than reused from Table~\ref{tab:monomer_eval-full}; there are therefore 20 rather than 50 samples per specification.

\begin{promptbox}{Monomer paraphrase examples}{box:monomer-paraphrase-examples}
\textbf{CATH class: Alpha Beta}\\
Direct: This protein has an Alpha Beta structural class.\\
Short-form: CATH class: Alpha Beta.\\
Paraphrased: The overall fold falls into the Alpha Beta category.\\
Reordered: An Alpha Beta structural class describes this protein.\\[4pt]

\textbf{Helix count: 6}\\
Direct: This protein contains 6 helices.\\
Short-form: 6 helices.\\
Paraphrased: The structure is built from six alpha helices.\\
Reordered: Six helices make up this protein.\\[4pt]

\textbf{Composite: Alpha Beta class and 4 helices}\\
Direct: This protein has an Alpha Beta structural class and contains 4 helices.\\
Short-form: Alpha Beta class; 4 helices.\\
Paraphrased: With four alpha helices, this protein adopts an Alpha Beta fold.\\
Reordered: Build a protein with four helices whose overall structural class is Alpha Beta.
\end{promptbox}

\begin{table}[H]
\centering
\caption{\textbf{Prompt paraphrases.}
CATH class uses three specifications, helix count uses five, and the composite uses six; each specification has 20 matched-length generations per format.
We report accuracy for CATH class, Spearman $\rho$ for helix count, and accuracy / $\rho$ for their composition.}
\label{tab:monomer-paraphrase}
\small
\setlength{\tabcolsep}{7pt}
\begin{tabular}{@{}lcccc@{}}
\toprule
\textbf{Evaluation} & \textbf{Direct} & \textbf{Short-form} & \textbf{Paraphrased} & \textbf{Reordered} \\
\midrule
CATH class accuracy (\%) & 100.0 & 100.0 & 95.0 & 96.7 \\
Helix count $\rho$ & 0.953 & 0.942 & 0.912 & 0.934 \\
\midrule
Class $\times$ helix (acc.\ \% / $\rho$) & 92.5\,/\,0.734 & 87.5\,/\,0.652 & 90.8\,/\,0.729 & 86.7\,/\,0.694 \\
\bottomrule
\end{tabular}
\end{table}

All four formats retain strong control of the tested properties (CATH class accuracy at least 95.0\%, helix-count $\rho$ at least 0.912, and composite class accuracy and helix-count $\rho$ at least 86.7\% and 0.652; Table~\ref{tab:monomer-paraphrase}).
Reversing the order of the two composite properties leaves both metrics similar, so control does not depend on whether CATH class or helix count appears first in the prompt.

\subsection{Designability}
\label{app:monomer-quality}

Following the standard self-consistency evaluation for generated backbones \citep{yimSE3DiffusionModel2023a,watsonNovoDesignProtein2023a}, we sample eight sequences for each structure with ProteinMPNN \citep{dauparasRobustDeepLearning2022a}, fold them with ESMFold, and take the minimum C$_\alpha$ self-consistency RMSD (scRMSD) to the generated structure.
This best-of-eight inverse-folding test asks whether at least one designed sequence is predicted to realize the proposed backbone, rather than requiring the generation model's own sequence to do so.
Following prior backbone-generation evaluations \citep{yimSE3DiffusionModel2023a,watsonNovoDesignProtein2023a}, we classify a structure as designable when scRMSD is below 2~\AA{} for every method.

\begin{figure}[!htbp]
\centering
\includegraphics[width=\linewidth,height=0.84\textheight,keepaspectratio]{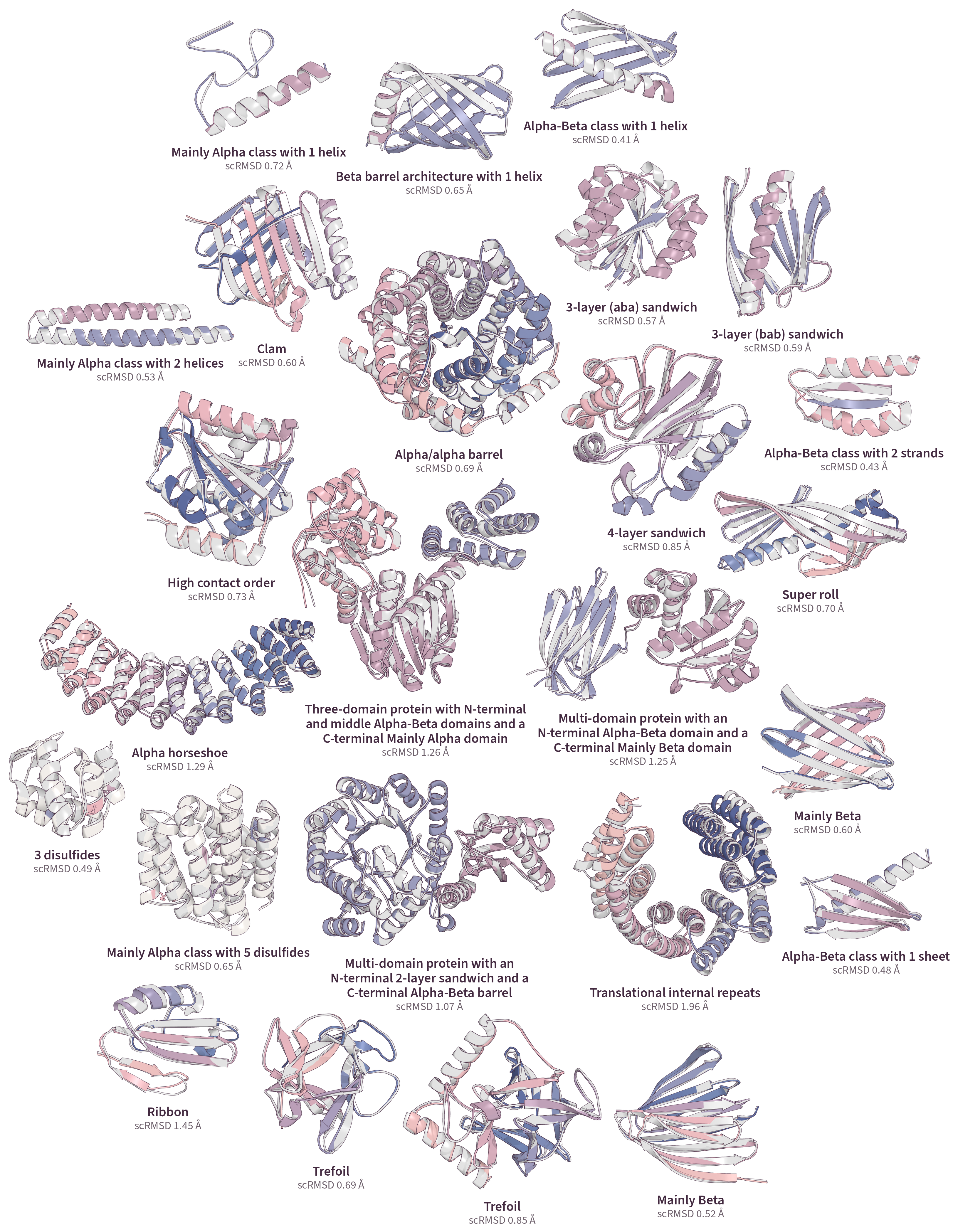}
\caption{\textbf{Additional monomer samples.}
Generated structures are colored and superposed with their best of eight ProteinMPNN--ESMFold refolds (light gray) after C$_\alpha$ alignment.
Labels give the requested property; values give scRMSD.}
\label{fig:monomer-designable-examples}
\end{figure}

For every prompt and method, we evaluate an evenly spaced 10-of-50 subset (every fifth structure after sorting by sample index); this preserves coverage of all test cases while controlling the cost of eight refolds per structure.
Each baseline supports a different benchmark subset, so each row of Table~\ref{tab:monomer-designability} restricts Prompteina to the test cases evaluated by that baseline; rows should therefore be read across, not columns down.
This avoids treating unsupported tasks as failures or comparing averages over different task mixtures.
Additional generated structures realize the properties requested in their prompts, together with the corresponding refolded structures used to evaluate designability (Fig.~\ref{fig:monomer-designable-examples}).

\begin{table}[H]
\centering
\caption{\textbf{Monomer designability.}
Each row compares Prompteina with one baseline on exactly the benchmark test cases both evaluated.
Median scRMSD is in \AA{} (lower is better); designable is the percentage with scRMSD $<2$~\AA{} (higher is better).
``Structures per model'' gives the sample size on each side of the comparison.
Complexa and the text-guided models share all 18 tasks.
The property-specific scopes are CATH architecture and class for Proteina, six tasks for Chroma label, and the three secondary-structure-count tasks for RFdiffusion.
\textbf{Bold} marks the better result within each pair.}
\label{tab:monomer-designability}
\footnotesize
\setlength{\tabcolsep}{2.5pt}
\begin{tabular*}{\linewidth}{@{\extracolsep{\fill}}lrrcccc@{}}
\toprule
\textbf{Baseline} & Shared test & Structures & \multicolumn{2}{c}{Median scRMSD ($\downarrow$)} & \multicolumn{2}{c}{Designable (\%) ($\uparrow$)} \\
& cases & per model & Prompteina & Baseline & Prompteina & Baseline \\
\midrule
Complexa & 251 & 2,510 & \textbf{0.83} & 3.11 & \textbf{79.7} & 29.1 \\
\midrule
\multicolumn{7}{@{}l}{Text-guided models} \\
Chroma (text) & 251 & 2,510 & \textbf{0.83} & 14.07 & \textbf{79.7} & 2.4 \\
Pinal & 251 & 2,510 & \textbf{0.83} & 11.10 & \textbf{79.7} & 16.5 \\
BioMatrix & 251 & 2,510 & \textbf{0.83} & 15.83 & \textbf{79.7} & 7.0 \\
\midrule
\multicolumn{7}{@{}l}{Property-specific models} \\
Proteina & 27 & 270 & \textbf{0.69} & 1.17 & \textbf{91.1} & 64.8 \\
Chroma (label) & 84 & 840 & \textbf{0.78} & 14.80 & \textbf{81.9} & 1.7 \\
RFdiffusion & 54 & 540 & \textbf{0.83} & 1.19 & \textbf{76.5} & 68.1 \\
\bottomrule
\end{tabular*}
\end{table}

Conditional and empty-prompt samples use the same fine-tuned checkpoint, the same 251 prompts, and identical per-prompt length distributions; the empty-prompt condition zeros the text embeddings and uses guidance weight $w=1$.
Holding the checkpoint and lengths fixed makes this a control for whether the designability gain comes from fine-tuning or from text conditioning, while preventing the strong dependence of designability on length from confounding the comparison.

Fine-tuned Prompteina is substantially more designable than pretrained Complexa with or without a prompt (79.7\% conditional and 85.2\% empty-prompt vs.\ 29.1\%).
Although the pooled empty-prompt rate is 5.5 percentage points higher, the median paired per-prompt difference is zero; this is not evidence that removing the prompt improves designability.
The conditional Prompteina advantage replicates with ESMFold2 (80.7\% vs.\ 26.7\% designable; paired scRMSD correlation $\rho=0.953$).
This checks that the conclusion is not specific to one structure predictor; we treat it as a replication with a second folding model, not as a better ground truth.

Prompteina is more designable than all three text-guided full-suite baselines (79.7\% vs.\ 2.4--16.5\%).
The Proteina comparison uses a different sequence-design instrument because Proteina emits only a C$_\alpha$ trace.
The specialist rows also cover narrower task sets.

\subsection{Structural novelty}
We measure novelty only among designable structures, preventing structurally failed outputs from appearing novel merely because they match no protein fold.
For each design, Foldseek \citep{vankempenFastAccurateProtein2024} with TM-align searches the training structures used by that model, and novelty is one minus the maximum query-normalized TM-score \citep{zhangScoringFunctionAutomated2004} to any training structure.
Taking the maximum tests the nearest potential training-set match rather than average similarity.
Query normalization asks how much of the generated design is explained by that match, so a short design contained within a longer training structure is conservatively treated as non-novel.
We use model-specific training references because the relevant memorization opportunity is the corpus to which each model was exposed; consequently, cross-model values are descriptive rather than a controlled comparison against a common database.

Prompteina's conditional and empty-prompt samples share the same 780,195-structure PDB--AFDB training reference, so their matched comparison isolates the effect of supplying text.
Both sets begin with 2,510 structures; 2,000 conditional and 2,138 empty-prompt structures pass the scRMSD $<2$~\AA{} designability criterion and enter the novelty analysis.

For Complexa, 731 of 2,510 structures are designable.
Complexa does not release its original chain-level monomer manifest, so we reconstruct the reported 110,976-chain PDB monomer set from the checkpoint filters and archived PDB metadata, retaining 110,969 after the checkpoint's explicit exclusions.
We combine these monomers with the released 45,856-entry PDB multimer manifest.
Expanding multimer entries into their constituent chains and structural models and combining them with the 289,222 AFDB targets gives 506,972 Foldseek targets.
Including both PDB components prevents Complexa from appearing artificially novel because part of its training exposure is absent from the search database.
Both model-specific references omit Teddymer structures; this can overestimate absolute novelty, but the reconstructed Complexa reference corrects the previous additional omission of its PDB training structures.

\begin{table}[H]
\centering
\caption{\textbf{Structural novelty.}
Each row is searched against the corresponding model-specific PDB--AFDB training reference; reference size counts Foldseek targets.
Maximum TM is the best query-normalized structural match to that reference; lower maximum TM and higher novelty indicate greater distance from the model's training set.
The two Prompteina conditions use the same checkpoint, prompts, lengths, and reference set.
\textbf{Bold} marks the better result in this matched conditional--empty-prompt comparison.}
\label{tab:monomer-novelty}
\footnotesize
\setlength{\tabcolsep}{6pt}
\begin{tabular}{@{}lrcc@{}}
\toprule
\textbf{Model / condition} & Reference size & Max TM ($\downarrow$) & Mean novelty ($\uparrow$) \\
& & mean / median & \\
\midrule
Complexa & 506,972 & 0.783 / 0.782 & 0.217 \\
\midrule
Prompteina (conditional) & 780,195 & \textbf{0.770 / 0.785} & \textbf{0.230} \\
Prompteina (empty prompt) & 780,195 & 0.835 / 0.847 & 0.165 \\
\bottomrule
\end{tabular}
\end{table}

Conditional Prompteina and Complexa have similar novelty under their model-specific PDB--AFDB references (mean novelty 0.230 vs.\ 0.217).
Text conditioning also moves Prompteina farther from its training structures than the matched empty-prompt control (mean novelty 0.230 vs.\ 0.165).
The matched Prompteina comparison isolates the prompt effect; the cross-model comparison remains descriptive because the reference sets differ and both omit Teddymer.

\subsection{Full prompts for Figure~\ref{fig:overview}a}
\label{app:fig1-prompts}

The two prompts abbreviated in Figure~\ref{fig:overview}a are:
\begin{itemize}[leftmargin=*]
\item ``Mechanobiology sensors need an element that reports tension by stretching reproducibly. It is built as a straight up-down bundle of six alpha helices packed side by side along one long axis, with two beta strands clasping one end. Two disulfide bonds staple the clasp and at least six salt bridges run along the bundle surface. The silhouette is elongated.''
\item ``This enzyme is intended to hydrolyse organophosphate pesticides under EC 3.1.8.1. It adopts a Super Roll architecture of the Alpha Beta class, built from four alpha helices and six beta strands, with the sheet rolled around the helices and a metal centre at the open mouth of the roll. The chain forms zero disulfide bonds and a square silhouette.''
\end{itemize}

\FloatBarrier
\section{Functional specification evaluation}

\label{app:functional-eval}

We test whether a text prompt can specify a protein family and generate recognizable members.
The structural properties in \S\ref{sec:exp-structural} have property-specific verifiers, but no corresponding classifier covers arbitrary protein families.

\paragraph{Benchmark families.}
The benchmark comprises 122 sequence-coherent protein families.
We define each candidate family as an ECOD topology group \citep{chengECODEvolutionaryClassification2014} whose PDB members exceed 30\% median pairwise sequence identity, the threshold below which alignment stops implying homology \citep{rostTwilightZoneProtein1999}.
This requirement removes topology groups that share a fold without a single family-level referent.

Eligibility is computed from the median identity over all pairs of PDB sequences in a group before any structure is generated.
No design, method, or evaluation score enters the benchmark definition.

\paragraph{Prompt formats.}
The models were trained with different textual vocabularies, so a single prompt format would favor the model for which that wording is in-distribution.
We evaluate every method on four formats and report its highest score for each metric.
The \emph{ECOD annotation} combines an ECOD topology name with an EC number or cofactor annotation when available, matching Prompteina's caption vocabulary.
The \emph{functional description} paraphrases a Swiss-Prot \texttt{FUNCTION} annotation with Claude, matching Pinal's training and evaluation setting.
The \emph{protein metadata} combines the protein name with an EC number and subcellular location when available, matching BioMatrix's keyword setting.
The \emph{InterPro label} contains only the InterPro entry name, which is a second format used by Pinal \citep{daiNovoProteinDesign2025,peiBioMatrixComprehensiveBiological2026}.

The exact prompts for one benchmark family illustrate these four formats (Box~\ref{box:functional-prompt-examples}).

\begin{promptbox}{Functional-evaluation prompt formats: globin example}{box:functional-prompt-examples}
\textbf{ECOD annotation}\\
This protein adopts a globin-like fold.
It coordinates a heme cofactor.\\[3pt]

\textbf{Functional description}\\
Carries oxygen from the lungs to peripheral tissues.
Its derived peptide LVV-hemorphin-7 enhances bradykinin activity and thereby lowers blood pressure, and it also inhibits enkephalin-degrading enzymes such as DPP3 and selectively antagonizes the pain-signaling receptor P2RX3, implicating it in the control of pain and inflammation.\\[3pt]

\textbf{Protein metadata}\\
hemoglobin subunit beta.\\[3pt]

\textbf{InterPro label}\\
Globin
\end{promptbox}

\paragraph{Generation protocol.}
Every method generates 50 structures per family and format under the common sampling protocol (App.~\ref{app:eval-protocol}).
Pinal and BioMatrix emit sequences, which we fold with ESMFold before computing structure similarity.

\paragraph{ProTrek similarity.}
ProTrek is a contrastively trained trimodal encoder mapping protein sequences, structures, and text into a shared embedding space \citep{suTrimodalProteinLanguage2025a}.
Both scores are temperature-scaled cosine similarities between the prompt embedding and a generated sequence or structure embedding.

Confidence-weighted similarity reduces the contribution of generated sequences that ESMFold predicts with low confidence without filtering any design or family.
For each design, it multiplies ProTrek similarity by mean per-residue $\mathrm{pLDDT}/100$.
We obtain pLDDT by applying ESMFold to the exact generated sequence, using Prompteina's co-designed sequence or the sequence emitted by Pinal or BioMatrix.
Scores are averaged first within each family and then across the 122 families, keeping the family as the unit of analysis.
We estimate 95\% confidence intervals by cluster-bootstrapping the 122 families 10,000 times and taking the 2.5th and 97.5th percentiles.

\begin{table}[!htbp]
\centering
\caption{\textbf{Functional control across prompt formats.}
Mean ProTrek similarity between a prompt and the sequence or folded structure generated from it, over 50 designs for each of 122 protein families.
Formats marked $\dagger$ are in-distribution for that method.
\textbf{Bold} marks each method's best format.
$\pm$ denotes the half-width of a 95\% cluster-bootstrap interval obtained by resampling the 122 families rather than designs within a family.}
\label{tab:functional_formats}
\small
\begin{tabular*}{\linewidth}{@{\extracolsep{\fill}}lcccc@{}}
\toprule
\multicolumn{5}{c}{ProTrek seq. sim. ($\uparrow$)} \\
\midrule
\textbf{Method} & ECOD ann. & Function & Metadata & InterPro \\
\midrule
Prompteina & \textbf{8.46}$^\dagger$ {\scriptsize$\pm$\,0.98} & 5.96 {\scriptsize$\pm$\,0.91} & 5.65 {\scriptsize$\pm$\,0.89} & 5.51 {\scriptsize$\pm$\,0.94} \\
Pinal      & \textbf{7.55} {\scriptsize$\pm$\,0.79} & 6.37$^\dagger$ {\scriptsize$\pm$\,0.82} & 6.97 {\scriptsize$\pm$\,0.85} & 7.52$^\dagger$ {\scriptsize$\pm$\,0.72} \\
BioMatrix  & 4.73 {\scriptsize$\pm$\,0.64} & 3.49 {\scriptsize$\pm$\,0.56} & 4.72$^\dagger$ {\scriptsize$\pm$\,0.62} & \textbf{5.73} {\scriptsize$\pm$\,0.68} \\
\bottomrule
\end{tabular*}

\vspace{6pt}

\begin{tabular*}{\linewidth}{@{\extracolsep{\fill}}lcccc@{}}
\toprule
\multicolumn{5}{c}{ProTrek struct. sim. ($\uparrow$)} \\
\midrule
\textbf{Method} & ECOD ann. & Function & Metadata & InterPro \\
\midrule
Prompteina & \textbf{7.36}$^\dagger$ {\scriptsize$\pm$\,1.04} & 4.33 {\scriptsize$\pm$\,0.91} & 4.12 {\scriptsize$\pm$\,0.87} & 4.55 {\scriptsize$\pm$\,0.96} \\
Pinal      & 6.84 {\scriptsize$\pm$\,0.78} & 5.27$^\dagger$ {\scriptsize$\pm$\,0.84} & 6.23 {\scriptsize$\pm$\,0.84} & \textbf{7.09}$^\dagger$ {\scriptsize$\pm$\,0.73} \\
BioMatrix  & 3.57 {\scriptsize$\pm$\,0.56} & 2.09 {\scriptsize$\pm$\,0.48} & 3.33$^\dagger$ {\scriptsize$\pm$\,0.55} & \textbf{4.64} {\scriptsize$\pm$\,0.63} \\
\bottomrule
\end{tabular*}
\end{table}

\paragraph{Baseline training and metric overlap.}
Pinal is designed for text-to-protein generation, whereas BioMatrix is a generalist model instruction-tuned for the same task.
Both generate protein sequences from functional descriptions and include UniProt-derived training data.
Prompteina is trained to generate all-atom structures from captions that combine structural and functional properties.

We retain ProTrek because the original Pinal and BioMatrix evaluations report this metric.
We note that the use of ProTrek for evaluation is not independent of Pinal's training.
Pinal's TrEMBL50-ProTrek corpus was built by scoring TrEMBL cluster representatives against Swiss-Prot descriptions with ProTrek and retaining about 412M of the highest-scoring sequence--description pairs \citep{daiNovoProteinDesign2025}.
Pinal's structure tokens use ProTrek's Foldseek tokenization, and Pinal selects candidates by ProTrek score.
Prompteina uses neither these data nor these selection procedures.

\paragraph{Method comparison.}
At each method's best prompt format, Prompteina has the highest mean sequence and structure similarity (Table~\ref{tab:functional_formats}).
Its unweighted advantage over Pinal is small and unresolved under a paired family bootstrap (sequence $+0.91$ [${-}0.14$, $+1.93$]; structure $+0.26$ [${-}0.73$, $+1.30$]).
The corresponding advantages over BioMatrix are larger and resolved (sequence $+2.73$ [$+1.75$, $+3.68$]; structure $+2.71$ [$+1.74$, $+3.71$]).

Confidence weighting increases the separation without changing the overall conclusion (Table~\ref{tab:functional_summary}).
Prompteina's advantage over Pinal resolves for sequence similarity but remains unresolved for structure similarity (sequence $+1.14$ [$+0.26$, $+2.03$]; structure $+0.72$ [${-}0.15$, $+1.60$]).
Both confidence-weighted advantages over BioMatrix resolve (sequence $+3.42$ [$+2.58$, $+4.25$]; structure $+3.01$ [$+2.18$, $+3.85$]).

\FloatBarrier
\section{Binder design protocols and evaluation}
\label{app:binder-eval}

\subsection{Interface-residue steering}
\label{app:interface-steering}

\paragraph{Interface residue detection.} Following \citet{jonesPrinciplesProteinproteinInteractions1996}, we define interface residues as binder residues whose solvent-accessible surface area (SASA) decreases by more than 1.0~\AA$^2$ upon complex formation.
SASA is computed with FreeSASA using default radii.
The interface residue count reported throughout \S\ref{sec:exp-nres-steering} is the number of binder-side residues meeting this criterion.

\paragraph{Benchmark targets.} We use all 20 targets in the Complexa benchmark \citep{didiScalingAtomisticProtein2026}, retaining its target structures, binder-length ranges, and hotspot residues where provided.
The agentic binder protocol lists the complete panel below.

\paragraph{Prompt template.} We specify only the requested binder--target interface residue count: \texttt{``The binding interface involves \{n\} residues.''}
Requested counts range from 8 to 30; scaffold topology, contact chemistry, and other binder properties remain unspecified.

\paragraph{Generation protocol.} We sample 50 structures per requested count with classifier-free guidance at $w=2.0$.
The single-pass protocol omits search, reward models, inverse folding, and structure prediction so that these components cannot alter the requested-count response.
Within a target, every requested count uses the same 50 binder lengths, sampled uniformly from the target-specific range with a fixed seed.
The target structure and benchmark hotspots also remain fixed, leaving the requested contact-residue count as the only manipulated variable.

\paragraph{Unconditional baseline.} We sample pretrained Complexa with unchanged target inputs, disabled text, and the same binder lengths as the text-conditioned samples.
Classifier-free guidance is undefined when both text and fold conditioning are absent.

\paragraph{Prompt paraphrases.}
We test interface-residue control under key--value, paraphrased, and reordered prompts on CD45, IL-17A, and PDL1 (Box~\ref{box:interface-paraphrase-examples}).
For each target and format, we request 8--37 interface residues and generate 20 structures at each of the 30 levels.
Every format uses the same 20 binder lengths at a given target and level.
Across the 30 requested levels, we report Spearman $\rho$ between requested and mean realized counts and the slope of mean realized on requested count (Table~\ref{tab:interface-paraphrase}).
Response magnitude differs by target, so we express each slope relative to that target's key--value slope rather than pooling targets.

\begin{promptbox}{Interface-residue paraphrase examples at a requested count of 22}{box:interface-paraphrase-examples}
\textbf{Key--value}\\
- Protein-protein interface: 22 interface residues\\[4pt]

\textbf{Paraphrased}\\
The binding interface involves 22 residues.\\[4pt]

\textbf{Reordered}\\
22 residues make up the binding interface.
\end{promptbox}

\begin{table}[!htbp]
\centering
\caption{\textbf{Interface-residue control across prompt formats.}
Each row contains 600 structures (30 requested counts $\times$ 20 matched lengths).
Relative slope divides by the key--value slope separately within each target.}
\label{tab:interface-paraphrase}
\small
\setlength{\tabcolsep}{7pt}
\begin{tabular}{@{}llccc@{}}
\toprule
\textbf{Target} & \textbf{Format} & Slope & Relative slope & Spearman $\rho$ \\
\midrule
\multirow{3}{*}{CD45}
 & Key--value & 0.851 & 1.00 & 0.989 \\
 & Paraphrased & 0.910 & 1.07 & 0.997 \\
 & Reordered & 0.647 & 0.76 & 0.976 \\
\midrule
\multirow{3}{*}{IL-17A}
 & Key--value & 0.808 & 1.00 & 0.997 \\
 & Paraphrased & 0.933 & 1.15 & 0.996 \\
 & Reordered & 0.881 & 1.09 & 0.997 \\
\midrule
\multirow{3}{*}{PDL1}
 & Key--value & 0.346 & 1.00 & 0.985 \\
 & Paraphrased & 0.437 & 1.26 & 0.993 \\
 & Reordered & 0.366 & 1.06 & 0.989 \\
\bottomrule
\end{tabular}
\end{table}

All tested phrasings preserve the requested-count ordering ($\rho\geq0.976$).
Paraphrased prompts retain or increase the key--value response slope on every target, while reordered prompts retain 76--109\% of the key--value response.

\begin{figure}[t]
\centering
\includegraphics[width=\linewidth]{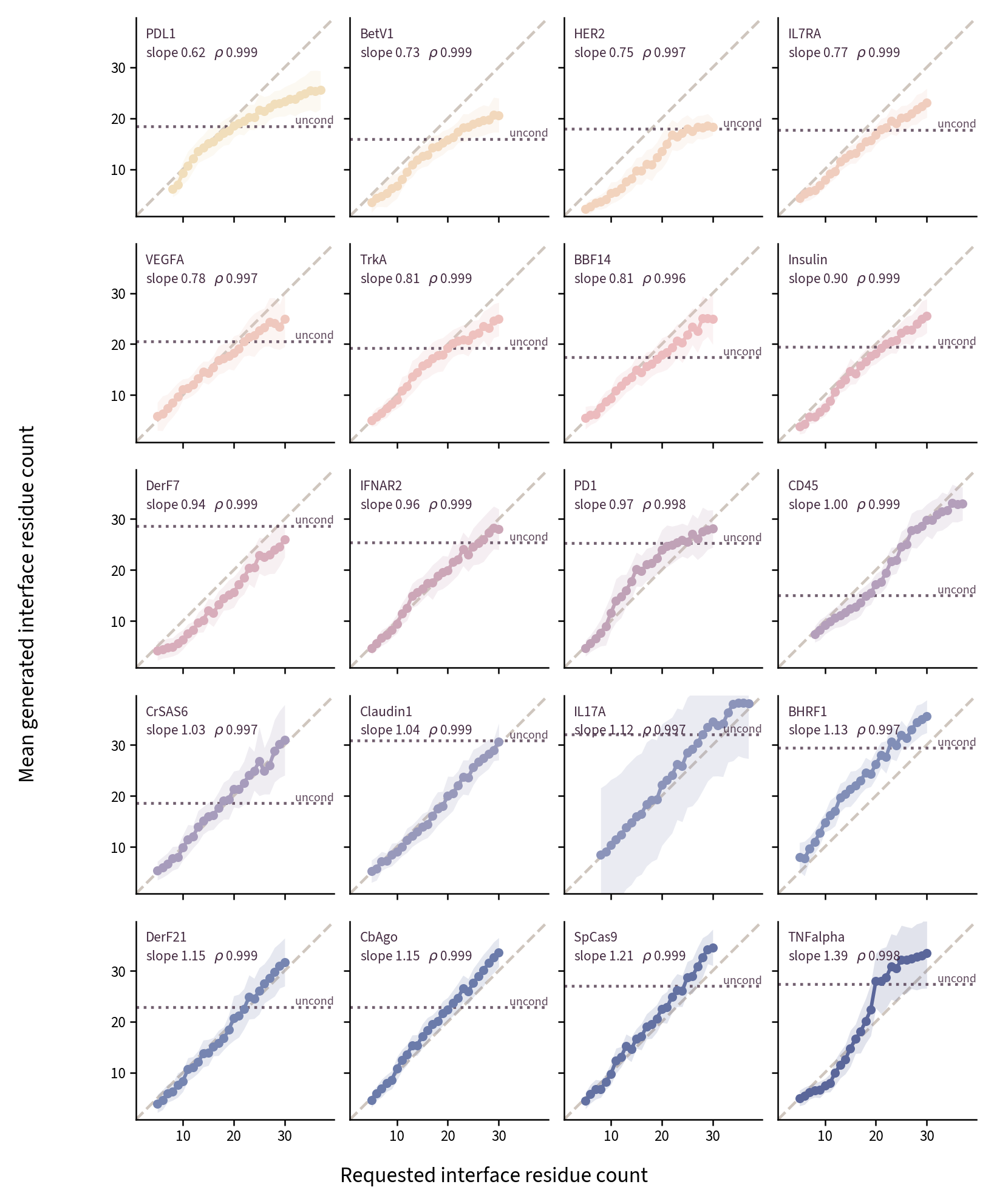}
\caption{\textbf{Interface-residue steering across targets.}
Each panel shows the mean realized count across 50 binders for requested counts 8--30; colored bands span one standard deviation.
Dashed diagonals mark equality.
Dotted horizontal lines mark the per-target unconditional mean.
Annotations give the fitted slope and Spearman's $\rho$; panels are ordered by slope.}
\label{fig:interface-steering-grid}
\end{figure}

\subsection{Agentic binder design protocol}
\label{app:agentic-binder}

\subsubsection{Target panel}

The panel contains 20 protein targets drawn from the AlphaProteo \citep{zambaldiNovoDesignHighaffinity2024} and BindCraft \citep{pacesaOneshotDesignFunctional2025} targets aggregated by \citet{didiScalingAtomisticProtein2026}.
We exclude influenza H1 and the SARS-CoV-2 receptor-binding domain because our prompt-expansion pipeline returned no usable specifications, and apply this exclusion to every method.
IL-17A and TNF-$\alpha$ remain in the panel, although Complexa excludes them after no evaluated method succeeds within its runtime budget.

Each target contributes a structure and binder-length range; curated hotspots are available for 18 targets.
In the agentic setting, each method receives the structure and length range and selects an epitope for each design.
In the non-agentic setting, each method receives those curated hotspots.

\subsubsection{Generation protocol}

Every method generates 200 designs per target in both the agentic and non-agentic settings.
We divide the target's binder-length range into 20 contiguous bands and draw 10 designs in each, preventing a method's length prior from concentrating its samples in part of the range.
Designs are sampled independently without inference-time search, a reward model, or candidate selection.
The common budget and generation procedure isolate differences among conditioning channels.

Prompteina uses 400 SDE denoising steps as in the monomer evaluation, with classifier-free guidance reduced to $w=1.0$ (App.~\ref{app:eval-protocol}).
Baselines use their published inference defaults.

\subsubsection{Sequence conditions}

Each generated backbone is scored under three sequence conditions.
The \emph{self-sequence} condition uses the sequence produced by the structure generation model.
The \emph{redesign} condition samples eight ProteinMPNN sequences per backbone and retains the highest-scoring one \citep{dauparasRobustDeepLearning2022a}.
The \emph{fixed-interface redesign} condition also uses ProteinMPNN but fixes the generated identities of binder interface residues, changing only the scaffold sequence.
A binder residue belongs to the interface when any of its heavy atoms lies within 6~\AA{} of a target heavy atom.

The gap between the two redesign conditions isolates how much of a method's score is carried by ProteinMPNN rewriting the interface rather than by the interface the model generated. \citet{didiScalingAtomisticProtein2026} report the same condition for this reason.

Reporting all three conditions permits comparison among methods that use inverse folding differently.
RFdiffusion3 co-designs sequence and structure, but its alanine-biased sequence head is followed by ProteinMPNN in every experiment reported by its authors.
Scoring only its self-sequences would omit part of the published method; Complexa and Prompteina require no redesign.

\subsubsection{Agentic specification generation}

We use Claude Opus 4.6 as the agent in a two-stage pipeline.
A research stage gives a language model tool access to web and scientific-literature search, PDB structure search and entry lookup, and UniProt lookup.
The model uses these sources to prepare a structured dossier about target structure, binding biology, known binders, and biophysical properties (Box~\ref{box:research-prompt}).
An expansion stage reads the dossier and emits the design specifications consumed by each structure generation model (Box~\ref{box:expand-prompt}).

Each method receives the same four classes of input, with only the method reference differing:

\begin{itemize}[leftmargin=1.2em, itemsep=2pt, topsep=3pt, parsep=0pt]
\item \textbf{The research dossier}, byte-identical across methods for a given target.
\item \textbf{The task instruction}, which states how many designs to produce, the length band, and the requirement that designs be diverse. It is byte-identical across methods and names none of them.
\item \textbf{A method reference}, the only input that differs. It describes what that method is, the channels it exposes, the closed set of values each channel accepts, and the exact emission format. Each is self-contained and refers to no other method.
\item \textbf{Structural query tools} over the target, listed in Table~\ref{tab:target-tools}, together with the target configuration (target chain and residue range, binder length range, PDB identifier).
\end{itemize}

\begin{table}[!htbp]
\centering
\caption{\textbf{Structural query tools.}
The tools return target properties rather than raw coordinates, which are too large for the language-model context.}
\label{tab:target-tools}
\footnotesize
\begin{tabular}{@{}>{\raggedright\arraybackslash}p{4.9cm} >{\raggedright\arraybackslash}p{8.3cm}@{}}
\toprule
\textbf{Tool} & Returns \\
\midrule
\texttt{target\_overview()} & Chains, residue ranges, sequence, length, PDB identifier \\
\texttt{secondary\_structure()} & DSSP assignment per chain \\
\texttt{surface\_residues(min\_sasa)} & Exposed residues with per-residue SASA \\
\texttt{neighbors(residue, radius)} & Residues within $r$~\AA, for assembling a contiguous site \\
\texttt{surface\_patches(kind)} & Contiguous hydrophobic, polar, or charged patches \\
\texttt{known\_interfaces()} & Contact residues from deposited complexes for this target \\
\bottomrule
\end{tabular}
\end{table}

The \texttt{known\_interfaces()} tool exposes information already present in the research dossiers and gives every method identical access to deposited complexes.
Removing it only from the structural tools would leave the same information available in prose and would not create a stricter comparison.

The agent selects a binding epitope for each design and writes it through the method's available channel: \texttt{hotspot\_residues} for Prompteina, \texttt{select\_hotspots} atom pairs for RFdiffusion3, and \texttt{binding} properties for BoltzGen.
We validate every index against the target's residue numbering.
A specification that names an absent residue is rejected and recorded, since replacing it with a nearby residue would change the site selected by the agent.

Apart from clamping proposed binder lengths into the target's range, agent output is not modified before it reaches the structure generation model.

\begin{promptbox}{Research stage system prompt (abbreviated)}{box:research-prompt}
You are a protein biology research assistant with access to search tools.\\

Your goal is to help design a STRONG protein binder against the target.
Build a factual research dossier that will inform binder design.\\

PHASE 1 --- IDENTIFY THE TARGET\\
Call pdb\_lookup and uniprot\_lookup to identify the target.\\

PHASE 2 --- GATHER INFORMATION\\
Search systematically: find co-crystal structures with binders, characterize the target surface (shape, epitopes, chemistry), find known binders (scaffold type, size, BSA, affinity, PDB ID).\\

PHASE 3 --- SYNTHESIZE (only from tool results)\\
Write a structured dossier: (1) Target structure, (2) Binding biology, (3) Successful binders, (4) Biophysics.\\

Every PDB ID must be verified by a pdb\_lookup call.
Specific numbers must come from tool results.
Accuracy > completeness.
\end{promptbox}

\begin{promptbox}{Expansion stage task instruction (abbreviated; identical across methods)}{box:expand-prompt}
A generative model will design protein binders against the target described in the research dossier.
Using the method reference provided, produce 10 design specifications.\\

Every design must have a binder length inside the given band.
Use the structure tools to choose a binding site on the target, and name that site explicitly in each design.\\

The 10 designs must be diverse: vary the scaffold, the binder size within the band, and the epitope.
Do not emit the same specification twice.\\

Express every design only through channels the method reference defines, using only values from its closed sets.
\end{promptbox}

\subsubsection{CD45 specification examples}
\label{app:cd45-agent-traces}

All three expansion runs received the same CD45 research dossier and target structure.
Claude generated each method's designs in batches and recorded a rationale for every design.
The examples below pair verbatim excerpts from these saved rationales with the corresponding input sent to each structure model.
The three expansion runs proposed designs independently, so these examples illustrate the available conditioning channels rather than matched design strategies.

\begin{figure}[!htbp]
\centering
\includegraphics[width=0.8477\textwidth]{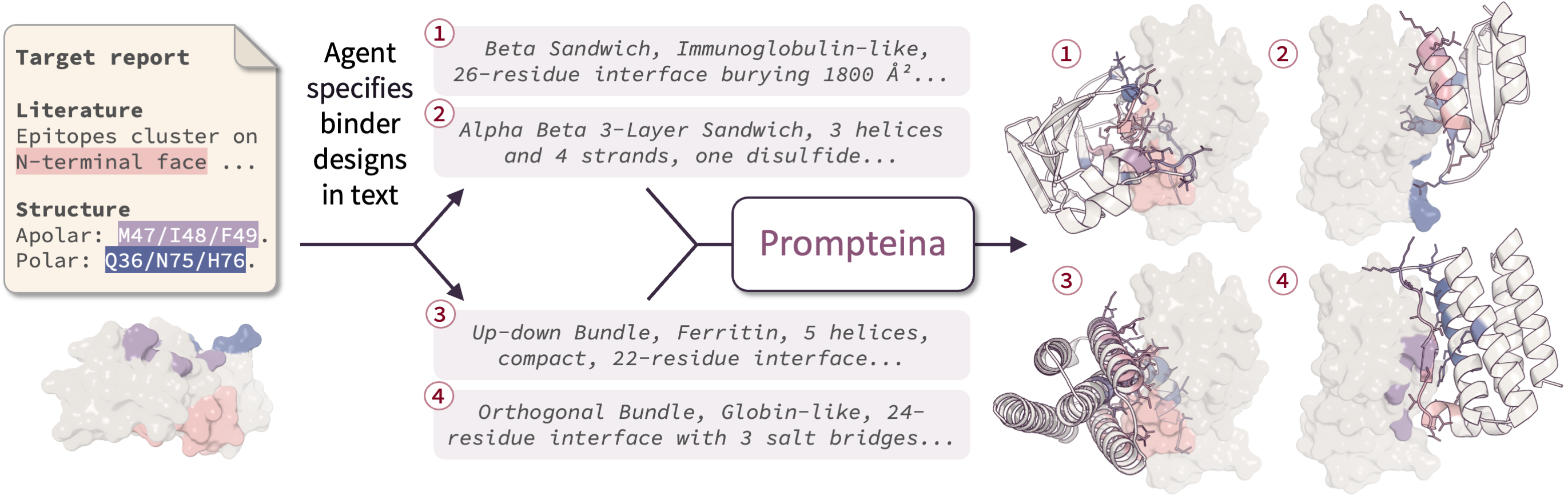}
\caption{\textbf{Prompteina realizes diverse agent specifications for CD45.}
The agent combines literature and structure analysis in a target report, specifies binders with distinct folds and interface properties in natural language, and Prompteina realizes them as structures targeting CD45 sites.}
\label{fig:cd45-designs}
\end{figure}

\paragraph{Prompteina.} For \texttt{ra\_cd45\_114}, Claude's batch-level analysis identifies the ``N-terminal face / BC8 epitope'' as a mixed charged and polar site and proposes ``VHH-inspired loop-mediated contacts to [the] N-terminal face.''
The natural-language specification sent to Prompteina retains the proposed fold, interface size, buried area, chemistry, and geometry:
\begin{quote}
\small\ttfamily
Structural class: Mainly Beta\\
Architecture: Sandwich\\
Topology: Immunoglobulin-like\\
Strands: 9\\
Secondary structure: 45\% strand, 55\% coil\\
Protein-protein interface: 26 interface residues\\
Buried surface area: 1800 A\textasciicircum2\\
Interface H-bonds: 10 total\\
Interface hydrophobic contacts: 12 total\\
Interface geometry: concave, heterologous, mixed
\end{quote}

\paragraph{RFdiffusion3.} For one design, Claude records the following rationale:
\begin{quote}
\small
``Engages the large polar surface patch on Face A (Q36, N37, T39, N75) with directed hydrogen bonds.
This is the most polar-dense contiguous surface on d1, and a helical binder presenting complementary polar side chains should achieve specificity through a network of H-bonds.
The acceptor on Q36 and donor on N75 anchor the interface polar core.''
\end{quote}
The emitted JSON retains a non-loop flag, the selected target atoms, and two atom-level hydrogen-bond labels:
\begin{quote}
\small\ttfamily
is\_non\_loopy: true\\
select\_hotspots: A36, A37, A39, A75\\
select\_hbond\_acceptor: A36.OE1\\
select\_hbond\_donor: A75.ND2
\end{quote}
Across the 200 CD45 specifications, 193 set the non-loop flag, 38 specify a hydrogen-bond acceptor, 17 specify a hydrogen-bond donor, and 7 contain an atom-level burial or exposure label.

\paragraph{BoltzGen.} For another design, Claude records the following rationale:
\begin{quote}
\small
``Targets the MFI hydrophobic patch (M47, I48, F49) in loop 3, a compact surface surrounded by polar/charged residues (R41, Q43, N46, D50, N51, K52).
A mixed $\alpha$/$\beta$ topology uses a $\beta$-hairpin to extend along the loop 3 edge for shape complementarity, while two helices provide hydrophobic packing against M47/I48/F49 and additional polar contacts.''
\end{quote}
The emitted YAML retains residue-level secondary-structure ranges and target binding residues:
\begin{quote}
\small\ttfamily
helix: 27..48, 55..76\\
sheet: 1..8, 13..20\\
loop: 9..12, 21..26, 49..54, 77..82\\
binding: 41, 43, 46, 47, 48, 49, 50, 51, 52
\end{quote}
All 200 BoltzGen CD45 specifications contain residue-level secondary-structure ranges.
These examples show that Claude considered site chemistry and geometry for all three methods, while each model's input interface determined which parts of that reasoning reached generation.

\subsection{Baseline agent specifications}
\label{app:agentic-baselines}

The released Complexa checkpoint takes the target structure, binder length, and target hotspots.
Its authors also report a CATH-conditioned binder checkpoint, but have not released it.
RFdiffusion3 specifications can set binder length and chain layout, atom-level target hotspots, hydrogen-bond partners, burial state, and non-loop secondary structure.
BoltzGen specifications can set binding and avoidance sites, per-residue secondary-structure ranges, and permitted amino acids.

Two protocol differences determine how the agentic and non-agentic runs should be compared.
First, the binding site comes from a different source.
Non-agentic runs receive expert-curated hotspots from the AlphaProteo \citep{zambaldiNovoDesignHighaffinity2024} and BindCraft \citep{pacesaOneshotDesignFunctional2025} campaigns on the 18 targets that provide them.
Agentic runs instead select an epitope for every design, including Prompteina's designs.

Second, the agent populates the baselines' conditioning channels instead of leaving them at their defaults.
A complete parse of all 4,000 specifications per method across the same 20 targets quantifies this use (Table~\ref{tab:knob-coverage}).
The six RFdiffusion3 specifications without an epitope use center-of-mass strategies, for which the agent specifies an inferred orientation instead of a hotspot.

\begin{table}[!htbp]
\centering
\caption{\textbf{Baseline conditioning-channel use.}
Coverage is the fraction of that method's 4{,}000 specification files across the 20 benchmark targets that populate each channel.}
\label{tab:knob-coverage}
\small
\begin{tabular}{@{}llc@{}}
\toprule
\textbf{Method} & \textbf{Channel} & Coverage \\
\midrule
RFdiffusion3 & Target hotspots (\texttt{select\_hotspots})      & 99.9\% \\
             & Non-loop secondary structure (\texttt{is\_non\_loopy}) & 95.3\% \\
             & Hydrogen-bond acceptors                           & 17.9\% \\
             & Hydrogen-bond donors                              & \phantom{0}7.9\% \\
             & Burial, buried (\texttt{select\_buried})          & \phantom{0}2.2\% \\
             & Burial, partial (\texttt{select\_partially\_buried}) & \phantom{0}1.9\% \\
             & Burial, exposed (\texttt{select\_exposed})        & \phantom{0}0.9\% \\
\midrule
BoltzGen     & Secondary structure, per residue                  & 100.0\% \\
             & Loop assignments                                  & 99.2\% \\
             & Negative design (\texttt{not\_binding})           & 81.2\% \\
             & Sheet assignments                                 & 14.2\% \\
             & Amino-acid constraints                            & \phantom{0}0.3\% \\
\bottomrule
\end{tabular}
\end{table}

\subsection{Binder score calibration}
\label{app:ipsae-calibration}

\paragraph{AlphaProteo AF2 filters used by Complexa.} AlphaProteo fit three AlphaFold2 thresholds to the binding data of \citet{caoDesignProteinbindingProteins2022}: interaction PAE below 7~\AA, complex pLDDT above 0.90, and binder-aligned RMSD below 1.5~\AA{} \citep{zambaldiNovoDesignHighaffinity2024}.
The thresholds were selected by ranking experimental success rates within each target, subject to at least ten designs passing the filters.
Complexa subsequently adopted these AF2 filters as its in-silico success criterion \citep{didiScalingAtomisticProtein2026}.
AlphaProteo instead used its AF3 filters for its final benchmark because they performed better across its targets.
These differences motivate recalibrating the evaluation criterion for method comparison.

\paragraph{Calibration data.} We assemble 5,227 designed binders with experimental bind/no-bind labels from four public releases: 3,532 designs over 13 targets from \citet{overathPredictingExperimentalSuccess2025}, 1,189 over 15 targets from \citet{shanehsazzadehAutonomousNovoProtein2026}, 361 over 13 targets from the BoltzGen deposit \citep{starkBoltzGenUniversalBinder2026}, and 145 over 13 targets from the BindCraft deposit \citep{pacesaOneshotDesignFunctional2025}.
We define a group as all designs for one target within one release because releases with the same target name may use different target constructs and different sets of designed binders.
Of 54 such groups, 46 contain both binders and non-binders and can be used to fit and evaluate a threshold; these groups contain 5,045 designs.
We therefore average results equally across these 46 groups rather than pooling all designs.

\paragraph{Prediction and calibration protocol.} We refold every design once with ESMFold2-Fast and ESMFold2 \citep{candidoLanguageModelingMaterializes2026}, AlphaFold2 \citep{jumperHighlyAccurateProtein2021}, and AlphaFold3 \citep{abramsonAccurateStructurePrediction2024a}.
AlphaFold2 uses initial guess, the standard binder-scoring configuration that initializes the structure module from the designed coordinates.
We recompute the published AlphaProteo AF2 filter terms from the saved structures and predicted aligned error matrices under all four models.
For 394 designs in the \citet{shanehsazzadehAutonomousNovoProtein2026} release, the released sequence and deposited design structure contain different sets of target residues, so binder-aligned RMSD cannot be computed for ESMFold2-Fast, ESMFold2, or AlphaFold3.
AlphaFold2 starts from the deposited design structure and therefore retains the residue correspondence needed for RMSD.
We count the 394 designs as not passing filters that require RMSD rather than removing them.
For thresholded criteria, we repeatedly fit the threshold on 45 groups and evaluate it on the remaining group.
The median fitted iPAE thresholds are 10.31, 18.90, 15.19, and 11.15~\AA{} for ESMFold2-Fast, ESMFold2, AlphaFold2, and AlphaFold3, respectively.
The corresponding median ipSAE$_\mathrm{min}$ thresholds are 0.341, 0.328, 0.301, and 0.504.
Average precision is computed directly from the ranking and requires no threshold.

\begin{table}[!htbp]
\centering
\caption{\textbf{Binder-score calibration.}
Each group is one target in one public release, and the 46 groups receive equal weight.
Rows with published thresholds use them unchanged; for rows with fitted thresholds, each group is tested using a threshold fitted on the other 45.
For average precision, a random ranking scores the binder fraction in each group; averaging these 46 fractions gives 0.299.
Runtimes are median seconds per design for the same 3,532 Overath designs under each predictor.}
\label{tab:calibration}
\small
\begin{tabular}{@{}l cccc@{}}
\toprule
& \textbf{ESMFold2-Fast} & \textbf{ESMFold2} & \textbf{AF2} & \textbf{AF3} \\
\midrule
\multicolumn{5}{@{}l}{\textit{Mean F1 across the 46 groups}} \\
\quad AlphaProteo AF2 filters & 0.232 & 0.222 & 0.398 & 0.375 \\
\quad \quad iPAE $<$ 7~\AA{} & 0.350 & 0.304 & 0.404 & 0.394 \\
\quad \quad complex pLDDT $>$ 0.90 & 0.235 & 0.223 & 0.419 & 0.385 \\
\quad \quad binder-aligned RMSD $<$ 1.5~\AA{} & 0.398 & 0.397 & 0.416 & 0.388 \\
\quad iPAE, fitted threshold & 0.463 & 0.387 & 0.437 & 0.442 \\
\quad ipSAE$_\mathrm{min}$, fitted threshold & \textbf{0.519} & 0.468 & 0.441 & 0.458 \\
\quad ipSAE$_\mathrm{min}$ + binder-aligned RMSD & 0.469 & 0.454 & 0.448 & 0.420 \\
\midrule
\multicolumn{5}{@{}l}{\textit{Mean average precision across the 46 groups}} \\
\quad iPAE & 0.577 & 0.564 & 0.502 & 0.547 \\
\quad ipSAE$_\mathrm{min}$ & \textbf{0.587} & 0.577 & 0.517 & 0.566 \\
\midrule
Seconds per design & \textbf{1.61} & 2.37 & 2.62 & 9.07 \\
Groups with $\geq$1 correctly identified binder & \textbf{45/46} & 44/46 & 44/46 & 43/46 \\
\bottomrule
\end{tabular}
\end{table}

\paragraph{AlphaProteo AF2 filter evaluation.} Under every prediction model, the AlphaProteo AF2 filter set discriminates binders less accurately than any one of its component thresholds (Table~\ref{tab:calibration}).
These comparisons use the published thresholds and do not retune them for the present evaluation.

\paragraph{ipSAE$_\mathrm{min}$ evaluation.} ipSAE$_\mathrm{min}$ restricts a TM-score-like transform of predicted aligned error to residue pairs across the interface and scales its distance parameter by interface size \citep{shanehsazzadehAutonomousNovoProtein2026}.
The \emph{min} variant takes the smaller of the asymmetric binder$\rightarrow$target and target$\rightarrow$binder scores.
Its macro average precision exceeds that of iPAE under every prediction model (Table~\ref{tab:calibration}).
Adding binder-aligned RMSD lowers macro-F1 under three predictors (0.519 to 0.469 for ESMFold2-Fast) and raises it slightly only for AlphaFold2 (0.441 to 0.448), and thus provides no consistent benefit.

ESMFold2-Fast and AlphaFold3 give macro average precision values of 0.587 and 0.566, with median runtimes of 1.61 and 9.07 seconds per design, respectively, when both process the same 3,532 Overath designs; the benchmark therefore uses ESMFold2-Fast.
We report continuous ipSAE$_\mathrm{min}$ directly because the best cutoff depends on the data used to fit it.
Within each target, we average the $k$ highest-scoring designs to measure the part of the distribution that a design campaign would order.
We then average the 20 target-level values so that each target receives equal weight.
Every method contributes 200 designs on each of the same 20 benchmark targets in every comparison.

Let $M(k)$ be the mean ipSAE$_\mathrm{min}$ of a target's $k$ best designs among the $N$ generated designs.
The expected top-$k$ statistic averages $M(k)$ under a log-uniform distribution on $k$, as follows:
\begin{equation}
\label{eq:etopk}
\mathbb{E}\bigl[M(K)\bigr]
=
\Bigl(\textstyle\sum_{k=1}^{N}\tfrac{1}{k}\Bigr)^{-1}
\textstyle\sum_{k=1}^{N} M(k)/k .
\end{equation}
This weighting gives equal mass to each order of magnitude of the design budget.

\paragraph{Limitations.} The calibration is not fully independent of the criteria it evaluates.
Of the 3,532 designs in the Overath release \citep{overathPredictingExperimentalSuccess2025}, 2,585 originate from \citet{caoDesignProteinbindingProteins2022}, whose data AlphaProteo used to set its AF2 filters.
The designs tested by \citet{shanehsazzadehAutonomousNovoProtein2026} were themselves selected using ipSAE$_\mathrm{min}$, so that release cannot independently establish a general cutoff for the same score.
For method comparisons, we therefore use continuous ipSAE$_\mathrm{min}$ and the threshold-free expected top-$k$ statistic rather than classifying designs with a fitted cutoff.
The calibration also covers only single-chain targets, whereas three benchmark targets contain multiple chains.
We use the same rule to merge target chains throughout, but have not measured how this choice affects multichain targets.
Finally, the negative examples are campaign designs selected for experimental testing rather than random sequences.
They provide a relevant comparison among plausible designs but cannot estimate the absolute hit rate of a new prospective campaign.

\subsection{Interface confidence across design budgets}
\label{app:topk-curves}

We compare interface confidence across the full range of retained design budgets in Table~\ref{tab:topk-curves}.
Under \texttt{self} and \texttt{MPNN-FI} Prompteina leads at every budget.
Under \texttt{MPNN}, Prompteina leads only in the middle of the range; RFdiffusion3 leads for $k \le 2$ and BoltzGen leads for $k \ge 64$.
No method therefore leads at every budget in this sequence condition.
The $k{=}1$ column rests on one design per target and separates the method--setting combinations least reliably in this cross-method comparison.

\begin{table}[!htbp]
\centering
\caption{\textbf{Interface confidence across design budgets.}
Each cell is the within-target mean of the target's $k$ best ipSAE$_\mathrm{min}$ values, averaged over 20 targets; the last column is the expected top-$k$ from Eq.~\ref{eq:etopk}.
Every method--setting combination contributes 200 designs per target, so $k{=}200$ is the mean over all designs.
Scores use ESMFold2-Fast at one seed.
\textbf{Bold} marks displayed column maxima.
At \texttt{MPNN} $k{=}4$, RFdiffusion3 exceeds Prompteina by 0.0003, but both round to 0.689 in the table.}
\label{tab:topk-curves}
\footnotesize
\setlength{\tabcolsep}{3.2pt}
\begin{tabular}{@{}ll ccccccccc c@{}}
\toprule
\textbf{Method} & \textbf{Agent} & $k$=1 & 2 & 4 & 8 & 16 & 32 & 64 & 128 & 200 & $\mathbb{E}[M(K)]$ \\
\midrule
\multicolumn{12}{@{}l}{\textit{\texttt{self}}} \\
Prompteina   & \cmark & \textbf{0.647} & \textbf{0.620} & \textbf{0.580} & \textbf{0.534} & \textbf{0.470} & \textbf{0.376} & \textbf{0.253} & \textbf{0.149} & \textbf{0.101} & \textbf{0.455} \\
Complexa     & \xmark & 0.571 & 0.543 & 0.501 & 0.455 & 0.402 & 0.333 & 0.240 & 0.144 & 0.096 & 0.400 \\
BoltzGen     & \cmark & 0.573 & 0.537 & 0.478 & 0.405 & 0.326 & 0.249 & 0.169 & 0.107 & 0.074 & 0.359 \\
BoltzGen     & \xmark & 0.569 & 0.530 & 0.486 & 0.430 & 0.364 & 0.284 & 0.199 & 0.128 & 0.088 & 0.376 \\
RFdiffusion3 & \cmark & 0.475 & 0.425 & 0.355 & 0.258 & 0.161 & 0.093 & 0.051 & 0.026 & 0.017 & 0.237 \\
RFdiffusion3 & \xmark & 0.461 & 0.426 & 0.362 & 0.286 & 0.200 & 0.125 & 0.071 & 0.036 & 0.023 & 0.251 \\
\midrule
\multicolumn{12}{@{}l}{\textit{\texttt{MPNN} redesign}} \\
Prompteina   & \cmark & 0.720 & 0.709 & \textbf{0.689} & \textbf{0.660} & \textbf{0.613} & \textbf{0.545} & 0.445 & 0.308 & 0.212 & \textbf{0.584} \\
Complexa     & \xmark & 0.680 & 0.661 & 0.634 & 0.600 & 0.553 & 0.489 & 0.405 & 0.290 & 0.200 & 0.538 \\
BoltzGen     & \cmark & 0.729 & 0.714 & 0.684 & 0.646 & 0.592 & 0.522 & 0.421 & 0.296 & 0.204 & 0.575 \\
BoltzGen     & \xmark & 0.711 & 0.695 & 0.668 & 0.636 & 0.594 & 0.535 & \textbf{0.454} & \textbf{0.335} & \textbf{0.234} & 0.577 \\
RFdiffusion3 & \cmark & \textbf{0.759} & \textbf{0.726} & \textbf{0.689} & 0.641 & 0.578 & 0.489 & 0.363 & 0.220 & 0.143 & 0.559 \\
RFdiffusion3 & \xmark & 0.698 & 0.680 & 0.658 & 0.624 & 0.581 & 0.521 & 0.434 & 0.320 & 0.218 & 0.563 \\
\midrule
\multicolumn{12}{@{}l}{\textit{\texttt{MPNN-FI} redesign}} \\
Prompteina   & \cmark & \textbf{0.696} & \textbf{0.679} & \textbf{0.659} & \textbf{0.629} & \textbf{0.582} & \textbf{0.503} & \textbf{0.394} & \textbf{0.256} & \textbf{0.174} & \textbf{0.549} \\
Complexa     & \xmark & 0.658 & 0.638 & 0.612 & 0.576 & 0.528 & 0.462 & 0.367 & 0.239 & 0.163 & 0.510 \\
BoltzGen     & \cmark & 0.674 & 0.652 & 0.620 & 0.569 & 0.498 & 0.406 & 0.299 & 0.194 & 0.133 & 0.490 \\
BoltzGen     & \xmark & 0.659 & 0.631 & 0.597 & 0.561 & 0.508 & 0.434 & 0.332 & 0.220 & 0.152 & 0.493 \\
RFdiffusion3 & \cmark & 0.675 & 0.642 & 0.599 & 0.537 & 0.459 & 0.349 & 0.229 & 0.127 & 0.082 & 0.454 \\
RFdiffusion3 & \xmark & 0.664 & 0.635 & 0.602 & 0.548 & 0.475 & 0.383 & 0.272 & 0.164 & 0.106 & 0.469 \\
\bottomrule
\end{tabular}
\end{table}

\begin{table}[!htbp]
\centering
\captionsetup{format=plain}
\caption{\textbf{Target-level head-to-head win rates.}
Each cell reports the number and percentage of the 20 targets on which Prompteina's per-target expected top-$k$ ipSAE$_\mathrm{min}$ exceeds the indicated baseline under the given sequence condition.}
\label{tab:binder-target-win-rates}
\footnotesize
\setlength{\tabcolsep}{4pt}
\begin{tabular}{@{}ll @{\hspace{7pt}} ccc@{}}
\toprule
& & \multicolumn{3}{c}{Targets led / 20 ($\uparrow$)} \\
\cmidrule(lr){3-5}
\textbf{Method} & \textbf{Agent} & self & MPNN & \mbox{MPNN-FI} \\
\midrule
Complexa     & \xmark & 13/20 (65\%) & 12/20 (60\%) & 14/20 (70\%) \\
BoltzGen     & \cmark & 15/20 (75\%) & 10/20 (50\%) & 14/20 (70\%) \\
BoltzGen     & \xmark & 16/20 (80\%) & 11/20 (55\%) & 15/20 (75\%) \\
RFdiffusion3 & \cmark & 18/20 (90\%) & 14/20 (70\%) & 17/20 (85\%) \\
RFdiffusion3 & \xmark & 18/20 (90\%) &  9/20 (45\%) & 16/20 (80\%) \\
\bottomrule
\end{tabular}
\end{table}

\subsection{Prediction-model replication}

\begin{table}[!htbp]
\centering
\caption{\textbf{AlphaFold3 replication.}
Expected top-$k$ ipSAE$_\mathrm{min}$ is computed from the same 200 designs per method--setting combination on each of the 20 targets.
Only the \texttt{self} condition was refolded; \texttt{MPNN} and \texttt{MPNN-FI} were not evaluated with AlphaFold3.
AlphaFold3 uses one seed and \texttt{num\_recycles=10}, versus three recycles in Table~\ref{tab:calibration}.}
\label{tab:binder-af3-self}
\small
\setlength{\tabcolsep}{8pt}
\begin{tabular}{@{}llcc@{}}
\toprule
\textbf{Method} & \textbf{Agent} & ESMFold2-Fast & AlphaFold3 \\
\midrule
Prompteina   & \cmark & \textbf{0.455} & \textbf{0.518} \\
\midrule
Complexa     & \xmark & \underline{0.400} & \underline{0.434} \\
BoltzGen     & \cmark & 0.359 & 0.420 \\
BoltzGen     & \xmark & 0.376 & 0.404 \\
RFdiffusion3 & \cmark & 0.237 & 0.276 \\
RFdiffusion3 & \xmark & 0.251 & 0.293 \\
\bottomrule
\end{tabular}
\end{table}

Prompteina remains first under AlphaFold3, with a larger margin over Complexa than under ESMFold2-Fast (0.084 vs.\ 0.055).
The ordering of agent-guided and non-agentic BoltzGen reverses between folding models, so we do not interpret the direction of that comparison.

\subsection{Generated binder--target interface properties}
\label{app:interface-properties}

We analyze generated-backbone properties to isolate the design model's contribution, then measure the same properties after refolding to test whether they persist under sequence redesign.
The \texttt{self} condition refolds the unchanged generated sequence with ESMFold2-Fast; differences from the generated backbone therefore reflect changes in structure rather than sequence.
The other two refolds use the best full ProteinMPNN redesign or the best fixed-interface redesign.
Among the baselines, full ProteinMPNN redesign increases the median interface hydrogen-bond count by one to three, while refolding the generated sequence or preserving interface residues changes the median by at most one (Table~\ref{tab:interface-properties}).
Prompteina's median buried-surface-area lead over the best baseline narrows from 149~\AA$^2$ on generated backbones to 99~\AA$^2$ after \texttt{MPNN-FI} and 55~\AA$^2$ after \texttt{MPNN}.

\begin{table}[!htbp]
\centering
\caption{\textbf{Interface hydrogen bonds after sequence redesign.}
Values are median interface hydrogen-bond counts per design, measured on the generated backbone and on the structure folded under each sequence condition.
Agentic and non-agentic runs are separate.}
\label{tab:interface-properties}
\small
\begin{tabular}{@{}ll cccc@{}}
\toprule
\textbf{Method} & \textbf{Agent} & Generated & \texttt{self} & \texttt{MPNN} & \texttt{MPNN-FI} \\
\midrule
Prompteina   & \cmark & \textbf{10} & \textbf{8} & \textbf{9} & \textbf{9} \\
Complexa     & \xmark & 7 & 7 & 8 & 8 \\
BoltzGen     & \cmark & 6 & 6 & 7 & 6 \\
BoltzGen     & \xmark & 7 & 7 & 8 & 7 \\
RFdiffusion3 & \cmark & 5 & 6 & 8 & 6 \\
RFdiffusion3 & \xmark & 6 & 5 & \textbf{9} & 6 \\
\bottomrule
\end{tabular}
\end{table}

\looseness=-1
Holding the agent fixed, Prompteina generates larger binder--target interfaces than the structured-conditioning baselines (median buried surface area 1,020~\AA$^2$ vs.\ 710--871~\AA$^2$; Table~\ref{tab:binder_mech}).
Its ipSAE$_\mathrm{min}$ rises across within-target buried-surface-area quartiles at every budget (Q1 to Q4: 0.280 to 0.676 at $k{=}1$, 0.016 to 0.421 at $k{=}32$, and 0.070 to 0.207 over all 200 designs).
Among generated-backbone properties, buried surface area has the strongest mean within-target correlation with the \texttt{MPNN-FI} score (0.18 across the six method--setting combinations, compared with 0.14 for interface residue count and 0.09 for interface hydrogen bonds).
The association holds for every method except RFdiffusion3 when the generated interface is preserved (Spearman $\rho$ 0.26 for Complexa, 0.22--0.23 for BoltzGen, 0.18 for Prompteina, and 0.07--0.09 for RFdiffusion3).
Under free \texttt{MPNN} redesign, RFdiffusion3's correlation rises to 0.24--0.27 as ProteinMPNN rebuilds interface contacts.

Matching designs on generated buried surface area changes Prompteina's lead only slightly (13 to 12 of 20 targets over BoltzGen and 17 to 18 of 20 over RFdiffusion3).
Matching on generated interface hydrogen bonds also retains the RFdiffusion3 gap (16 of 20 targets), as do the best-design score and the number of designs above 0.5 (mean best score 0.696 vs.\ 0.675; median yield 22 vs.\ 8 per target).
Buried surface area therefore explains only part of the difference; on unseen targets, no combination of measured per-design properties predicts ipSAE$_\mathrm{min}$ beyond $r\approx0.4$.

\begin{table}[!htbp]
\centering
\caption{\textbf{Generated interface properties and interface confidence.}
All methods use the same agent, and properties are measured on generated backbones.
Buried-surface-area quartiles are defined within target across methods.
The final rows report \texttt{MPNN-FI} scores for Q1 and Q4.}
\label{tab:binder_mech}
\small
\setlength{\tabcolsep}{4.0pt}
\begin{tabular}{@{}l ccc@{}}
\toprule
& \textbf{Prompteina} & \textbf{BoltzGen} & \textbf{RFdiffusion3} \\
\midrule
\multicolumn{4}{@{}l}{\textit{Realized properties}} \\
\quad Buried surface area, p10 (\AA$^2$)    & \textbf{691}  & 468  & 400  \\
\quad Buried surface area, median (\AA$^2$) & \textbf{1020} & 728  & 710  \\
\quad Buried surface area, p90 (\AA$^2$)    & \textbf{1422} & 1073 & 1205 \\
\quad Designs in the top quartile of area   & \textbf{47\%} & 11\% & 17\% \\
\midrule
\multicolumn{4}{@{}l}{\textit{ipSAE$_\mathrm{min}$ across quartiles of buried surface area}} \\
\quad Q1 (508 \AA$^2$) & 0.116 & 0.279 & 0.361 \\
\quad Q4 (1175 \AA$^2$) & \textbf{0.596} & 0.442 & 0.313 \\
\bottomrule
\end{tabular}
\end{table}

\subsection{Binder specification ablation}
\label{app:prompt-ablation}

We re-run the agentic pipeline of \S\ref{sec:exp-ra-binder} with a reduced specification interface to measure the contribution of agent-specified properties.
At each level the agent plans from scratch: it reads the same research dossier, but the design properties it may write are restricted, and it chooses its own epitope and binder length under that restriction.
Fold-only exposes structural class, architecture and topology.
Fold + shape adds secondary structure, helix and strand counts, compactness, surface geometry, contact order, charge character and surface hydrophobicity.
The default interface adds interface residue count, buried surface area, interface geometry, hydrogen bonds, hydrophobic contacts and salt bridges.

Each level generates 200 designs per target on the same 20 targets, and we score the generated sequence, since the ablation targets what conditioning produces rather than what ProteinMPNN recovers.
The default level is the Prompteina row of Table~\ref{tab:binder_eval} and is not regenerated.

The full specification exceeds fold + shape by 0.091 expected top-$k$ (95\% CI $[0.041, 0.145]$, paired bootstrap over the 20 targets) and has the highest score at every selection budget (Table~\ref{tab:prompt-ablation}).
Fold + shape, in contrast, exceeds fold only by 0.002 ($[-0.036, 0.037]$).
Thus, the improvement comes from specifying interface properties rather than from adding more properties in general.
Nine of the 20 targets follow the aggregate ordering of full specification, fold + shape, and fold only, compared with 3.3 expected under a random ordering of the three settings (Binomial$(20, 1/6)$, $p = 0.003$).
For each of these nine targets, the confidence interval for the difference between the full and fold-only settings excludes zero.
CrSAS6 and Insulin favor the reverse ordering with confidence intervals that exclude zero, while five targets show no clear difference among the settings.

\subsection{Structural diversity}
\label{app:diversity}

Diversity is measured on the same design pools as interface confidence, allowing the two quantities to be compared at a common selection budget.
For each method and target, we rank 200 designs by ipSAE$_\mathrm{min}$, retain the best $k$, remove the target chain, and align every binder pair with exact TMalign.
Pair similarity is $\mathrm{TM}=\min(\mathrm{qTM},\mathrm{tTM})$, where lower similarity indicates greater structural diversity.
TM-score depends on length, so we compare binders only within the same 25-residue length bin and average first within bins, then across bins, and finally with equal weight across targets.

Designs with ipSAE$_\mathrm{min}=0$ have an interface but no confident inter-chain geometry, and deep top-$k$ pools can otherwise be dominated by ties at zero.
We report a target at a given $k$ only when every method has at least $k$ positive-scoring designs.
At $k=32$, full and fixed-interface redesign retain 19 and 18 of 20 targets (Table~\ref{tab:diversity}).
We emphasize $k=32$ because $k=64$ retains only 17 and 13 targets, respectively, and preferentially drops difficult ones, changing panel difficulty as well as size.

\begin{table}[!htbp]
\centering
\caption{\textbf{Structural diversity among high-scoring designs.}
Values are mean pairwise TM-scores among each method's $k$ best designs under \texttt{MPNN} and \texttt{MPNN-FI} redesign, where lower is more diverse.
The four columns retain 20, 19, 19, and 17 targets under \texttt{MPNN} redesign and 19, 19, 18, and 13 under \texttt{MPNN-FI} redesign.}
\label{tab:diversity}
\small
\begin{tabular}{@{}ll cccc@{}}
\toprule
\textbf{Method} & \textbf{Agent} & $k{=}8$ & 16 & 32 & 64 \\
\midrule
\multicolumn{6}{@{}l}{\textit{\texttt{MPNN} redesign}} \\
Prompteina   & \cmark & 0.433 & 0.441 & 0.424 & 0.423 \\
Complexa     & \xmark & 0.492 & 0.492 & 0.467 & 0.478 \\
BoltzGen     & \cmark & 0.480 & 0.486 & 0.496 & 0.496 \\
BoltzGen     & \xmark & 0.635 & 0.653 & 0.619 & 0.612 \\
RFdiffusion3 & \cmark & 0.418 & 0.412 & 0.414 & 0.414 \\
RFdiffusion3 & \xmark & 0.482 & 0.470 & 0.461 & 0.466 \\
\midrule
\multicolumn{6}{@{}l}{\textit{\texttt{MPNN-FI} redesign}} \\
Prompteina   & \cmark & 0.455 & 0.438 & 0.416 & 0.410 \\
Complexa     & \xmark & 0.536 & 0.504 & 0.480 & 0.468 \\
BoltzGen     & \cmark & 0.463 & 0.476 & 0.484 & 0.494 \\
BoltzGen     & \xmark & 0.640 & 0.616 & 0.601 & 0.630 \\
RFdiffusion3 & \cmark & 0.434 & 0.413 & 0.413 & 0.412 \\
RFdiffusion3 & \xmark & 0.459 & 0.454 & 0.459 & 0.451 \\
\bottomrule
\end{tabular}
\end{table}

The pooled mean and per-target comparison differ for agent-guided RFdiffusion3, which has the lowest pooled similarity but splits the targets almost evenly with Prompteina.
We therefore report both summaries.
At $k=32$, Prompteina is more diverse under full and fixed-interface redesign on 13 of 19 and 15 of 18 targets against Complexa, 16 of 19 and 15 of 18 against agent-guided BoltzGen, 19 of 19 and 18 of 18 against non-agentic BoltzGen, and 13 of 19 and 14 of 18 against non-agentic RFdiffusion3, respectively.
It splits the targets with agent-guided RFdiffusion3 under both conditions (8 of 19 and 9 of 18).
Prompteina's binding-quality advantage therefore does not correspond to lower structural diversity in these top-ranked pools.

\begin{table}[!htbp]
\centering
\caption{\textbf{Secondary-structure composition among high-scoring designs.}
Statistics use the same top-32 \texttt{MPNN-FI} pools, computed within each target and averaged over the 18 retained targets.
No column has a preferred direction.
Responsiveness is the standard deviation of a method's per-target mean helix fraction, so a low value indicates little change among targets.}
\label{tab:ss-composition}
\small
\setlength{\tabcolsep}{4.5pt}
\begin{tabular}{@{}ll ccc ccc c@{}}
\toprule
& & \multicolumn{3}{c}{Helix fraction} & & & \\
\cmidrule(lr){3-5}
\textbf{Method} & \textbf{Agent} & p10 & p50 & p90 & Mainly & Strand & Respon- \\
& & & & & helical & segments & siveness \\
\midrule
Prompteina   & \cmark & 0.312 & 0.729 & 0.874 & 45\% & 0.7 & 0.127 \\
Complexa     & \xmark & 0.278 & 0.565 & 0.777 & 27\% & 2.3 & 0.201 \\
BoltzGen     & \cmark & 0.664 & 0.823 & 0.880 & 68\% & 0.0 & 0.035 \\
BoltzGen     & \xmark & 0.163 & 0.406 & 0.764 & 29\% & 5.0 & 0.227 \\
RFdiffusion3 & \cmark & 0.320 & 0.753 & 0.866 & 50\% & 0.5 & 0.120 \\
RFdiffusion3 & \xmark & 0.529 & 0.796 & 0.872 & 63\% & 0.2 & 0.102 \\
\bottomrule
\end{tabular}
\end{table}

Agent-guided BoltzGen varies least among targets in secondary-structure composition (responsiveness 0.035 vs.\ 0.201 and 0.227 for the two non-agentic methods; Table~\ref{tab:ss-composition}).
It produces no designs below 0.20 helix fraction and has a median of zero strand segments.
It is also the only method whose pairwise similarity increases with $k$ (0.463 at $k=8$ to 0.494 at $k=64$), indicating that its additional designs repeat similar folds.

Secondary-structure composition and structural diversity measure different variation.
Non-agentic BoltzGen has the widest helix spread, the most strand segments, and the most $\beta$-containing designs, yet also the highest mean pairwise TM-score.
A method can therefore vary secondary-structure content while emitting similar overall folds.

\FloatBarrier
\section{Iterative binder design}

\subsection{Iterative design protocol and analysis}
\label{app:iteropt}

Each iterative loop uses the same 200-design budget as the one-shot benchmark, divided into four rounds of 50 (\S\ref{sec:exp-ra-binder}).
The first round follows the one-shot agentic protocol, so iterative and independent generation differ only after the first 50 designs are scored (App.~\ref{app:agentic-binder}).
Generation, redesign, and scoring otherwise remain identical.

Each 50-design round divides the target's binder-length range into 10 bands and draws five designs per band.
We recompute these bands from the full range instead of merging the 20-band grid used for 200 designs, reproducing the length distribution of an independent 50-design run.

\paragraph{Feedback.} After each round the agent receives every design produced so far, ranked by the objective
\begin{equation*}
y = \max\bigl(\mathrm{ipSAE}_\mathrm{min}(\texttt{self}),\; \max_j \mathrm{ipSAE}_\mathrm{min}(\texttt{MPNN},\, j),\; \max_j \mathrm{ipSAE}_\mathrm{min}(\texttt{MPNN-FI},\, j)\bigr),
\end{equation*}
maximized over the eight redesigned sequences $j$ for each design.
Each feedback entry contains the binder length, epitope, generating specification, per-condition scores, and the maximum and median across redesigned sequences.
Reporting both distinguishes a design with one favorable sequence, such as maximum 0.53 and median 0.05, from a backbone that tolerates redesign, such as maximum 0.40 and median 0.33.

The feedback contains no summary statistics, and the agent is asked to draw its own conclusions.
Roll-ups by design property are omitted because they are confounded by length: ranking architectures by mean score promotes those that happen to occur in long binders, and architecture does not predict the score once length is accounted for.

\paragraph{Rounds.} A single agent call plans all 50 designs in each round.
Parallel calls would receive the same prompt without seeing one another's choices and can therefore produce overlapping designs.
Planning the round jointly also lets the agent distribute designs across binding sites and scaffolds.

\paragraph{Claudin-1 trajectory.}
\label{app:iteropt-claudin}
The seed, round-1, and round-2 designs compared in Fig.~\ref{fig:iteropt} use the same six-residue epitope, up-down bundle architecture, convex mixed interface, and two requested intra-chain disulfides.
The complete reasoning and planning passages excerpted in the figure are reproduced below.

\begin{quote}\small
\emph{Seed to round 1---reasoning.}
``Pulling together the key patterns across all designs: Up-down Bundle is clearly the dominant successful architecture, the helical face epitope scored best overall at 0.667, small ECL1 binders work well for site B, larger Up-down Bundles suit the strand-helix junction, disulfide bonds seem to boost binding, and convex/mixed interface geometries outperform flatter or more hydrophobic ones, with moderate interface sizes around 20--22 residues looking optimal.''

\emph{Seed to round 1---planning.}
``I'm sketching 15 designs for this group, staying close to the winning design while probing topology and parameter variations: several PLA2-based scaffolds at different sizes (125--140 residues) with convex or flat mixed/hydrophobic surfaces and 2 disulfides, plus alternative topologies using colicin Ia, a smaller 4-helix bundle, acyl-CoA binding protein, growth hormone, and bacteriocin As-48 scaffolds, each tested with similar disulfide and interface parameters.''

\emph{Round 1 to round 2---reasoning.}
``Comparing \#2 vs \#4, the convex geometry with more interface residues, BSA, and H-bonds outperforms the flat version. And comparing \#1 vs \#2, growth hormone's 4-helix topology edges out PLA2's 5-helix design despite fewer interface residues and less BSA, suggesting topology choice matters as much as raw interface size.''

\emph{Round 1 to round 2---planning.}
``I should focus most designs on refining around the two best epitopes: A89,A92,A93,A96,A118,A121 paired with the Growth Hormone topology (which produced the best result) and also tested with PLA2 at optimal size, plus the A86,A89,A92,A93,A96,A121 variant with both topologies. I'll also try small perturbations like swapping A85 or A86 in, and tune interface parameters toward convex geometry with 20--24 interface residues around 1500--1700.''
\end{quote}

The selected round-1 design changes the requested topology from phospholipase A2 to growth hormone.
The selected round-2 design keeps that topology and increases the requested interface from 20 to 24 residues and from 1500 to 1700~\AA$^2$ buried surface area (Table~\ref{tab:claudin-trajectory}).

\begin{table}[!htbp]
\centering
\caption{\textbf{Claudin-1 trajectory under iterative prompting.}
Generated properties are measured directly on model outputs; ipSAE$_\mathrm{min}$ is computed on their ESMFold2-Fast refolds.}
\label{tab:claudin-trajectory}
\small
\setlength{\tabcolsep}{8pt}
\begin{tabular}{@{}lccc@{}}
\toprule
Property & Initial & Round 1 & Round 2 \\
\midrule
\multicolumn{4}{@{}l}{\textit{Requested specification}} \\
\quad Topology & Phospholipase A2 & Growth hormone & Growth hormone \\
\quad Helices & 5 & 4 & 5 \\
\quad Interface residues & 22 & 20 & 24 \\
\quad Buried surface area (\AA$^2$) & 1600 & 1500 & 1700 \\
\midrule
\multicolumn{4}{@{}l}{\textit{Generated structure}} \\
\quad Binder length & 135 & 136 & 141 \\
\quad Helices (DSSP) & 7 & 4 & 4 \\
\quad Interface residues & 32 & 21 & 33 \\
\quad Buried surface area (\AA$^2$) & 1298 & 898 & 1327 \\
\midrule
ipSAE$_\mathrm{min}$ & 0.667 & 0.737 & 0.802 \\
\bottomrule
\end{tabular}
\end{table}

\paragraph{Claudin-1 comparison.}
Figure~\ref{fig:iteropt}a shows the top-1 gain; the same comparison at top-8 gives the same method ordering (Table~\ref{tab:iteropt-claudin}).

\begin{table}[!htbp]
\centering
\caption{\textbf{Iterative improvement on Claudin-1.}
Each value is the gain in cumulative top-$k$ mean ipSAE$_\mathrm{min}$ over the exact expectation from a matched-budget i.i.d.\ extension of the shared 50-design seed.}
\label{tab:iteropt-claudin}
\footnotesize
\setlength{\tabcolsep}{6pt}
\begin{tabular}{@{}llrrrr@{}}
\toprule
& & \multicolumn{4}{c}{Designs evaluated} \\
\cmidrule(lr){3-6}
Method & $k$ & 50 & 100 & 150 & 200 \\
\midrule
Prompteina   & 1 & 0.000 & $+$0.060 & $+$0.114 & $+$0.119 \\
             & 8 & 0.000 & $+$0.121 & $+$0.167 & $+$0.181 \\
\midrule
BoltzGen     & 1 & 0.000 & $-$0.002 & $+$0.019 & $+$0.029 \\
             & 8 & 0.000 & $+$0.013 & $+$0.051 & $+$0.056 \\
\midrule
RFdiffusion3 & 1 & 0.000 & $+$0.002 & $-$0.012 & $+$0.013 \\
             & 8 & 0.000 & $+$0.048 & $+$0.035 & $+$0.041 \\
\bottomrule
\end{tabular}
\end{table}

\paragraph{Baseline reasoning traces.}
When paired with BoltzGen or RFdiffusion3, the agent also extracts coherent patterns from feedback; both iterative runs improve over matched-budget i.i.d.\ generation on Claudin-1, although by less than Prompteina.
Before round 2, the BoltzGen agent concludes that ``Helix-$\beta$hairpin-helix-helix (H-SS-H-H) dominates'' and assigns 70\% of the next batch to variants of that topology at the identified length range.
Before round 3, the RFdiffusion3 agent finds that ``The A47+A67 core (ECL1 $\beta$-sheet hydrophobic crown) dominates all top scores'' and assigns about 70\% of the next batch to systematic hotspot and length sweeps around that site.
Thus, the smaller baseline gains do not reflect a failure to use the feedback.
These single-target traces suggest that Prompteina gives the agent more ways to act on feedback: it can revise topology, architecture, interface geometry and size, and disulfides, whereas through BoltzGen and RFdiffusion3 it mainly revises secondary-structure ranges, atom-level target constraints, and binder length.

\subsection{Full-panel evaluation}
\label{app:iteropt-panel}

We run the same iterative loop on all 20 targets to test whether these gains extend across the full benchmark.
Each method receives 200 designs per target.

We define target difficulty as the mean expected top-$k$ ipSAE$_\mathrm{min}$ of the three i.i.d.\ runs on that target, averaged over the three sequence-redesign conditions.
A lower value indicates that every method performs worse on the target.
Difficulty is a property of the i.i.d.\ runs alone, so the loop cannot move it.
IL-17A and TNF-$\alpha$ rank second and third hardest of the 20.
We split the panel into difficulty thirds and report each method's gain over its own i.i.d.\ pool under the same top-$k$ metric (Table~\ref{tab:iteropt-panel}).

\paragraph{Independent difficulty check.}
An independent check separates the difficulty definition from the gain estimate.
We define difficulty using a deterministic, length-stratified subset of 50 i.i.d.\ designs per method and target, then evaluate gains using the disjoint remaining 150 i.i.d.\ designs and 150 agent-generated designs from rounds 1--3.
This independent definition produces exactly the same hardest, middle, and easiest target sets.
We use this check only to verify the target grouping rather than to compare gain magnitudes because it changes the evaluation budget and removes the shared seed used in the primary analysis.

Prompteina gains most on the hardest third at both $k$=1 and $k$=8, and its gains there exceed those of the other methods.
BoltzGen follows the same pattern at $k$=1 but gains most on the middle third at $k$=8.
RFdiffusion3 loses ground on the hardest targets at $k$=1 and is at chance across the panel; its gain comes from the easier two thirds and from $k$=8 rather than its best design.

\begin{table}[!htbp]
\centering
\caption{\textbf{Full-panel iterative prompting by target difficulty.}
Every method receives 200 designs per target across the 20-target panel.
Each cell is the mean gain in top-$k$ ipSAE$_\mathrm{min}$ over the same method's 200 i.i.d.\ designs, averaged over the targets in that difficulty third.
Targets improved counts those for which the loop beats i.i.d.\ generation.
\textbf{Bold} marks the best method within a column.}
\label{tab:iteropt-panel}
\footnotesize
\setlength{\tabcolsep}{6pt}
\begin{tabular}{@{}ll ccc c@{}}
\toprule
& & \multicolumn{3}{c}{Gain over i.i.d. ($\uparrow$)} & Targets \\
\cmidrule(lr){3-5}
\textbf{Method} & $k$ & Hardest 7 & Middle 6 & Easiest 7 & improved \\
\midrule
Prompteina   & 1 & \textbf{$+$0.098} & \textbf{$+$0.046} & \textbf{$+$0.014} & \textbf{17/20} \\
BoltzGen     & 1 & $+$0.051 & $+$0.042 & $+$0.012 & 16/20 \\
RFdiffusion3 & 1 & $-$0.021 & $+$0.004 & $+$0.003 & 11/20 \\
\midrule
Prompteina   & 8 & \textbf{$+$0.105} & $+$0.073 & $+$0.023 & \textbf{20/20} \\
BoltzGen     & 8 & $+$0.067 & \textbf{$+$0.087} & $+$0.022 & \textbf{20/20} \\
RFdiffusion3 & 8 & $+$0.028 & $+$0.055 & \textbf{$+$0.030} & 16/20 \\
\bottomrule
\end{tabular}
\end{table}

The hardest third contains CbAgo, IL-17A, TNF-$\alpha$, BBF14, CD45, HER2, and SpCas9.
The middle third contains Claudin1, DerF7, CrSAS6, VEGFA, DerF21, and BetV1.
The easiest third contains TrkA, IL7RA, IFNAR2, PD1, Insulin, PD-L1, and BHRF1.
Ranking the targets by top-1 or top-8 rather than expected top-$k$ leaves the six hardest unchanged.
At most six targets cross a boundary, all at the middle-to-easiest edge where the difficulty values are closely spaced.

\section{GM2A experimental validation}

\subsection{Candidate selection}
\label{app:gm2a-assay}

We generated 1,000 backbone designs in five parallel iterative-optimization threads, each contributing 200 designs over four rounds (\S\ref{sec:exp-iteropt}).
Four threads targeted distinct GM2A epitopes; the fifth also required $\beta$-rich binder topologies.
Each backbone yielded 33 sequence variants: the generated sequence and eight samples from each of ProteinMPNN, ProteinMPNN-FI, SolubleMPNN, and SolubleMPNN-FI.
We screened every variant with one seed of ESMFold2-Fast and retained the top 50 backbones from each thread, yielding 250 backbones for deeper evaluation.
For deeper evaluation, we folded each of their 8,250 sequences with two ESMFold2-Fast seeds and two AlphaFold3 seeds.

\paragraph{Sequence and interface filters.}
We rejected sequences containing a homopolymer run of at least four residues, a hydrophobic run of at least 12 residues, four consecutive Lys/Arg residues, or a \texttt{PP} pair within the first 20 residues.
Additional vetoes covered the PURE-expression stall motifs \texttt{PPP}, \texttt{DPP}, \texttt{PPN}, \texttt{PPG}, \texttt{APP}, \texttt{GPP}, \texttt{PPD}, \texttt{PPE}, and \texttt{PPW} \citep{peilDistinctXPPXSequence2013}.
A backbone remained eligible when at least one of its sequence variants passed these filters.
Beyond sequence-level quality, the calibration data revealed a second failure mode: very small predicted interfaces.
Designs in the lowest decile of interface buried surface area performed worse in the calibration data, whereas interface size did not distinguish designs above this cutoff.
We therefore removed sequences below the within-target 10th percentile of ESMFold2-Fast interface buried surface area, which was 574~\AA$^2$ for GM2A.
The sequence and interface filters retained 6,031 sequences from 249 backbones.

\paragraph{Metric development for experimental selection.}
Metrics based on absolute model confidence can rank complexes highly even when none of the designs bind.
We therefore required the ranking metric to enrich binders within targets that yield binders and to avoid assigning high scores to designs from targets where every tested design failed.
The development panel combined data from the threshold-calibration panel in Appendix~\ref{app:ipsae-calibration} with 2,221 additional designs from Adaptyv ProteinBase.
We define a group as all designs for one target within one release because releases may use different target constructs and design sets.
The development panel contained 6,402 designs in 64 groups, including 1,057 binders and seven groups with no binders.

We evaluated established metrics and developed 16 new candidates on this panel.
To limit selection bias during this search, we stated a physical hypothesis and a negative control before evaluating each candidate, then tested candidates individually without fitting combinations of existing features.

Our GM2A campaign selects 90 of 1,000 generated backbones for experimental testing, which sets the relevant selection fraction to $q=0.09$.
A hard cutoff requires rounding for groups of different sizes, so we instead give each design a continuous rank weight.
Let $s_{gi}$ be the score of design $i$ in group $g$, oriented so that higher values rank better, and let $r_{gi}$ be its percentile rank within the group, with tied designs assigned their average rank.
We assign each design the weight
\begin{equation}
w_{gi}=r_{gi}^{1/q-1}.
\end{equation}
This weight has mean $q$ under random ranking and approaches one near the top of the ranking.

We score each metric using two quantities: binder selection and non-binder rejection.
Binder selection measures whether confirmed binders receive more weight than expected under random ranking.
Let $B_g$ denote the confirmed binders in group $g$.
We first average their weights within each group and then average across groups with at least one binder:
\begin{equation}
\mathrm{Binder\ selection}
=\frac{
\displaystyle \operatorname*{mean}_{g:B_g\ne\emptyset}
\left(\operatorname*{mean}_{i\in B_g} w_{gi}\right)-q
}{1-q}.
\end{equation}
The subtraction and rescaling set random ranking to zero and perfect binder selection to one.

The binder-selection term excludes groups with no binders, even though these groups expose the failure mode that motivated the search.
The second term therefore asks whether the non-binders ranked highly within one group also score highly relative to non-binders from other groups.
Let $N_g$ denote the confirmed non-binders in group $g$.
For each non-binder, we calculate its percentile among the non-binders in every other group and average those percentiles to obtain $r_{gi}^{\mathrm{cross}}$.
We then use the within-group weights to summarize the cross-group ranks of the non-binders placed near the top of group $g$:
\begin{equation}
\mathrm{Nonbinder\ rank}_g
=\frac{\sum_{i\in N_g}w_{gi}r_{gi}^{\mathrm{cross}}}
{\sum_{i\in N_g}w_{gi}}.
\end{equation}
High $\mathrm{Nonbinder\ rank}_g$ means that the metric gives unusually high scores to the non-binders it places near the top of group $g$.
Before the search, we gave groups with no binders twice the weight of groups with binders to match the planned campaign's mix of all-failure and successful targets.
We invert their weighted mean so that better rejection gives a larger value:
\begin{equation}
\begin{aligned}
\mathrm{Nonbinder\ rejection}
&=1-\left(
\frac{2}{3}\operatorname*{mean}_{g:B_g=\emptyset}\mathrm{Nonbinder\ rank}_g
+\frac{1}{3}\operatorname*{mean}_{g:B_g\ne\emptyset}\mathrm{Nonbinder\ rank}_g
\right), \\
\mathrm{Score}
&=\mathrm{Binder\ selection}\times\mathrm{Nonbinder\ rejection}.
\end{aligned}
\end{equation}
The product is high only when a metric performs well on both parts of the ranking problem.

\paragraph{Computation of $\mathrm{iPAE}_{\text{target-norm}}$.}
Target-normalized interface PAE ($\mathrm{iPAE}_{\text{target-norm}}$) was identified as the best-performing criterion for experimental selection (Table~\ref{tab:gm2a-ranking-metric}).
Its main advantage over ipSAE$_\mathrm{min}$ was stronger rejection of non-binders from all-failure groups.

\begin{table}[H]
\centering
\caption{\textbf{$\mathrm{iPAE}_{\text{target-norm}}$ better rejects non-binders from all-failure groups.}
The final score multiplies the binder-selection and non-binder-rejection terms.
Intervals for the final score are 95\% confidence intervals from resampling the 64 groups.}
\label{tab:gm2a-ranking-metric}
\small
\begin{tabular}{@{}lccc@{}}
\toprule
\textbf{Metric} & Binder selection & Non-binder rejection & Final score ($\uparrow$) \\
\midrule
$\mathrm{iPAE}_{\text{target-norm}}$ & 0.1030 & 0.3527 & 0.0363 [0.0197, 0.0559] \\
ipSAE$_\mathrm{min}$               & 0.0956 & 0.2703 & 0.0258 [0.0118, 0.0428] \\
\bottomrule
\end{tabular}
\end{table}

$\mathrm{iPAE}_{\text{target-norm}}$ is computed by comparing predicted error across the binder--target interface with predicted error within the target's own fold.
For each target residue that contacts the binder, we use its mean within-target PAE to target residues 8--20~\AA\ away as a target-specific reference.
We divide the larger directional interchain PAE for every binder residue within 8~\AA\ by this reference value and transform the ratio $x$ as $1/(1+x^2)$.
The mean transformed ratio over contacting binder residues gives the interface score for that target residue.
The final score is the 25th percentile of these per-target-residue averages.

Let $T$ and $B$ denote the sets of target and binder residues, respectively.
Let $\mathrm{PAE}_{ij}$ denote the predicted aligned error from residue $i$ to residue $j$, and let $d_{ij}$ denote the distance between their representative atoms, using C$_\beta$ when present and C$_\alpha$ otherwise.
Let $Q_{0.25}$ denote the 25th percentile.
\begin{samepage}
For $i\in T$, define $A_i=\{k\in T\mid8\leq d_{ik}\leq20\}$ and $I_i=\{j\in B\mid d_{ij}<8\}$.
\begin{equation}
\begin{aligned}
\overline{\mathrm{PAE}}_i^{\mathrm{own}} &= \frac{1}{|A_i|}\sum_{k\in A_i}\mathrm{PAE}_{ik}, \\
s_i &= \frac{1}{|I_i|}\sum_{j\in I_i}
\left[1+\left(\frac{\max(\mathrm{PAE}_{ij},\mathrm{PAE}_{ji})}{\overline{\mathrm{PAE}}_i^{\mathrm{own}}}\right)^2\right]^{-1}, \\
\mathrm{iPAE}_{\text{target-norm}}
    &= Q_{0.25}\!\left(\{s_i:i\in T,\ |I_i|>0\}\right).
\end{aligned}
\end{equation}
\end{samepage}

A high value indicates that the predicted interface is confident relative to the target's own fold.
Using the 25th percentile prevents a few confident contacts from dominating the score.

\paragraph{Ranking and final selection.}
To preserve candidates from every design thread during the initial screen, we allocated deeper evaluation evenly across the five threads.
We represented each backbone by the highest $\mathrm{iPAE}_{\text{target-norm}}$ among its 33 sequence variants and selected the top 50 backbones within each thread, reducing 1,000 backbones to 250.

The final ranking aggregates predictions across random seeds and folding models so that no single prediction determines selection.
For each metric and folding model, we took the median across two seeds and standardized these medians over all 8,250 sequences.
We then averaged each standardized metric across ESMFold2-Fast and AlphaFold3.
The final score combines $\mathrm{iPAE}_{\text{target-norm}}$ with measures of binder--target interface confidence, complex fold confidence, and agreement between the designed and refolded complexes:
\begin{equation}
\label{eq:gm2a-final-ranking}
\begin{aligned}
S ={}& z(\mathrm{iPAE}_{\text{target-norm}})
  + 0.50\,z(\mathrm{ipSAE}_{\mathrm{min}}) \\
&+ 0.50\,z(\mathrm{pLDDT}_{\mathrm{complex}})
  - 0.25\,z(\mathrm{RMSD}_{\mathrm{complex}}).
\end{aligned}
\end{equation}
Here $z$ denotes the standardized value.
ipSAE$_\mathrm{min}$ measures binder--target interface confidence, complex pLDDT measures overall fold confidence, and complex RMSD measures disagreement between the designed and refolded complexes.
Prior work and the calibration data supported the direction of each term.
Nearby weightings performed similarly, so we do not interpret these coefficients as fitted optima.
We retained only the highest-ranked sequence for each backbone to avoid testing several sequence variants of the same structure, then sent the top 90 distinct backbones to Adaptyv Bio.

\subsection{Binding assay results}
\label{app:gm2a-binding-results}

SPR identified nine binders among the 89 loaded designs: six medium and three weak binders.
Eight yielded fitted overall apparent $K_{\mathrm{D}}$ values, whereas the remaining weak binder had no fit.
The strongest binder reached an overall apparent $K_{\mathrm{D}}$ of 107.4~nM (95\% CI: 80.9--142.5~nM), with replicate-level fits of 68.8, 129.1, and 137.6~nM.
We use the overall fit of 107~nM when summarizing this design and show the three sensorgrams with their replicate-level values (Fig.~\ref{fig:gm2a-campaign}c).
The ESMFold2-Fast predictions for the six medium binders are computational models rather than experimentally determined complexes (Fig.~\ref{fig:gm2a-campaign}a).

Table~\ref{tab:gm2a-binders} reports the experimental results alongside the in silico quantities used during selection.
Design rank is the candidate's position among the 90 designs selected for testing.
MPNN and SolMPNN denote ProteinMPNN and SolubleMPNN, respectively; FI denotes fixed-interface redesign, which retains the generated interface residues while redesigning the remaining sequence.
Round 0 denotes the initial generation, and Rounds 1--3 denote successive iterative-optimization rounds.
The five metrics are raw rather than standardized values.
iPAE is the mean predicted aligned error across binder--target residue pairs, while pLDDT and RMSD are calculated over the full binder--target complex.
For each metric, we took the median over two seeds within each folding model and then averaged the ESMFold2-Fast and AlphaFold3 values.

\begin{table}[H]
\centering
\caption{\textbf{Nine GM2A designs produced measurable binding.}
Apparent $K_{\mathrm{D}}$ values are overall fits across the available replicates.
Rows are ordered by fitted apparent $K_{\mathrm{D}}$.}
\label{tab:gm2a-binders}
\setlength{\tabcolsep}{1.5pt}
\begin{tabular}{@{}ccccccccc@{}}
\toprule
\multicolumn{1}{c}{Rank} &
\multicolumn{1}{c}{$K_{\mathrm{D}}^{\mathrm{app}}$ (nM)} &
Seq. redesign &
\multicolumn{1}{c}{Round} &
\multicolumn{1}{c}{iPAE (\AA)} &
\multicolumn{1}{c}{$\mathrm{iPAE}_{\text{target-norm}}$} &
\multicolumn{1}{c}{$\mathrm{ipSAE}_{\mathrm{min}}$} &
\multicolumn{1}{c}{pLDDT} &
\multicolumn{1}{c}{RMSD (\AA)} \\
\midrule
16 & 107    & MPNN       & 2 & 5.92 & 0.589 & 0.713 & 0.916 & 2.60 \\
78 & 144    & MPNN       & 3 & 5.95 & 0.425 & 0.665 & 0.910 & 2.82 \\
17 & 238    & MPNN-FI    & 0 & 6.99 & 0.658 & 0.666 & 0.896 & 7.61 \\
68 & 281    & SolMPNN    & 3 & 7.83 & 0.504 & 0.679 & 0.885 & 4.80 \\
39 & 335    & SolMPNN    & 0 & 7.09 & 0.621 & 0.586 & 0.884 & 2.75 \\
13 & 868    & SolMPNN-FI & 3 & 5.82 & 0.630 & 0.689 & 0.913 & 3.02 \\
37 & 3,589  & SolMPNN    & 0 & 6.22 & 0.558 & 0.663 & 0.904 & 2.85 \\
21 & 10,000 & MPNN       & 1 & 4.88 & 0.560 & 0.696 & 0.926 & 1.72 \\
61 & No fit  & SolMPNN    & 3 & 5.86 & 0.469 & 0.695 & 0.914 & 3.52 \\
\bottomrule
\end{tabular}
\end{table}

\end{document}